\documentclass[twoside,11pt]{article}

\usepackage{jmlr2e}
\usepackage{microtype,marvosym} 
\usepackage{amsmath,amssymb,graphicx}

\usepackage{tikz,pgfplots}
\pgfplotsset{compat=newest}
\pgfplotsset{plot coordinates/math parser=false}       

\usepackage{natbib}

\definecolor{lred}{RGB}{200,0,0}
\definecolor{dred}{RGB}{130,0,0}
\definecolor{ldre}{RGB}{190.0515  127.5000  127.5000}

\definecolor{dblu}{RGB}{0,0,130}
\definecolor{lblu}{RGB}{127.5000, 127.5000, 192.3975}

\definecolor{dgre}{RGB}{0,130,0} 
\definecolor{dgra}{RGB}{50,50,50}
\definecolor{mgra}{RGB}{100,100,100}
\definecolor{lgra}{RGB}{220,220,220}

\definecolor{MPG}{RGB}{000,125,122}
\definecolor{ora}{HTML}{FF9933}

\definecolor{AMPurple}{HTML}{663366}
\definecolor{Burgundy}{HTML}{993333}
\definecolor{Coffee}{HTML}{7B6049}
\definecolor{ForestGreen}{HTML}{005826}
\definecolor{Lavender}{HTML}{6E6AB1}
\definecolor{PSLightBlue}{HTML}{7DA7D9}

\newcommand{\mpg}[1]{{\color{MPG} #1}}  
  
\newcommand{\dre}[1]{{\color{dred} #1}}  
\newcommand{\ora}[1]{{\color{ora} #1}}   
\newcommand{\blu}[1]{{\color{dblu} #1}}  

\usepackage{booktabs}

\newcommand{\g}{\,|\,} 
\newcommand{\de}{\partial}
 
\newcommand{\eps}{\epsilon}

\newcommand{\Exp}{\mathbb{E}}

\newcommand{\erf}{\operatorname{erf}} 
\newcommand{\sign}{\operatorname{sign}}

\renewcommand{\Re}{\mathbb{R}}

\newcommand{\diag}{\operatorname{diag}}
 
\newcommand{\N}{\mathcal{N}}

\newcommand{\Trans}{^{\intercal}}

\newcommand{\q}{\quad}
\newcommand{\qq}{\qquad}
\newcommand{\qqq}{\quad\qquad}
\newcommand{\qqqq}{\qquad\qquad}

\renewcommand{\vec}{\boldsymbol}

\newcommand{\GP}{\mathcal{GP}}
\newcommand{\Id}{\vec{I}}

\newcommand{\II}{\mathbb{I}}

\newcommand{\y}{\vec{y}}

\newcommand{\V}{\mathbb{V}}

\usepackage{colonequals}

\usepackage{nicefrac}

\usetikzlibrary{arrows,shapes,plotmarks}

\tikzset{>=stealth'} 
\tikzstyle{graphnode} = 
   [circle,draw=black,minimum size=22pt,text centered,text
     width=22pt,inner sep=0pt] 
\tikzstyle{var}   =[graphnode,fill=white]
\tikzstyle{obs}   =[graphnode,fill=black,text=white]
\tikzstyle{fac}   =[rectangle,draw=black,fill=black!25,minimum size=5pt]
\tikzstyle{facprior} =[rectangle,draw=black,fill=black,text=white,minimum size=5pt]
\tikzstyle{edge}  =[draw=white,double=black,thick,-]
\tikzstyle{prior} =[rectangle, draw=black, fill=black, minimum size=
5pt, inner sep=0pt]
\tikzstyle{dirprior} = [circle, draw=black, fill=black, minimum
size=5pt, inner sep=0pt]

\DeclareSymbolFont{stmry}{U}{stmry}{m}{n}
\DeclareMathSymbol\leftarrowtriangle\mathrel{stmry}{"5E}
\DeclareMathSymbol\rightarrowtriangle\mathrel{stmry}{"5F}
\renewcommand{\gets}{\operatorname*{\leftarrowtriangle}}
\renewcommand{\to}{\operatorname*{\rightarrowtriangle}}

\newcounter{PHcomment}

\usepackage{tensor}
\newcommand{\dk}{\tensor[^{\de}]{k}{}}
\newcommand{\ddk}{\tensor[^{\de^2}]{k}{}}
\newcommand{\ddkd}{\tensor[^{\de^2}]{k}{^{\de}}}
\newcommand{\dddk}{\tensor[^{\de^3}]{k}{}}
\newcommand{\dddkd}{\tensor[^{\de^3}]{k}{^{\de}}}
\newcommand{\kd}{\tensor{k}{^{\de}}}
\newcommand{\dkd}{\tensor[^{\de}]{k}{^{\de}}}

\usepackage{algorithm}
\usepackage{algpseudocode}
\newcommand{\LineComment}[1]{\State \(\blacktriangleright\) \begin{minipage}{0.9\textwidth}
    \textcolor{dblu}{#1}
\end{minipage}}
\newcommand{\ColorComment}[1]{\Comment{\textcolor{dblu}{#1}}}
\newcommand{\Separator}{--------------------------------------------------}

\newcommand{\Hm}{\odot}
\newcommand{\Hd}{\oslash}

\newcommand{\cone}{\textcolor{black}{\mathit{c_1}}}
\newcommand{\ctwo}{\textcolor{black}{\mathit{c_2}}}
\newcommand{\pwolfe}{\textcolor{black}{\mathit{c_W}}}
\newcommand{\offs}{\textcolor{black}{\mathit{\tau}}}
\newcommand{\limi}{\textcolor{black}{\mathit{L}}}
\newcommand{\alphaextr}{\textcolor{black}{\mathit{\alpha_{\text{ext}}}}}
\newcommand{\betaf}{\beta}
\newcommand{\fres}{\textcolor{black}{\mathit{\theta_{\text{reset}}}}}
\newcommand{\BoxComment}[1]{\State\fbox{\parbox[c]{0.9\linewidth}{#1}}}

\newcommand{\outsCol}[1]{\textcolor{dred}{#1}}

\newcommand{\globIn}[1]{\textcolor{ora}{#1}}
\newcommand{\returnCol}[1]{\textcolor{dred}{#1}}
\newcommand{\GPCol}[1]{\textcolor{MPG}{#1}}

\newcommand{\StatexIndent}[1][3]{%
  \setlength\@tempdima{\algorithmicindent}%
  \Statex\hskip\dimexpr#1\@tempdima\relax}

\usepackage[pdftex,color]{changebar}\cbcolor{dred} 

\newcommand{\sgd}{{\sc sgd}}
\newcommand{\AdaGrad}{{\sc AdaGrad}}
\newcommand{\Adam}{{\sc Adam}}
\newcommand{\gp}{{\sc gp}}
\newcommand{\pW}{p^\text{Wolfe}}
\newcommand{\probLS}{{\sc probLS}}

\usepackage{pifont}
\usepackage{placeins}
\makeatletter
\newcommand{\manuallabel}[2]{\def\@currentlabel{#2}\label{#1}}
\makeatother

\newlength\figheight
\newlength\figwidth 

\usetikzlibrary{external}
\tikzset{external/force remake=false}
\tikzsetexternalprefix{ext_figs/}

\jmlrheading{}{}{}{}{}{Maren Mahsereci and Philipp Hennig}

\ShortHeadings{Probabilistic Line Searches}{Mahsereci and Hennig}
\firstpageno{1}

\begin{document}

\title{Probabilistic Line Searches for Stochastic Optimization}

\author{\name Maren Mahsereci \email mmahsereci@tue.mpg.de \\
%       \AND
       \name Philipp Hennig \email phennig@tue.mpg.de \\
       \addr Max Planck Institute for Intelligent Systems\\
       Spemannstra{\ss}e, 72076 T\"ubingen, Germany}

\editor{}

\maketitle

% =======================================================================
\begin{abstract}%   <- trailing '%' for backward compatibility of .sty file
In deterministic optimization, line searches are a standard tool ensuring stability and efficiency. Where only stochastic gradients are available, no direct equivalent has so far been formulated, because uncertain gradients do not allow for a strict sequence of decisions collapsing the search space. We construct a probabilistic line search by combining the structure of existing deterministic methods with notions from Bayesian optimization. Our method retains a Gaussian process surrogate of the univariate optimization objective, and uses a probabilistic belief over the Wolfe conditions to monitor the descent. The algorithm has very low computational cost, and no user-controlled parameters. Experiments show that it effectively removes the need to define a learning rate for stochastic gradient descent.
\end{abstract}


% =======================================================================
\section{Introduction}

\label{sec:introduction}
This work substantially extends the work of \cite{MahHen2015} published at NIPS 2015.
Stochastic gradient descent \citep[\sgd,][]{robbins1951stochastic} is currently the standard in machine learning for the optimization of highly multivariate functions if their gradient is corrupted by noise. This includes the online or mini-batch training of neural networks, logistic regression \citep{Zhang:2004,bottou2010large} and variational models \citep[e.g.][]{hoffman2013stochastic,hensman2012fast,StreamingBayes}. In all these cases, noisy gradients arise because an exchangeable loss-function $\mathcal{L}(x)$ of the optimization parameters $x\in\Re^D$, across a large dataset $\{d_i\}_{i=1\,\dots,M}$, is evaluated only on a subset $\{d_j\}_{j=1,\dots,m}$:
\begin{equation}
  \label{eq:1}
  \mathcal{L}(x) := \frac{1}{M}\sum_{i=1} ^M \ell(x,d_i) 
  \approx \frac{1}{m}\sum_{j=1} ^{m} \ell(x,d_j) =: \hat{\mathcal{L}}(x)\qqqq m\ll M.
\end{equation}%
If the indices $j$ are i.i.d.~draws from $[1,M]$, by the Central Limit Theorem, the error $\hat{\mathcal{L}}(x)-\mathcal{L}(x)$ is unbiased and approximately normal distributed. Despite its popularity and its low cost per step, \sgd~has well-known deficiencies that can make it inefficient, or at least tedious to use in practice. Two main issues are that, first, the gradient itself, even without noise, is not the optimal search \emph{direction}; and second, \sgd~requires a \emph{step size} (learning rate) that has drastic effect on the algorithm's efficiency, is often difficult to choose well, and virtually never optimal for each individual descent step. The former issue, adapting the search direction, has been addressed by many authors \citep[see][for an overview]{george2006adaptive}. Existing approaches range from lightweight `diagonal preconditioning' approaches like \Adam~\citep{DBLP:journals/corr/KingmaB14}, \AdaGrad~\citep{duchi2011adaptive}, and `stochastic meta-descent' \citep{schraudolph1999local}, to empirical estimates for the natural gradient \citep{amari2000adaptive} or the Newton direction \citep{roux2010fast}, to problem-specific algorithms \citep{ranganath13}, and more elaborate estimates of the Newton direction \citep{StochasticNewton}. Most of these algorithms also include an auxiliary adaptive effect on the learning rate. 
\citet{schaul2013no} provided an estimation method to explicitly adapt the learning rate from one gradient descent step to another.
Several very recent works have proposed the use of reinforcement learning and `learning-to-learn' approaches for parameter adaption \citep{DBLP:journals/corr/AndrychowiczDGH16, DBLP:journals/corr/Hansen16, DBLP:journals/corr/LiM16b}.
Mostly these methods are designed to work well on a specified subset of optimization problems, which they are also trained on; they thus need to be re-learned for differing objectives.
The corresponding algorithms are usually orders of magnitude more expensive than the low-level black box proposed here, and often require a classic optimizer (e.g \sgd) to tune their internal hyper-parameters.

None of the mentioned algorithms change the size of the \emph{current} descent step. Accumulating statistics across steps in this fashion requires some conservatism: If the step size is initially too large, or grows too fast, \sgd~can become unstable and `explode', because individual steps are not checked for robustness at the time they are taken.

\begin{figure}[ht]
  \begin{minipage}[c]{0.3\textwidth}
  \centering
  \setlength{\figwidth}{\textwidth}
  \setlength{\figheight}{.38\textheight}
  %\tikzset{external/remake next}
%  {\scriptsize \input{poster_figs/classicLS_Sketch2_07.tikz}}
  \includegraphics[scale=1.0]{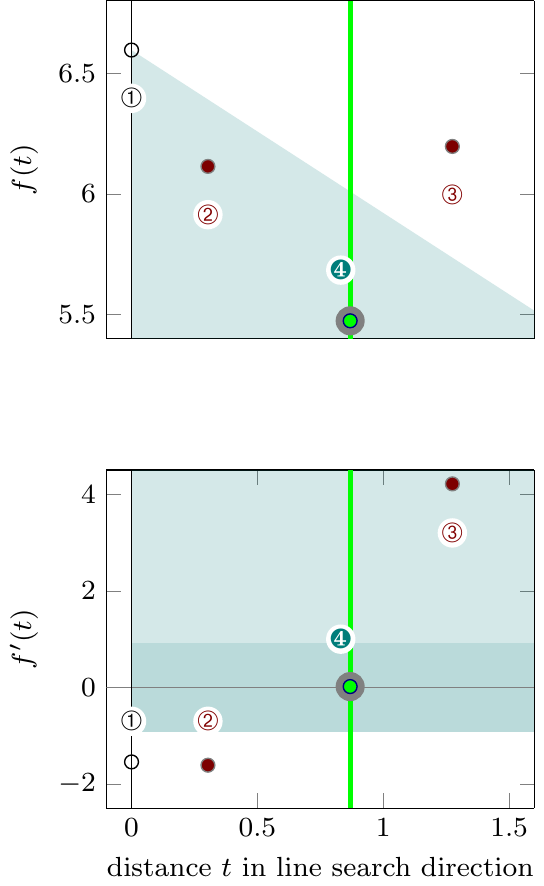}
  \end{minipage}\hfill
  \begin{minipage}[c]{.6\textwidth}
  \caption{Sketch: The task of a \emph{classic line search} is to tune the step taken by an optimization algorithm along a univariate search direction. The search starts at the endpoint \ding{192} of the previous line search, at $t=0$. The \emph{upper} plot shows function values, the \emph{lower} plot corresponding gradients. A sequence of extrapolation steps \ding{193},\ding{194} finds a point of positive gradient at \ding{194}. It is followed by interpolation steps until an acceptable point \ding{185} is found. Points of insufficient decrease, above the line $f(0) + c_1tf'(0)$ (white area in upper plot) are excluded by the Armijo condition W-I, while points of steep negative gradient (white area in lower plot) are excluded by the curvature condition W-II (the strong extension of the Wolfe conditions also excludes the light green area in the lower plot). Point \ding{185} is the first to fulfil both conditions, and is thus accepted.}
\label{fig:minimize-sketch}
  \end{minipage}
\end{figure}

In essence, the same problem exists in deterministic (noise-free) optimization problems. There, providing stability is one of several tasks of the \emph{line search} subroutine. It is a standard constituent of algorithms like the classic nonlinear conjugate gradient \citep{fletcher1964function} and BFGS \citep{broyden1969new,fletcher1970new,goldfarb1970family,shanno1970conditioning} methods \citep[\textsection 3]{nocedal1999numerical}.\footnote{In these algorithms, another task of the line search is to guarantee certain properties of the surrounding estimation rule. In BFGS, e.g., it ensures positive definiteness of the estimate. This aspect will not feature here.} In the noise-free case, line searches are considered a solved problem \citep[\textsection 3]{nocedal1999numerical}. But the methods used in deterministic optimization are not stable to noise. They are easily fooled by even small disturbances, either becoming overly conservative or failing altogether. The reason for this brittleness is that existing line searches take a sequence of hard decisions to shrink or shift the search space. This yields efficiency, but breaks hard in the presence of noise. Section~\ref{sec:method} constructs a probabilistic line search for noisy objectives, stabilizing optimization methods like the works cited above. As line searches only change the length, not the direction of a step, they could be used in combination with the algorithms adapting \sgd's direction, cited above. 
In this paper we focus on parameter tuning of the \sgd~algorithm and  leave other search directions to future work.

% =========================================================================
\section{Connections}
\label{sec:connections}

% =========================================================================
\subsection{Deterministic Line Searches}
\label{sec:determ-lines}
There is a host of existing line search variants \citep[\textsection 3]{nocedal1999numerical}. In essence, though, these methods explore a univariate domain `to the right' of a starting point, until an `acceptable' point is reached (Figure~\ref{fig:minimize-sketch}). More precisely, consider the problem of minimizing $\mathcal{L}(x):\Re^D\to \Re$, with access to $\nabla \mathcal{L}(x):\Re^D\to\Re^D$. At iteration $i$, some `outer loop' chooses, at location $x_i$, a search direction $s_i\in\Re^D$ (e.g.~by the BFGS rule, or simply $s_i=-\nabla \mathcal{L} (x_i)$ for gradient descent). It will \emph{not} be assumed that $s_i$ has unit norm. The line search operates along the univariate domain $x(t) = x_i + t s_i$ for $t\in\Re_+$. Along this direction it collects scalar function values and projected gradients that will be denoted $f(t) = \mathcal{L} (x(t))$ and $f'(t) = s_i\Trans \nabla \mathcal{L} (x(t))\in\Re$.  Most line searches involve an initial extrapolation phase to find a point $t_r$ with $f'(t_r)>0$. This is followed by a search in $[0,t_r]$, by interval nesting or by interpolation of the collected function and gradient values, e.g.~with cubic splines.\footnote{\label{fn:minimize}This is the strategy in {\scriptsize\tt minimize.m} by C.~Rasmussen, which provided a model for our implementation. At the time of writing, it can be found at {\scriptsize\url{http://learning.eng.cam.ac.uk/carl/code/minimize/minimize.m}}}

% =========================================================================
\subsubsection{The Wolfe Conditions for Termination}
\label{sec:wolfe-cond-conv}

\begin{figure}
\begin{minipage}[c]{.45\textwidth}
  \centering
  \setlength{\figwidth}{0.85\textwidth}
  \setlength{\figheight}{.35\textheight}
  %\tikzset{external/remake next}
%  {\scriptsize \input{fig/problinesearch-sketch.tikz}}  
  \includegraphics[scale=1.0]{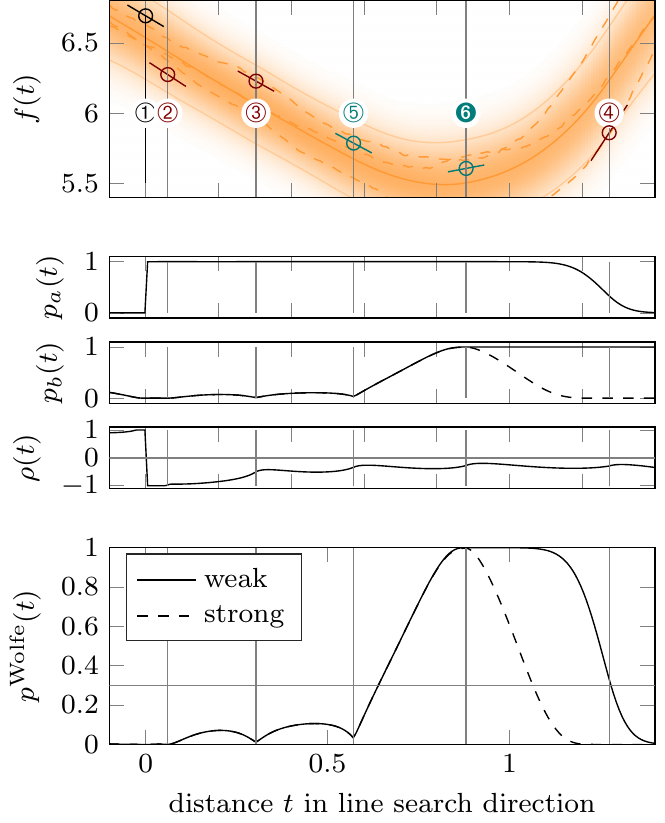}
\end{minipage}
\begin{minipage}[c]{.55\textwidth}
  \caption{Sketch of a \emph{probabilistic line search}. As in Fig.~\ref{fig:minimize-sketch}, the algorithm performs extrapolation (\ding{193},\ding{194},\ding{195}) and interpolation (\ding{196},\ding{207}), but receives unreliable, noisy function and gradient values. These are used to construct a \gp~posterior (top.~solid posterior mean, thin lines at 2 standard deviations, local pdf marginal as shading, three dashed sample paths). This implies a bivariate Gaussian belief (\textsection\ref{sec:determ-conv}) over the validity of the weak Wolfe conditions (middle three plots. $p_a(t)$ is the marginal for W-I, $p_b(t)$ for W-II, $\rho(t)$ their correlation). Points are considered acceptable if their joint probability $\pW(t)$ (bottom) is above a threshold (gray). An approximation (\textsection\ref{sec:appr-strong-wolfe}) to the strong Wolfe conditions is shown dashed.}
\label{fig:prob-ls-sketch}
\end{minipage}
\end{figure}

As the line search is only an auxiliary step within a larger iteration, it need not find an exact root of $f'$; it suffices to find a point `sufficiently' close to a minimum. The \emph{Wolfe conditions} \citep{wolfe1969convergence} are a widely accepted formalization of this notion; they consider $t$ acceptable if it fulfills
\begin{equation}
  \label{eq:4}
  f(t) \leq f(0) + c_1 t f'(0)\q\text{(W-I)}\qq\text{and}\qq  f'(t) \geq c_2 f'(0) \q\text{(W-II)},
\end{equation}
using two constants $0\leq c_1<c_2\leq 1$ chosen by the designer of the line search, not the user. W-I is the \emph{Armijo} or \emph{sufficient decrease} condition \citep{armijo1966minimization}. It encodes that acceptable functions values should lie below a linear extrapolation line of slope $c_1f'(0)$.  W-II is the \emph{curvature condition}, demanding a decrease in slope. The choice $c_1=0$ accepts any value below $f(0)$, while $c_1=1$ rejects all points for convex functions. For the curvature condition, $c_2=0$ only accepts points with $f'(t)\geq 0$; while $c_2=1$ accepts any point of greater slope than $f'(0)$. W-I and W-II are known as the \emph{weak} form of the Wolfe conditions. The \emph{strong} form replaces W-II with $|f'(t)|\leq c_2 |f'(0)|$.  This guards against accepting points of low function value but large positive gradient. Figure \ref{fig:minimize-sketch} shows a conceptual sketch illustrating the typical process of a line search, and the weak and strong Wolfe conditions. The exposition in \textsection\ref{sec:determ-conv} will initially focus on the weak conditions, which can be precisely modeled probabilistically. Section \ref{sec:appr-strong-wolfe} then adds an approximate treatment of the strong form.

% =========================================================================
\subsection{Bayesian Optimization}
\label{sec:bayes-optim}
A recently blossoming sample-efficient approach to global optimization revolves around modeling the objective $f$ with a probability measure $p(f)$; usually a Gaussian process (\gp). Searching for extrema, evaluation points are then chosen by a utility functional $u[p(f)]$. Our line search borrows the idea of a Gaussian process surrogate, and a popular acquisition function, \emph{expected improvement} \citep{jones1998efficient}. Bayesian optimization (\textsc{bo}) methods are often computationally expensive, thus ill-suited for a cost-sensitive task like a line search. But since line searches are governors more than information extractors, the kind of sample-efficiency expected of a Bayesian optimizer is not needed.
The following sections develop a lightweight algorithm which adds only minor computational overhead to stochastic optimization.

% =========================================================================
\section{A Probabilistic Line Search}
\label{sec:method}
We now consider minimizing $f(t)=\hat{\mathcal{L}} (x(t))$ from Eq.~\ref{eq:1}.
% based on noisy value and gradient observations $y_t,y' _t$ at location $t$.
That is, the algorithm can access only noisy function values and gradients $y_t,y' _t$ at location $t$, with Gaussian likelihood
\begin{equation}
  \label{eq:3}
  p(y _t,y' _t\g f) = \N\left(
    \begin{bmatrix}
      y _t\\ y' _t
    \end{bmatrix};
    \begin{bmatrix}
      f(t) \\ f'(t)
    \end{bmatrix},
    \begin{bmatrix}
      \sigma_{f}^2 & 0 \\ 0 & \sigma_{f'} ^2
    \end{bmatrix}
\right).
\end{equation}
The Gaussian form is supported by the Central Limit argument at Eq.~\ref{eq:1}. The function value $y_t$ and the gradient $y_t'$ are assumed independent for simplicity; see \textsection\ref{sec:hyperp-estim} and Appendix A regarding estimation of the variances $\sigma^2 _f,\sigma^2 _{f'}$, and some further notes on the independence assumption of $y$ and $y'$. Each evaluation of $f(t)$ uses a newly drawn mini-batch.

Our algorithm is modeled after the classic line search routine \texttt{minimize.m}\textsuperscript{\ref{fn:minimize}} and translates each of its ingredients one-by-one to the language of probability.
The following table illustrates the four ingredients of the probabilistic line search and their corresponding classic parts.
 \begin{center}
   \renewcommand{\arraystretch}{1.3}
    \begin{tabular}{p{4.5cm} p{4.5cm} p{4.7cm}}
      \textbf{building block} & \textbf{classic} & \textbf{probabilistic}\\
      \hline
      \textbf{1)} 1D surrogate for 

      ~~~~objective $f(t)$ 
      & piecewise cubic splines
      &\gp~where the mean are piecewise cubic splines
      \\
%      \hline
      \textbf{2)} candidate selection 
      & \emph{one} local  minimizer of cubic splines \emph{xor} extrapolation
      & local  minimizers of cubic splines \emph{and} extrapolation
      \\
%      \hline
      \textbf{3)} choice of best candidate 
      & \qqq\q --------- %(only 1 candidate)
      & \textsc{bo} acquisition function 
      \\
%      \hline
      \textbf{4)} acceptance criterion 
      & classic Wolfe conditions
      & probabilistic Wolfe conditions
      \\
    \end{tabular}
\end{center}
The table already motivates certain design choices, for example the particular choice of the \gp-surrogate for $f(t)$, which strongly resembles the classic design.
Probabilistic line searches operate in the same scheme as classic ones: 
1)~they construct a surrogate for the underlying 1D-function 
2)~they select candidates for evaluation which can interpolate between datapoints or extrapolate
3)~a heuristic chooses among the candidate locations and the function is evaluated there
4)~the evaluated points are checked for Wolfe-acceptance.
The following sections introduce all of these building blocks with greater detail:
A robust yet lightweight Gaussian process surrogate on $f(t)$ facilitating analytic optimization (\textsection~\ref{sec:gauss-proc-surr}); a simple Bayesian optimization objective for exploration (\textsection~\ref{sec:select-eval-points}); and a probabilistic formulation of the Wolfe conditions as a termination criterion (\textsection~\ref{sec:determ-conv}). 
Appendix \ref{app:pseudocode} contains a detailed pseudocode of the probabilistic line search; algorithm \ref{alg:sketch} very roughly sketches the structure of the probabilistic line search and highlights its essential ingredients.
\begin{algorithm}[t]
\manuallabel{alg:sketch}{1}
%\label{alg:sketch}
\caption{\textsc{probLineSearchSketch}($f$, $y_0$, $y'_0$, $\sigma_{f_0}$, $\sigma_{f'_0}$)}
    \begin{algorithmic}
%        \Function{probLineSearchSketch}{$f$, $y_0$, $y'_0$, $\sigma_{f_0}$, $\sigma_{f'_0}$}
        \State $GP\gets $\Call{initGP}{$y_0$, $y'_0$, $\sigma_{f_0}$, $\sigma_{f'_0}$} 
        \State $T, Y, Y'\gets $\Call{initStorage}{$0$, $y_0$, $y'_0$}\Comment{for observed points} 
        \State $t\gets 1$\Comment{scaled position of initial candidate}
        \State
        \While{\textbf{budget not used and no Wolfe-point found}}
            \State $[y, y']\gets f(t)$ 
            \Comment{evaluate objective}
            \State $T, Y, Y'\gets $\Call{updateStorage}{$t$, $y$, $y'$}%\Comment{update \gp} 
            \State $GP\gets $\Call{updateGP}{$t$, $y$, $y'$}%\Comment{update \gp} 
            \State $P^{\mathrm{Wolfe}}\gets$\Call{probWolfe}{$T$, $GP$}
            \Comment{compute Wolfe probability at points in $T$}
            \State
            \If{\textbf{any $P^{\mathrm{Wolfe}}$ above Wolfe threshold $c_W$}}
               \State\Return Wolfe-point
            \Else
               \State $T_{\text{cand}}\gets$\Call{computeCandidates}{$GP$} 
               \Comment{positions of new candidates}
               \State $EI\gets$\Call{expectedImprovement}{$T_{\text{cand}}$, $GP$} 
               \State $PW\gets$\Call{probWolfe}{$T_{\text{cand}}$, $GP$} 
               \State $t\gets$ where $(PW\Hm EI)$ is maximal\Comment{find best candidate among $T_{\text{cand}}$}
            \EndIf
        \EndWhile
        \State
        \State\Return observed point in $T$ with lowest \gp~mean since no Wolfe-point found
%        \EndFunction
    \end{algorithmic}
\end{algorithm}

% =========================================================================
\subsection{Lightweight Gaussian Process Surrogate}
\label{sec:gauss-proc-surr}
\begin{figure}
  \centering
  \setlength{\figwidth}{.9\textwidth}
  \setlength{\figheight}{.2\textheight} % .32
  %\tikzset{external/remake next}
  %{\scriptsize \input{fig/sketch_WPandPolyReg_12345678.tikz}}    
  \includegraphics[scale=1.0]{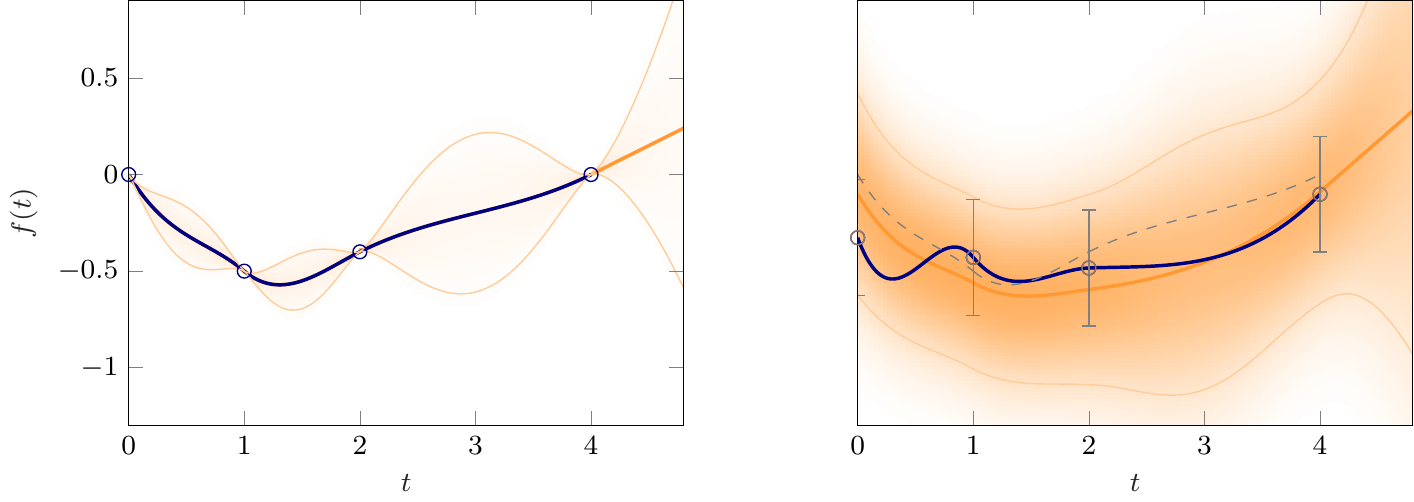}
%  {\scriptsize \input{fig/sketch_WPandPolyReg_123456.tikz}}  
  \caption{\emph{Integrated Wiener process:} \gp~marginal posterior of function values; posterior mean in solid orange and, two standard deviations in thinner solid orange, local pdf marginal as shading; function value observations as gray circles (corresponding gradients not shown). Classic interpolation by piecewise cubic spline in dark blue. \emph{Left:} observations are exact; the mean of the \gp~and the cubic spline interpolator of a classic line search coincide. \emph{Right:} same observations with additive Gaussian noise (error-bars indicate $\pm$ 1 standard deviations); noise free interpolator in dashed gray for comparison.  The classic interpolator in dark blue, which exactly matches the observations, becomes unreliable; the \gp~reacts robustly to noisy observations; the \gp-mean still consists of piecewise cubic splines.}
\label{fig:WPSketch}
\end{figure}
We model information about the objective in a probability measure $p(f)$. There are two requirements on such a measure: First, it must be robust to irregularity (low and high variability) of the objective. And second, it must allow analytic computation of discrete candidate points for evaluation, because a line search should not call yet another optimization subroutine itself. Both requirements are fulfilled by a once-integrated Wiener process, i.e.~a zero-mean Gaussian process prior $p(f)=\GP(f;0,k)$ with covariance function
\begin{equation}
  \label{eq:2}
  k(t,t') = \theta^2 \left[\nicefrac{1}{3}\operatorname{min}^3(\tilde{t},\tilde{t}') +
    \nicefrac{1}{2}|t-t'|\operatorname{min}^2(\tilde{t},\tilde{t}') \right].
\end{equation}%
Here $\tilde{t}:=t+\tau$ and $\tilde{t}':=t'+\tau$ denote a shift by a constant $\tau>0$. This ensures this kernel is positive semi-definite, the precise value $\tau$ is irrelevant as the algorithm only considers positive values of $t$ (our implementation uses $\tau=10$). See \textsection\ref{sec:hyperp-estim} regarding the scale $\theta^2$. With the likelihood of Eq.~\ref{eq:3}, this prior gives rise to a \gp~posterior whose mean function is a cubic spline\footnote{Eq.~\ref{eq:2} can be generalized to the `natural spline', removing the need for the constant $\tau$ \citep[\textsection 6.3.1]{RasmussenWilliams}. However, this notion is ill-defined in the case of a single observation, which is crucial for the line search.} \citep{wahba1990spline}. We note in passing that regression on $f$ and $f'$ from $N$ observations of pairs $(y_t,y'_t)$ can be formulated as a filter \citep{sarkka2013bayesian} and thus performed in $\mathcal{O}(N)$ time. However, since a line search typically collects $<10$ data points, generic \gp~inference, using a Gram matrix, has virtually the same, low cost.

Because Gaussian measures are closed under linear maps \citep[\textsection 10]{papoulis91:probab_random}, Eq.~\ref{eq:2} implies a Wiener process (linear spline) model on $f'$:
\begin{equation}
  \label{eq:6}
    p(f;f') = \GP\left(
      \begin{bmatrix}
        f \\ f'
      \end{bmatrix};% \vec{0},
     \begin{bmatrix}
        0 \\ 0
      \end{bmatrix},
      \begin{bmatrix}
        k & \kd\\ \dk & \dkd
      \end{bmatrix}
    \right),
\end{equation}
with (using the indicator function $\II(x)=1$ if $x$, else 0)
\begin{alignat}{2}
 \label{eq:23}
  \kd_{tt'} 
   &:=\frac{\partial k(t, t')}{\partial t'} 
   &&= \theta^2 \left[ 
       \II(t<t') \frac{\tilde{t}^2}{2} 
       + \II(t\geq t') \left(
        \tilde{t}\tilde{t}' 
        - \frac{\tilde{t}'^2}{2}
      \right) \right]\notag\\
    \dk_{tt'}
   &:=\frac{\partial k(t, t')}{\partial t} 
    &&= \theta^2 \left[ \II(t'<t) \frac{\tilde{t}'^2}{2} + \II(t'\geq t) \left(\tilde{t}\tilde{t}' - \frac{\tilde{t}^2}{2}\right) \right]\\
   \dkd_{tt'}
   &:=\frac{\partial^2 k(t, t')}{\partial t'\partial t} 
   &&= \theta^2 \min(\tilde{t},\tilde{t}').\notag
\end{alignat}
Given a set of evaluations $(\vec{t},\y,\y')$ (vectors, with elements $t_i,y_{t_i},y'_{t_i}$) with independent likelihood \ref{eq:3}, the posterior $p(f\g \y,\y')$ is a \gp~with posterior mean function $\mu$ and covariance function $\tilde{k}$ as follows:
\begin{equation}
  \label{eq:7}
  \begin{split}
  \begin{bmatrix}
    \mu(t)\\\mu'(t)
  \end{bmatrix}  &=
\underbrace{
  \begin{bmatrix}
    k_{t\vec{t}} & \kd_{t\vec{t}}\\
    \dk_{\vec{t}t} & \dkd_{t\vec{t}}
  \end{bmatrix}
    \begin{bmatrix}
      k_{\vec{t}\vec{t}}+\sigma_f ^2 \Id & \kd_{\vec{t}\vec{t}}\\
      \dk_{\vec{t}\vec{t}} & \dkd_{\vec{t}\vec{t}} +\sigma_{f'} ^2 \Id
    \end{bmatrix}^{-1}
  }_{=:\vec{g}\Trans(t)}
  \begin{bmatrix}
    \y \\ \y'
  \end{bmatrix}\\
  \begin{bmatrix}
    \tilde{k}(t,t') & \tilde{\kd}(t,t') \\
    \tilde{\dk}(t',t) & \tilde{\dkd}(t,t') 
  \end{bmatrix}  &= 
  \begin{bmatrix}
    k_{tt'} & \kd_{tt'}\\
    \dk_{t't} & \dkd_{tt'}
  \end{bmatrix}
  - 
  \vec{g}\Trans(t)
  \begin{bmatrix}
    k_{\vec{t}t'} & \kd_{\vec{t}t'}\\
    \dk_{t'\vec{t}} & \dkd_{\vec{t}t'}
  \end{bmatrix}
  \end{split}
\end{equation}
The posterior marginal variance will be denoted by $\V(t)=\tilde{k}(t,t)$. To see that $\mu$ is indeed piecewise cubic (i.e.~a cubic spline), we note that it has at most three non-vanishing derivatives\footnote{There is no well-defined probabilistic belief over $f''$ and higher derivatives---sample paths of the Wiener process are almost surely non-differentiable almost everywhere \citep[\textsection 2.2]{adler1981geometry}. But $\mu(t)$ is always a member of the reproducing kernel Hilbert space induced by $k$, thus piecewise cubic \citep[\textsection 6.1]{RasmussenWilliams}.}, because
\begin{alignat}{4}
\label{eq:9}
  \ddk_{tt'}
  &:=\frac{\partial^2 k(t, t')}{\partial t^2} 
  &&= \theta^2 \II(t\leq t')
  &\dddk_{tt'}
  &:=\frac{\partial^3 k(t, t')}{\partial t^3} 
  &&= \theta^2 \II(t\leq t')(t' - t) \notag\\
  \ddkd_{tt'}
  &:=\frac{\partial^4 k(t, t')}{\partial t^2\partial t'} 
  &&= -\theta^2 \II(t\leq t') 
  &\qq\dddkd_{tt'}
  &:=\frac{\partial^4 k(t, t')}{\partial t^3\partial t'} 
  &&= 0.
\end{alignat}
This piecewise cubic form of $\mu$ is crucial for our purposes: having collected $N$ values of $f$ and $f'$, respectively, all local minima of $\mu$ can be found analytically in $\mathcal{O}(N)$ time in a single sweep through the `cells' $t_{i-1}<t<t_{i}$, $i=1,\dots,N$ (here $t_0=0$ denotes the start location, where $(y_0,y'_0)$ are `inherited' from the preceding line search. For typical line searches $N<10$, c.f.~\textsection\ref{sec:experiments}. In each cell, $\mu(t)$ is a cubic polynomial with at most one minimum in the cell, found by an inexpensive quadratic computation from the three scalars $\mu'(t_i),\mu''(t_i),\mu'''(t_i)$. This is in contrast to other \gp~regression models---for example the one arising from a squared exponential kernel---which give more involved posterior means whose local minima can be found only approximately. Another advantage of the cubic spline interpolant is that it does not assume the existence of higher derivatives (in contrast to the Gaussian kernel, for example), and thus reacts robustly to irregularities in the objective. %  \citep{runge1901empirische}
In our algorithm, after each evaluation of $(y_N,y'_N)$, we use this property to compute a short list of \emph{candidates} for the next evaluation, consisting of the $\leq N$ local minimizers of $\mu(t)$ and one additional \emph{extrapolation} node at $t_{\max}+\alpha$, where $t_{\max}$ is the currently largest evaluated $t$, and $\alpha$ is an extrapolation step size starting at $\alpha=1$ and doubled after each extrapolation step. \footnote{For the integrated Wiener process and heteroscedastic noise, the variance \emph{always} attains its maximum exactly at the mid-point between two evaluations; including the variance into the candidate selection biases the existing candidates towards the center (additional candidates might occur between evaluations without local minimizer, even for noise free observations/classic line searches). We did not explore this further since the algorithm showed very good sample efficiency already with the adopted scheme.}

Another motivation for using the integrated Wiener process as surrogate for the objective, as well as for the described candidate selection, are classic line searches. There, the 1D-objective is modeled by piecewise cubic interpolations between neighboring datapoints. In a sense, this is a non-parametric approach, since a new spline is defined, when a datapoint is added. Classic line searches always only deal with one spline at a time, since they are able to collapse all other parts of the search space. For noise free observations, the mean of the posterior \gp~is identical to the classic cubic interpolations, and thus candidate locations are identical as well; this is illustrated in Figure~\ref{fig:WPSketch}.
The non-parametric approach also prevents issues of over-constrained surrogates for more than two datapoints. For example, unless the objective is a perfect cubic function, it is impossible to fit a parametric third order polynomial to it, for more than two noise free observations. All other variability in the objective would need to be explained away by artificially introducing noise on the observations. An integrated Wiener process very naturally extends its complexity with each newly added datapoint without being overly assertive -- the encoded assumption is, that the objective has \emph{at least} one derivative (which is also observed in this case). 

% =========================================================================
\subsection{Choosing Among Candidates}
\label{sec:select-eval-points}

\begin{figure}
  \centering
  \setlength{\figwidth}{.9\textwidth}
  \setlength{\figheight}{.32\textheight} % .32
 % \tikzset{external/remake next}
%  {\scriptsize \input{poster_figs/EI_Sketch.tikz}}  
  \includegraphics[scale=1.0]{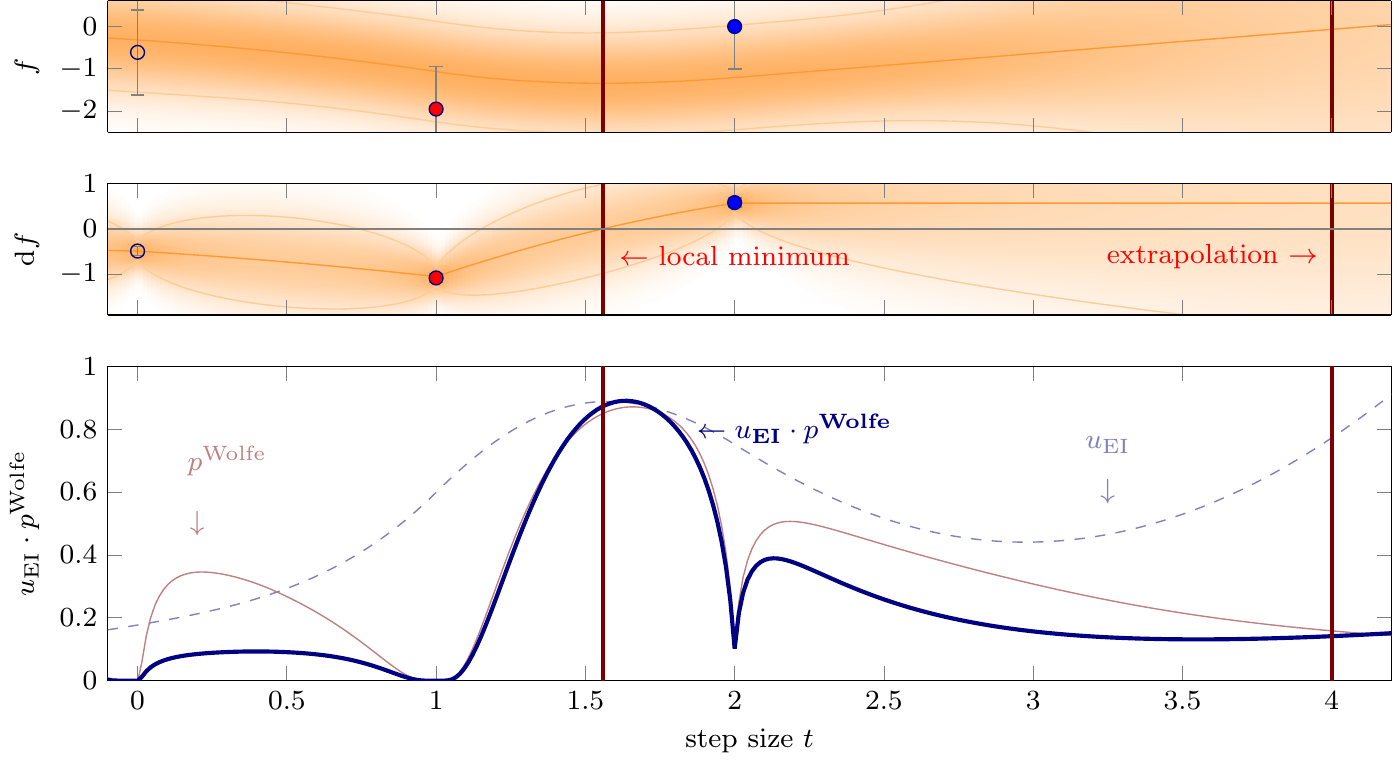}
  \caption{\emph{Candidate selection} by Bayesian optimization. \emph{Top:} \gp~marginal posterior of function values. Posterior mean in solid orange and, two standard deviations in thinner solid orange, local pdf marginal as shading. The red and the blue point are evaluations of the objective function, collected by the line search. \emph{Middle:} \gp~marginal posterior of corresponding gradients. Colors same as in top plot. In all three plots the locations of the \emph{two} candidate points (\textsection\ref{sec:gauss-proc-surr}) are indicated as vertical dark red lines. The left one at about $t^{\text{cand}}_1\approx 1.54$ is a local minimum of the posterior mean in between the red and blue point (the mean of the gradient belief (solid orange, middle plot) crosses through zero here). The right one at $t^{\text{cand}}_2=4$ is a candidate for extrapolation. \emph{Bottom:} Decision criterion in arbitrary scale: The expected improvement $u_{\text{EI}}$ (Eq.~\ref{eq:10}) is shown in dashed light blue, the Wolfe probability $\pW$ (Eq.~\ref{eq:16} and Eq.~\ref{eq:19}) in light red and their decisive product in solid dark blue. For illustrative purposes all criteria are plotted for the whole $t$-space. In practice solely the values at $t^{\text{cand}}_1$ and $t^{\text{cand}}_2$ are computed, compared, and the candidate with the higher value of $u_{\text{EI}}\cdot \pW$ is chosen for evaluation. In this example this would be the candidate at $t^{\text{cand}}_1$.}
\label{fig:EI-sketch}
\end{figure}

\begin{figure}
  \centering
  \setlength{\figwidth}{.9\textwidth}
  \setlength{\figheight}{.32\textheight}
  %\tikzset{external/remake next}
%  {\scriptsize \input{fig/probLS_Sketch_N3_0_EI0.tikz}}  
  \includegraphics[scale=1.0]{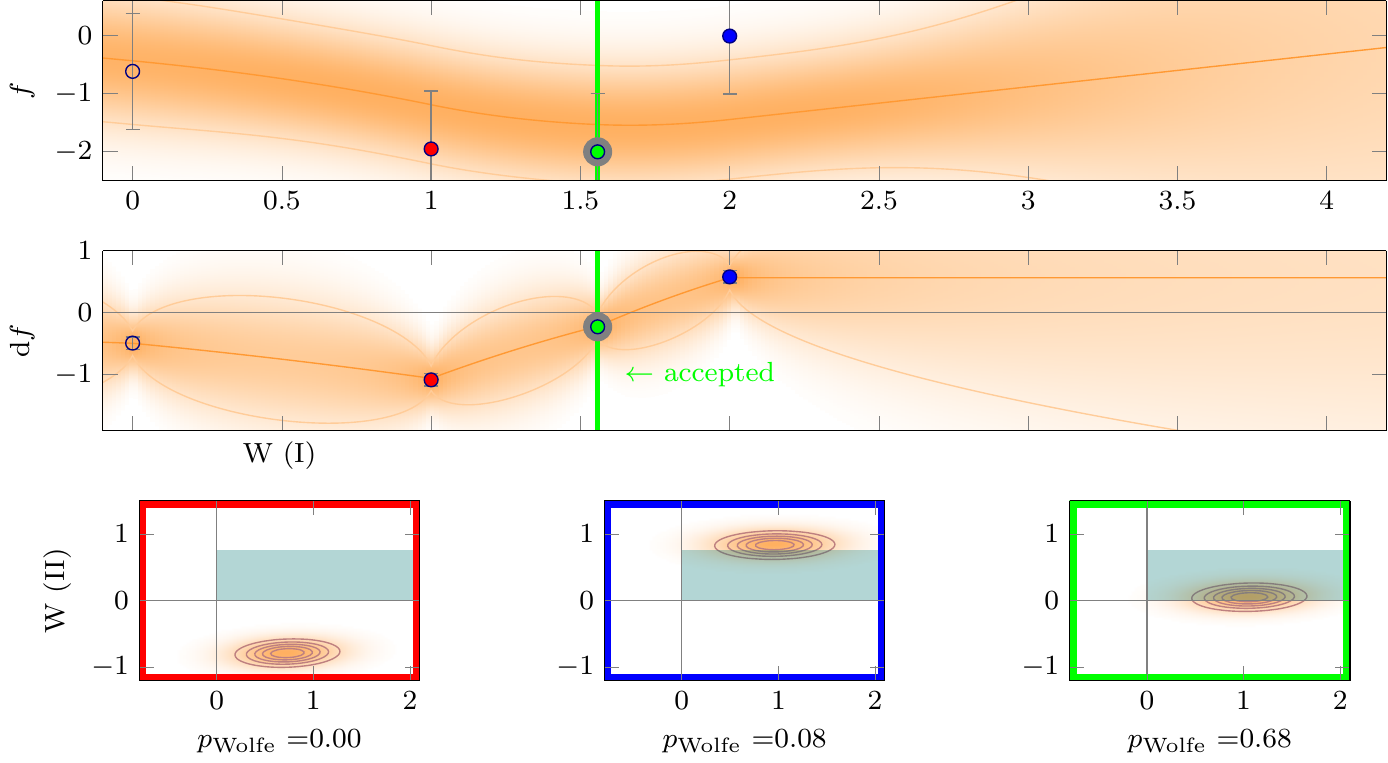}
  \caption{\emph{Acceptance procedure}. \emph{Top and middle:} plot and colors as in Figure~\ref{fig:EI-sketch} with an additional `green' observation.
\emph{Bottom:} Implied bivariate Gaussian belief over the validity of the Wolfe conditions (Eq.~\ref{eq:12}) at the red, blue and green point respectively. Points are considered acceptable if their Wolfe probability $\pW_t$ is above a threshold $c_W=0.3$; this means that at least $30\%$ of the orange 2D Gauss density must cover greenish shaded area. Only the green point fulfills this condition and is therefore accepted.}
\label{fig:prob_ls_accept}
\end{figure}
\begin{figure*}
  \centering
  \setlength{\figwidth}{.14\textwidth}
  \setlength{\figheight}{.1\textheight}
%  {\scriptsize %
  %\tikzset{external/remake next}
  %  \input{fig/snapshot1.tikz} \hspace{-8mm}%
  %\tikzset{external/remake next}
  %  \input{fig/snapshot11.tikz}\hspace{-7mm}%
  %\tikzset{external/remake next}
  %  \input{fig/snapshot3.tikz} \hspace{-8mm}%
  %\tikzset{external/remake next}
  %  \input{fig/snapshot4.tikz} \hspace{-7mm}%
  %\tikzset{external/remake next}
  %  \input{fig/snapshot10.tikz}%
%  }
    \includegraphics[scale=1.0]{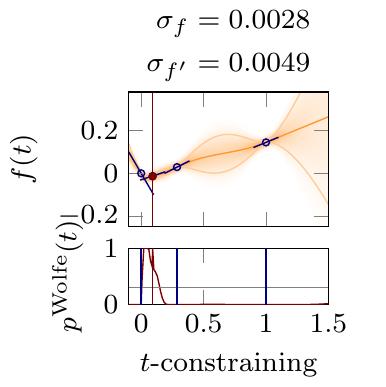} \hspace{-8mm}%
    \includegraphics[scale=1.0]{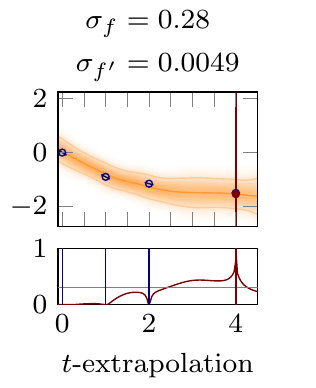}\hspace{-7mm}%
    \includegraphics[scale=1.0]{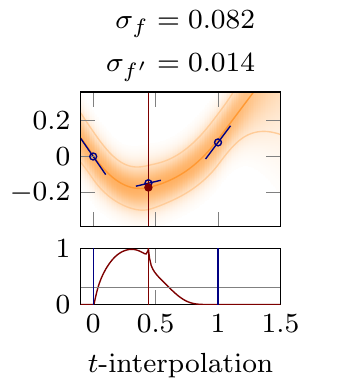}\hspace{-8mm}%
    \includegraphics[scale=1.0]{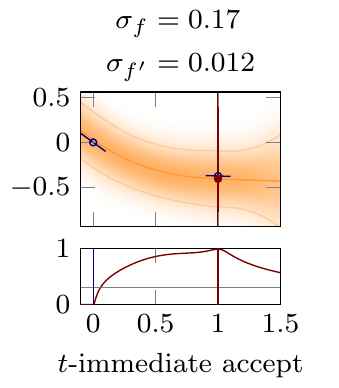}\hspace{-7mm}%
    \includegraphics[scale=1.0]{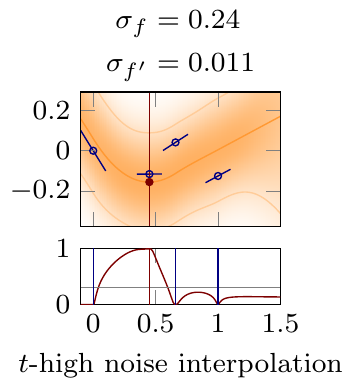}
  \caption{\emph{Curated snapshots} of line searches (from N-I on MNIST), showing variability of the objective's shape and the decision process. \emph{Top row:} \gp~marginal posterior of function values and evaluations, \emph{bottom row:} approximate $\pW$ over strong Wolfe conditions. Accepted point marked red.}
\label{fig:snapshots}
\end{figure*}
The previous section described the construction of $<N+1$ discrete candidate points for the next evaluation. To decide at which of the candidate points to actually call $f$ and $f'$, we make use of a popular acquisition function from Bayesian optimization. \emph{Expected improvement} \citep{jones1998efficient} is the expected amount, under the \gp~surrogate, by which the function $f(t)$ might be smaller than a `current best' value $\eta$ (we set $\eta=\min_{i=0,\dots,N}\{\mu(t_i)\}$, where $t_i$ are observed locations),
\begin{equation}
  \label{eq:10}
  \begin{split}%\raisetag{1.64cm}
    u_{\text{EI}}(t) &= \Exp_{p(f_t\g \vec{y},\vec{y}')}[\min\{0,\eta -
    f(t)\}]\\
    &= \frac{\eta - \mu(t)}{2}\left(1+\erf\frac{\eta-\mu(t)}{\sqrt{2\V(t)}}\right) +
    \sqrt{\frac{\V(t)}{2\pi}}\exp\left(-\frac{(\eta-\mu(t))^2}{2\V(t)} \right).
  \end{split}
\end{equation}
The next evaluation point is chosen as the candidate maximizing the product of Eq.~\ref{eq:10} and Wolfe probability $\pW$, which is derived in the following section. 
The intuition is that $\pW$ precisely encodes properties of desired points, but has poor exploration properties; $u_{\text{EI}}$ has better exploration properties, but lacks the information that we are seeking a point with low curvature;
$u_{\text{EI}}$ thus puts weight on (by W-II) clearly ruled out points.
An illustration of the candidate proposal and selection is shown in Figure \ref{fig:EI-sketch}. 

In principle other acquisition functions (e.g. the upper-confidence bound, \textsc{gp-ucb} \citep{srinivasgaussian}) are possible, which might have a stronger explorative behavior; we opted for $u_{\text{EI}}$ since exploration is less crucial for line searches than for general \textsc{bo} and some (e.g. \textsc{gp-ucb}) had one additional parameter to tune.
We tracked the sample efficiency of $u_{\text{EI}}$ instead and it was very good (low);
the experimental Subsection~\ref{sec:learning-rate-traces} contains further comments and experiments on the alternative choices of $u_{\text{EI}}$ and $\pW$ as standalone acquisition functions; they performed equally well (in terms of loss and sample efficiency) to their product.

% =========================================================================
\subsection{Probabilistic Wolfe Conditions for Termination}
\label{sec:determ-conv}

The key observation for a probabilistic extension of the Wolfe conditions W-I and W-II is that they are positivity constraints on two variables $a_t,b_t$ that are both linear projections of the (jointly Gaussian) variables $f$ and $f'$:
\begin{equation}
  \label{eq:11}
  \begin{bmatrix}
    a_t\\b_t
  \end{bmatrix} = 
  \begin{bmatrix}
    1 & c_1t & -1 & 0\\
    0 & -c_2 & 0 & 1
  \end{bmatrix}
  \begin{bmatrix}
    f(0)\\ f'(0) \\ f(t) \\ f'(t)
  \end{bmatrix} \geq 0.
\end{equation}
The \gp~of Eq.~\eqref{eq:6}~on $f$ thus implies, at each value of $t$, a bivariate Gaussian distribution
\begin{align}
  \label{eq:12}
  p(a_t,b_t) &= \N\left(
    \begin{bmatrix}
      a_t\\b_t
    \end{bmatrix};
    \begin{bmatrix}
      m^a _t\\m^b _t
    \end{bmatrix},
    \begin{bmatrix}
      C^{aa} _{t} & C^{ab} _t\\ C^{ba} _t & C^{bb} _t
    \end{bmatrix}\right),\\
    \label{eq:13}
    \text{with}\qq m^a _t &= \mu(0) - \mu(t) + c_1 t \mu'(0)\notag\\%\qq\text{and}\qq
    m^b _t &= \mu'(t) - c_2 \mu'(0)\\
  \begin{split}
  \label{eq:14}
    \text{and}\qq C^{aa} _t &= \tilde{k}_{00} + (c_1t)^2 \tensor*[^\de]{\tilde{k}}{^\de _{00}}
    + \tilde{k}_{tt}
    + 2[c_1t(\tilde{k}^{\de} _{00} - \tensor[^\de]{\tilde{k}}{_{0t}}) - \tilde{k}_{0t}]\\
    C^{bb} _t &= c_2 ^2 \tensor*[^\de]{\tilde{k}}{^\de _{00}} - 2c_2
    \tensor*[^{\de}]{\tilde{k}}{^{\de}_{0t}} + \tensor*[^{\de}]{\tilde{k}}{^{\de} _{tt}}\\
    C^{ab} _t = C^{ba} _t &= -c_2(\tilde{k}^{\de} _{00} + c_1 t
    \tensor*[^{\de}]{\tilde{k}}{^{\de} _{00}})
    + c_2 \tensor[^{\de}]{\tilde{k}}{_{0t}}+ \tensor[^{\de}]{\tilde{k}}{_{t0}} + c_1t
    \tensor[^{\de}]{\tilde{k}}{^{\de} _{0t}} - \tilde{k}^{\de} _{tt}.
  \end{split}
\end{align}
The quadrant probability $\pW_t=p(a_t>0 \wedge b_t>0)$ for the Wolfe conditions to hold, is an integral over a bivariate normal probability,
\begin{equation}
  \label{eq:16}%\raisetag{1.7cm}
  \pW_t = 
  \int_{-\frac{m^a _t}{\sqrt{C^{aa} _t}}} ^\infty
  \int_{-\frac{m^b _t}{\sqrt{C^{bb} _t}}}
  ^\infty \N\left(
    \begin{bmatrix}
      a\\b
    \end{bmatrix};
    \begin{bmatrix}
      0\\0
    \end{bmatrix}
    ,
    \begin{bmatrix}
      1 & \rho_t \\ \rho_t & 1
    \end{bmatrix}
\right) \,da \,db,
\end{equation}
with correlation coefficient $\rho_t = C^{ab} _t / \sqrt{C^{aa} _t C^{bb} _t}$. It can be computed efficiently \citep{drezner1990computation}, using readily available code.\footnote{e.g.~{\scriptsize\url{http://www.math.wsu.edu/faculty/genz/software/matlab/bvn.m}}} 
The line search computes this probability for all evaluation nodes, after each evaluation. If any of the nodes fulfills the Wolfe conditions with $\pW_t>c_W$, greater than some threshold $0<c_W\leq 1$, it is accepted and returned. If several nodes simultaneously fulfill this requirement, the most recently evaluated node is returned; there are additional safeguards for cases where e.g. no Wolfe-point can be found, which can be deduced from the pseudo-code in Appendix D; they are similar to standard safeguards of classic line search routines (e.g. returning the node of lowest mean). Section~\ref{sec:design-param-c-1} below motivates fixing $c_W=0.3$. The acceptance procedure is illustrated in Figure \ref{fig:prob_ls_accept}.

% =========================================================================
\subsubsection{Approximation for Strong Conditions:}
\label{sec:appr-strong-wolfe}

As noted in Section~\ref{sec:wolfe-cond-conv}, deterministic optimizers tend to use the strong Wolfe conditions, which use $|f'(0)|$ and $|f'(t)|$. A precise extension of these conditions to the probabilistic setting is numerically taxing, because the distribution over $|f'|$ is a non-central $\chi$-distribution, requiring customized computations. However, a straightforward variation to \ref{eq:16} captures the spirit of the strong Wolfe conditions, that large positive derivatives should not be accepted: Assuming $f'(0)<0$ (i.e.~that the search direction is a descent direction), the strong second Wolfe condition can be written exactly as
\begin{equation}
  \label{eq:18}
  0 \leq b_t = f'(t)-c_2f'(0) \leq -2c_2 f'(0).
\end{equation}
The value $-2c_2 f'(0)$ is bounded to $95\%$ confidence by
\begin{equation}
  \label{eq:19}
  -2c_2 f'(0)  \lesssim 2 c_2 (|\mu'(0)| + 2\sqrt{\V'(0)}) =:\bar{b}.
\end{equation}
Hence, an approximation to the strong Wolfe conditions can be reached by replacing the infinite upper integration limit on $b$ in Eq.~\ref{eq:16} with $(\bar{b}-m^b _t)/\sqrt{C^{bb} _t}$. The effect of this adaptation, which adds no overhead to the computation, is shown in Figure~\ref{fig:prob-ls-sketch} as a dashed line.

% =========================================================================
\subsection{Eliminating Hyper-parameters}
\label{sec:hyperp-estim}

As a black-box inner loop, the line search should not require any tuning by the user. The preceding section introduced six so-far undefined parameters: $c_1,c_2,c_W,\theta,\sigma_f,\sigma_{f'}$. We will now show that $c_1,c_2,c_W$, can be fixed by hard design decisions: $\theta$ can be eliminated by standardizing the optimization objective within the line search; and the noise levels can be estimated at runtime with low overhead for finite-sum objectives of the form in Eq.~\ref{eq:1}. The result is a parameter-free algorithm that effectively \emph{removes} the one most problematic parameter from \sgd---the learning rate. 

\subsubsection{Design Parameters \texorpdfstring{$c_1,c_2,c_W$}{c1,c2,cW}}
\label{sec:design-param-c-1}

Our algorithm inherits the Wolfe thresholds $c_1$ and $c_2$ from its deterministic ancestors. We set $c_1=0.05$ and $c_2 = 0.5$. This is a standard setting that yields a `lenient' line search, i.e.~one that accepts most descent points. The rationale is that the stochastic aspect of \sgd~is not always problematic, but can also be helpful through a kind of `annealing' effect.

The acceptance threshold $c_W$ is a new design parameter arising only in the probabilistic setting. We fix it to $c_W=0.3$. To motivate this value, first note that in the noise-free limit, all values $0<c_W<1$ are equivalent, because $\pW$ then switches discretely between 0 and 1 upon observation of the function. A back-of-the-envelope computation, assuming only two evaluations at $t=0$ and $t=t_1$ and the same fixed noise level on $f$ and $f'$ (which then cancels out), shows that function values barely fulfilling the conditions, i.e.~$a_{t_1}=b_{t_1}=0$, can have $\pW\sim 0.2$ while function values at $a_{t_1}=b_{t_1}=-\eps$ for $\eps\to 0$ with `unlucky' evaluations (both function and gradient values one standard-deviation from true value) can achieve $\pW\sim 0.4$. The choice $c_W=0.3$ balances the two competing desiderata for precision and recall. Empirically (Fig.~\ref{fig:snapshots}), we rarely observed values of $\pW$ close to this threshold. Even at high evaluation noise, a function evaluation typically either clearly rules out the Wolfe conditions, or lifts $\pW$ well above the threshold. A more in-depth analysis of $c_1$, $c_2$, and $c_W$ is done in the experimental Section \ref{sec:wolfe-ii-parameter}.

\subsubsection{Scale \texorpdfstring{$\theta$}{theta}}
\label{sec:scale-theta}

The parameter $\theta$ of Eq.~\ref{eq:2} simply scales the prior variance. It can be eliminated by scaling the optimization objective: We set $\theta=1$ and scale
  $y_i \gets \nicefrac{(y_i - y_0)}{|y' _0|}, y_i' \gets \nicefrac{y'_i}{|y' _0|}$ within the code of the line search. This gives $y(0)=0$ and $y'(0)=-1$, and typically ensures the objective ranges in the single digits across $0<t<10$, where most line searches take place. The division by $|y' _0|$ causes a non-Gaussian disturbance, but this does not seem to have notable empirical effect.

\subsubsection{Noise Scales \texorpdfstring{$\sigma_f,\sigma_{f'}$}{sigmaf, sigmadf}}
\label{sec:noise-variances}

The likelihood \ref{eq:3} requires standard deviations for the noise on both function values ($\sigma_f$) and gradients ($\sigma_{f'}$).  One could attempt to learn these across several line searches. However, in exchangeable models, as captured by Eq.~\ref{eq:1}, the variance of the loss and its gradient can be estimated directly for the mini-batch, at low computational overhead---an approach already advocated by \citet{schaul2013no}. We collect the empirical statistics
\begin{equation}
  \label{eq:21}
    \hat{S}(x) := \frac{1}{m}\sum_{j} ^m \ell^2(x,y_j),\q\text{and} \q
    \hat{\nabla S}(x):= \frac{1}{m}\sum_{j} ^m \nabla\ell(x,y_j)^{\Hm 2} 
\end{equation}
(where $^{\Hm 2}$ denotes the element-wise square) and estimate, at the beginning of a line search from $x_k$,
\begin{equation}
  \label{eq:22}
    \sigma_f ^2=\frac{1}{m-1}\left(\hat{S}(x_k) - \hat{\mathcal{L}}(x_k)^2\right)~\text{and}~
    \sigma_{f'} ^2={s_i^{\Hm 2}}\Trans\left[\frac{1}{m-1}\left(\hat{\nabla S}(x_k) - (\nabla\hat{\mathcal{L}}(x_k))^{\Hm 2}\right)\right].
\end{equation}
This amounts to the assumption that noise on the gradient is independent. We finally scale the two empirical estimates as described in Section \textsection\ref{sec:scale-theta}: $\sigma_f \gets \sigma_f / |y'(0)|$, and ditto for $\sigma_{f'}$. The overhead of this estimation is small if the computation of $\ell(x,y_j)$ itself is more expensive than the summation over $j$. In the neural network examples N-I and N-II of the experimental Section \ref{sec:experiments},
the additional steps added only $\sim 1\%$ cost overhead to the evaluation of the loss. A more general statement about memory and time requirements can be found in Sections \ref{sec:requ-time} and \ref{sec:requ-mem}. Of course, this approach requires a mini-batch size $m>1$. For single-sample mini-batches, a running averaging could be used instead (single-sample mini-batches are not necessarily a good choice. In our experiments, for example, vanilla \sgd~with batch size 10 converged faster in wall-clock time than unit-batch \sgd). Estimating noise separately for each input dimension captures the often inhomogeneous structure among gradient elements, and its effect on the noise along the projected direction. For example, in deep models, gradient noise is typically higher on weights between the input and first hidden layer, hence line searches along the corresponding directions are noisier than those along directions affecting higher-level weights. A detailed description of the noise estimator can be found in Appendix \ref{app:noise-estimation}.

% =========================================================================
\subsubsection{Propagating Step Sizes Between Line Searches}
\label{sec:step-size-alpha}

As will be demonstrated in \textsection\ref{sec:experiments}, the line search can find good step sizes even if the length of the direction $s_i$ 
is mis-scaled. Since such scale issues typically persist over time, it would be wasteful to have the algorithm re-fit a good scale in each line search. Instead, we propagate step lengths from one iteration of the search to another: We set the initial search direction to $s_0=-\alpha_0 \nabla \hat{\mathcal{L}} (x_0)$ with some initial learning rate $\alpha_0$. Then, after each line search ending at $x_i=x_{i-1} + t_* s_i$, the next search direction is set to $s_{i+1}=-\alpha_{\text{ext}}\cdot t_* \alpha_0 \nabla \hat{\mathcal{L}} (x_i)$ (with $\alpha_{\text{ext}}=1.3$). Thus, the next line search starts its extrapolation at $1.3$ times the step size of its predecessor (Section \ref{sec:extr-fact-alph} for details).

% =========================================================================
\subsection{Relation to Bayesian Optimization and Noise-Free Limit}
\label{sec:diff-betw-bo}
%\emph{Relation to Bayesian optimization:}
The probabilistic line search algorithm is closely related to Bayesian optimization (\textsc{bo}) since it approximately minimizes a 1D-objective under potentially noisy function evaluations. It thus uses notions of \textsc{bo} (e.g. a \gp-surrogate for the objective, and an acquisition function to discriminate locations for the next evaluation of the loss), but there are some differences concerning the aim, requirements on computational efficiency, and termination condition, which are shortly discussed here:
(i)~\emph{Performance measure:} The final performance measure in \textsc{bo} is usually the lowest found value of the objective function.
Line searches are subroutines inside of a greedy, iterative optimization machine, which usually performs several thousand steps (and line searches);
many, very approximate steps often performs better than taking less, but preciser steps.  
(ii)~\emph{Termination:} The termination condition of a line search is imposed from the outside in the form of the Wolfe conditions. Stricter Wolfe conditions do not usually improve the performance of the overall optimizer, thus, no matter if a better (lower) minimum could be found, any Wolfe-point is acceptable at all times. 
(iii)~\emph{Sample efficiency:} Since the last evaluation from the previous line search can be re-used in the current line search, only one additional value and gradient evaluation is enough to terminate the procedure. This `immediate-accept' is the desired behavior if the learning rate is currently well calibrated.
(iv)~\emph{Locations for evaluation:} \textsc{bo}, usually calls an optimizer to maximize some acquisition function,
and the preciseness of this optimization is crucial for performance. 
Line searches just need to find a Wolfe-acceptable point;
classic line searches suggest, that it is enough to look at plausible locations, like minimizer of a local interpolator, or some rough extrapolation point; 
this inexpensive heuristic usually works rather well.
(v)~\emph{Exploration:} 
\textsc{bo} needs to solve an intricate trade-off problem in between exploring enough of the parameters space for possible locations of minima, and exploiting locations around them further. Since line searches are only concerned with finding a Wolfe-point, they do not need to explore the parameter space of possible step sizes to that extend; crucial features are rather the possibility to explore somewhat larger steps than previous ones (which is done by extrapolation-candidates), and likewise to shorted steps (which is done by interpolation-candidates). 

In the limit of noise free observed gradients and function values ($\sigma_f=\sigma_{f'}=0$) the probabilistic line search behaves like its classic parent, except for very slight variations in the candidate choice (building block 3): The \gp-mean reverts to the classic interpolator; all candidate locations are thus identical, but the probabilistic line search might propose a second option, since (even if there is a local minimizer) it \emph{always} also proposes an extrapolation candidate.
For intuitive purposes, this is illustrated in the following table. 
 \begin{center}
   \renewcommand{\arraystretch}{1.3}
    \begin{tabular}{p{4.5cm} p{4.5cm} p{4.7cm}}
      \textbf{building block} & \textbf{classic} & \textbf{probabilistic (noise free)}\\
      \hline
      \textbf{1)} 1D surrogate for 

      ~~~~objective $f(t)$ 
      & piecewise cubic splines
      & \gp-mean identical to classic interpolator
      \\
%      \hline
      \textbf{2)} candidate selection 
      & local minimizer of cubic splines \emph{xor} extrapolation
      & local minimizer of cubic splines \emph{or} extrapolation
      \\
%      \hline
      \textbf{3)} choice of best candidate 
      & \qqq\q --------- %(only 1 candidate)
      & \textsc{bo} acquisition function 
      \\
%      \hline
      \textbf{4)} acceptance criterion 
      & classic Wolfe conditions
      & $\pW$ identical to classic Wolfe conditions
      \\
    \end{tabular}
\end{center}

\subsection{Computational Time Overhead}
\label{sec:requ-time}
The line search routine itself has little memory and time overhead; most importantly it is independent of the dimensionality of the optimization problem. After every call of the objective function the \gp~(\textsection \ref{sec:gauss-proc-surr}) needs to be updated, which at most is at the cost of inverting a $2N\times 2N$-matrix, where $N$ usually is equal to $1, 2$, or $3$ but never $>10$. In addition, the bivariate normal integral $\pW_t$ of Eq.~\ref{eq:16} needs to be computed at most $N$ times. On a laptop, one evaluation of $\pW_t$ costs about 100 microseconds. For the choice among proposed candidates (\textsection \ref{sec:select-eval-points}), again at most $N$, for each, we need to evaluate $\pW_t$ and $u_{\text{EI}}(t)$ (Eq.~\ref{eq:10}) where the latter comes at the expense of evaluating two error functions. Since all of these computations have a fixed cost (in total some milliseconds on a laptop), the relative overhead becomes less the more expensive the evaluation of $\nabla\hat{\mathcal{L}}(x)$.

The largest overhead actually lies outside of the actual line search routine. In case the noise levels $\sigma_f$ and $\sigma_{f'}$ are not known, we need to estimate them. The approach we took is described in Section \ref{sec:noise-variances} where the variance of $\nabla\hat{\mathcal{L}}$ is estimated using the sample variance of the mini-batch, each time the objective function is called. Since in this formulation the variance estimation is about half as expensive as one backward pass of the net, the time overhead depends on the relative cost of the feed forward and backward passes \citep{2016arXiv161205086B}. If forward and backward pass are the same cost, the most straightforward implementation of the variance estimation would make each function call 1.25 times as expensive.\footnote{It is desirable to decrease this value in the future reusing computation results or by approximation but this is beyond this discussion.} At the same time though, \emph{all} exploratory experiments which very considerably increase the time spend when using \sgd~with a hand tuned learning rate schedule need not be performed anymore. In Section \ref{sec:minibatch} we will also see that \sgd~using the probabilistic line search often needs less function evaluations to converge, which might lead to overall faster convergence in wall clock time than classic \sgd~in a single run. 
%Due to inconsistency on the cluster timing, we could unfortunately not perform these tests.
\subsection{Memory Requirement}
\label{sec:requ-mem}
Vanilla \sgd, at all times, keeps around the current optimization parameters $x\in\Re^{D}$ and the gradient vector $\nabla\hat{\mathcal{L}(x)}\in\Re^{D}$. In addition to this, the probabilistic line search needs to store the estimated gradient variances $\Sigma'(x) = (1-m)^{-1}(\hat{\nabla S(x)}-\nabla\hat{\mathcal{L}}(x)^{\Hm 2})$ (Eq.~\ref{eq:22}) of same size. The memory requirement of \sgd+\probLS~is thus comparable to \AdaGrad~or \Adam. If combined with a search direction other than \sgd~always one additional vector of size $D$ needs to be stored.

\FloatBarrier

% =========================================================================
\section{Experiments}
\label{sec:experiments}
This section reports on an extensive set of experiments to characterise and test the line search. The overall evidence from these tests is that the line search performs well and is relatively insensitive to the choice of its internal hyper-parameters as well the mini-batch size.
We performed experiments on two multi-layer perceptrons N-I and N-II; both were trained on two well known datasets MNIST and CIFAR-10.
\begin{itemize}
\item N-I: fully connected net with $1$ hidden layer and $800$ hidden
  units + biases, and 10 output units, sigmoidal activation functions
  and a cross entropy loss. Structure without biases: 784-800-10. 
  Many authors used similar nets and reported performances.\footnote{\label{note-mnist}\scriptsize\url{http://yann.lecun.com/exdb/mnist/}}
\item N-II: fully connected net with $3$ hidden layers and 10 output units, $\tanh$-activation functions
  and a squared loss. Structure without biases:
  784-1000-500-250-10. Similar nets were also used for example in
  \citet{Martens2010} and \citet{icml2013_sutskever13}.
\item MNIST \citep{lecun1998gradient}: 
multi-class classification task with 10 classes: hand-written digits in gray-scale of size $28\times 28$ (numbers `0' to '9'); training set size 60~000, test set size 10~000. 
\item CIFAR-10 \citep{krizhevsky2009learning}: 
multi-class classification task with 10 classes: color images of natural objects (horse, dog, frog,\dots) of size $32\times 32$; training set size 50~000, test set size 10~000; like other authors, we only used the ``batch 1'' sub-set of CIFAR-10 containing 10~000 training examples.
\end{itemize}
In addition we train logistic regressors with sigmoidal output (N-III) on the following binary classification tasks:
\begin{itemize}
   \item Wisconsin Breast Cancer Dataset 
(WDBC) \citep{citeulike:8632439}: binary classification of tumors as either `malignant' or `benign'. The set consist of 569 examples of which we used 169 to monitor generalization performing; thus 400 remain for the training set; 30 features describe for example radius, area, symmetry, et cetera. In comparison to the other datasets and networks, this yields a very low dimensional optimization problem with only 30 (+1 bias) input parameters as well as just a small number of datapoints.
\item GISETTE \citep{NIPS2004_2728}: binary classification of the handwritten digits `4' and `9'. The original $28\times 28$ images are taken from the MNIST datset; then the feature set was expanded and consists of the original normalized pixels, plus a randomly selected subset of products of pairs of features, which are slightly biased towards the upper part of the image; in total there are 5000 features, instead of 784 as in the original MNIST. The size of the training set and test set is 6000 and 1000 respectively.
\item EPSILON: synthetic dataset from the PASCAL Challenge 2008 for binary classification. It consists of 400~000 training set datapoint and 100~000 test set datapoints, each having 2000 features.
\end{itemize}
In the text and figures, \sgd~using the probabilistic line search will occasionally be denoted as \sgd+\probLS. Section \ref{sec:minibatch} contains experiments on the sensitivity to varying gradient noise levels (mini-batch sizes) performed on both multi-layer perceptrons N-I and N-II, as well as on the logistic regressor N-III. Section \ref{sec:sens-design-param} discusses sensitivity to the hyper-parameters choices introduced in Section \ref{sec:hyperp-estim} and Section \ref{sec:learning-rate-traces} contains additional diagnostics on step size statistics. Each single experiment was performed $10$ times with different random seeds that determined the starting weights and the mini-batch selection and seeds were shared across all experiments. We report all results of the $10$ instances as well as means and standard deviations.

\subsection{Varying Mini-batch Sizes}
\label{sec:minibatch}

\begin{figure}%[ht]
  \centering
  \setlength{\figwidth}{.9\textwidth}
  \setlength{\figheight}{.12\textheight}
%  {\scriptsize 
%    %\tikzset{external/remake next}
%    \input{fig/MNIST_ARCH_1000_500_250_10_NEWPARAS_noiseLevel_ter_vs_alpha.tikz}\\
%    %\tikzset{external/remake next}
%    \input{fig/MNIST_ARCH_1000_500_250_10_NEWPARAS_noiseLevel_trainter_vs_epoch_ALL.tikz}\\
%    %\tikzset{external/remake next}
%    \input{fig/MNIST_ARCH_1000_500_250_10_NEWPARAS_noiseLevel_testter_vs_epoch_ALL.tikz}}
%%    %\tikzset{external/remake next}
%%    \input{fig/MNIST_ARCH_1000_500_250_10_NEWPARAS_noiseLevel_ter_vs_epoch.tikz}
%}
    \includegraphics[scale=1.0]{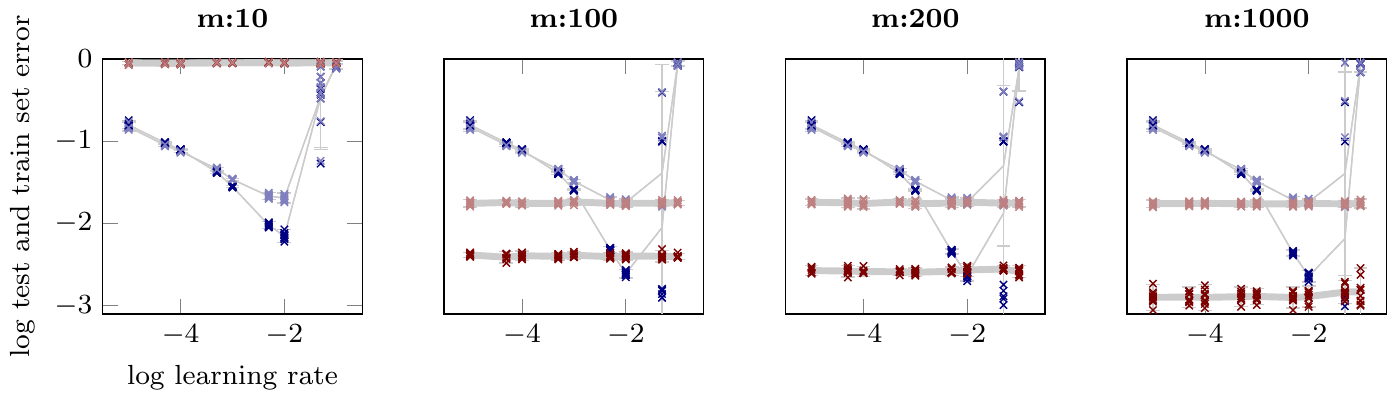}\\
    \includegraphics[scale=1.0]{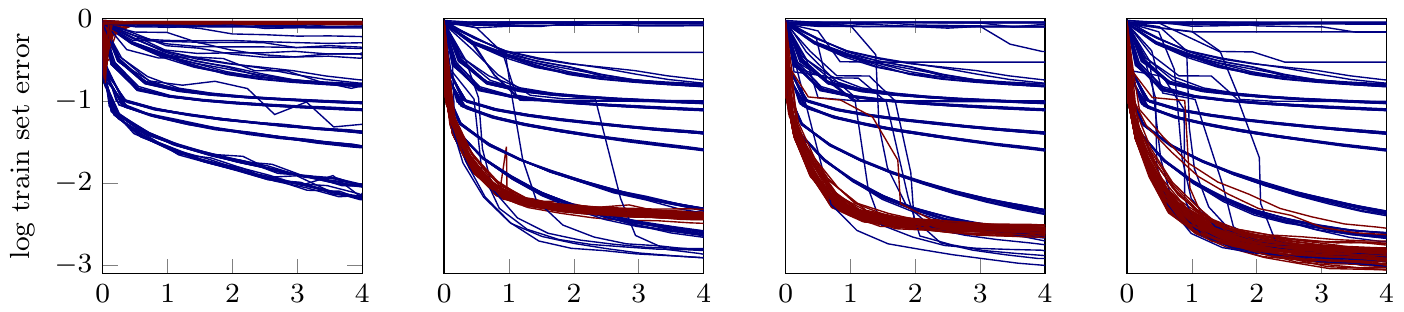}\\
    \includegraphics[scale=1.0]{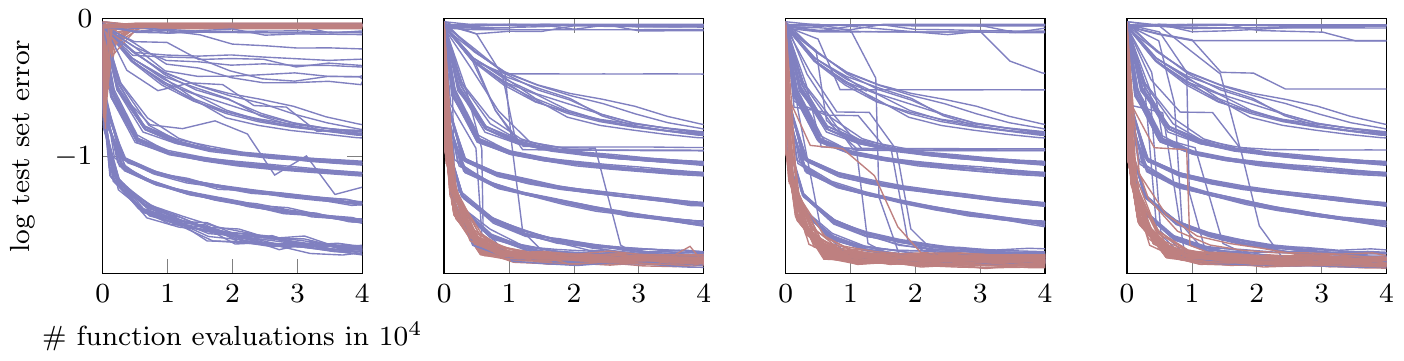}
  \caption{Performance of N-II on MNIST for \emph{varying mini-batch sizes}. \emph{Top:} final logarithmic test set and train set error  after $40~000$ function evaluations of training versus a large range of learning rates each for $10$ different initializations. \sgd-runs with fixed learning rates are shown in light blue (test set) and dark blue (train set); \sgd+\probLS-runs in light red (test set) and dark red (train set); means and two standard deviations for each of the $10$ runs in gray. \emph{Columns} from left to right refer to different mini-batch sizes $m$ of $10$, $100$, $200$ and $1000$ which correspond to decreasing relative noise in the gradient observations. 
Not surprisingly the performance of \sgd-runs with a fixed step size are very sensitive to the choice of this step size. \sgd~using the probabilistic line search adapts initially mis-scaled step sizes and performs well across the whole range of initial learning rates. 
\emph{Middle and bottom:} Evolution of the logarithmic test and train set error respectively for all \sgd-runs and \sgd+\probLS-runs versus \# function evaluations (colors as in top plot). For mini-batch sizes of $m= 100, 200$ and $ 1000$ \emph{all} instances of \sgd~using the probabilistic line search reach the same best test set error. Similarly a good train set error is reached very fast by \sgd+\probLS. Only very few instances of \sgd~with a fixed learning rate reach a better train set error (and this advantage usually does not translate to test set error). For very small mini-batch sizes ($m=10$ and first column in the plot) the line search becomes unstable with this architecture, possibly because of the variance estimation becoming too inaccurate (see Appendix \ref{app:noise-estimation}).}
\label{fig:noiseLevel}
\end{figure}
The noise level of the gradient estimate $\nabla\hat{\mathcal{L}}(x)$ and the loss $\hat{\mathcal{L}}(x)$ is determined by the mini-batch size $m$ and ultimately there should exist an optimal $m$ that maximizes the optimizer's performance in wall-clock-time. 
In practice of course the cost of computing $\nabla\hat{\mathcal{L}}(x)$ and $\hat{\mathcal{L}}(x)$ is not necessarily linear in $m$ since it is upper bounded by the memory capacity of the hardware used. We assume here, that the mini-batch size is chosen by the user;
thus we test the line search with the default hyper-parameter setting (see Sections \ref{sec:hyperp-estim} and \ref{sec:sens-design-param}) on four different mini-batch sizes:
\begin{itemize}
  \item $m= 10, 100, 200$ and $1000$ (for MNIST, CIFAR-10, and EPSILON)
  \item $m =10, 50, 100$, and $400 $ (for WDBC and GISETTE)
\end{itemize}
which correspond to increasing signal-to-noise ratios. Since the training set of WDBC only consists of 400 datapoints, the run with the larges mini-batch size of 400 in fact runs full-batch gradient descent on WDBC; this is not a problem, since---as discussed above---the probabilistic line search can also handle noise free observations.\footnote{Since the dataset size $M$ of WDBC is very small, we used the factor $\nicefrac{(M-m)}{(mM)}$ instead of $\nicefrac{1}{m}$ to scale the sample variances of Eq.~\ref{eq:21}; for $m\ll M$ both factors are nearly identical. The former measures the noise level relative to the empirical risk, the latter relative to the risk; so both choices are sensible depending on what is the desired objective.}
We compare to \sgd-runs using a fixed step size (which is typical for these architectures) and an annealed step size with annealing schedule $\alpha_t = \nicefrac{\alpha_0}{t}$. Because annealed step sizes performed much worse than \sgd+fixed step size, we will only report on the latter results in the plots.\footnote{An example of annealed step size performance can be found in \citet{MahHen2015}.} Since classic \sgd~without the line search needs a hand crafted learning rate we search on exhaustive logarithmic grids of 
\begin{equation*}
  \begin{split}
 \alpha_{\text{\sgd}}^{\text{N-I}} &= [ 10^{-5}, 5\cdot 10^{-5},  10^{-4}, 5\cdot 10^{-4},  10^{-3}, 5\cdot 10^{-3},  10^{-2}, 5\cdot 10^{-2},  10^{-1}, 5\cdot 10^{-1}] \\
 \alpha_{\text{\sgd}}^{\text{N-II}} &= [\alpha_{\text{\sgd}}^{\text{N-I}}, ~1.0, ~1.5, ~2.0, ~2.5, ~3.0, ~3.5, ~4.0] \\
 \alpha_{\text{\sgd}}^{\text{N-III}} &= [ 10^{-8},  10^{-7},  10^{-6},  10^{-5},  10^{-4},  10^{-3},  10^{-2},  10^{-1},  10^{0},  10^{1}, 10^{2}].    
  \end{split}
\end{equation*}
We run $10$ different initialization for each learning rate, each mini-batch size and each net and dataset combination ($10\cdot 4\cdot (2\cdot 10+2\cdot 17 + 3\cdot 11) = 3480 $ runs in total) for a large enough budget to reach convergence; and report all numbers. Then we perform the same experiments using the same seeds and setups with \sgd~using the probabilistic line search and compare the results. For \sgd+\probLS, $\alpha_{\text{\sgd}}$ is the initial learning rate which is used in the very first step. After that, the line search automatically adapts the learning rate, and shows no significant sensitivity to its initialization. 

Results of N-I and N-II on both, MNIST and CIFAR-10 are shown in Figures \ref{fig:noiseLevel}, \ref{fig:noiseMNIST-NI}, \ref{fig:noiseCIFAR-NII}, and \ref{fig:noiseCIFAR-NI}; results of N-III on WDBC, GISETTE and EPSILON are shown in Figures \ref{fig:noiseWBCD-NIII}, \ref{fig:noiseGISETTE-NIII}, and \ref{fig:noiseEPSILON-NIII} respectively. All instances (\sgd~and \sgd+\probLS) get the same computational budget (number of mini-batch evaluations) and not the same number of optimization steps. The latter would favour the probabilistic line search since, on average, a bit more than one mini-batch is evaluated per step. 
Likewise, all plots show performance measure versus the number of mini-batch evaluations, which is proportional to the computational cost.

All plots show similar results: While classic \sgd~is sensitive to the learning rate choice, the line search-controlled \sgd~performs as good, close to, or sometimes even better than the (in practice unknown) optimal classic \sgd~instance. In Figure \ref{fig:noiseLevel}, for example, \sgd+\probLS~converges much faster to a good test set error than the best classic \sgd~instance. In all experiments, across a reasonable range of mini-batch sizes $m$ and of initial $\alpha_{\text{\sgd}}$ values, the line search quickly identified good step sizes $\alpha_t$, stabilized the training, and progressed efficiently, reaching test set errors similar to those reported in the literature for tuned versions of these kind of architectures and datasets. The probabilistic line search thus effectively removes the need for exploratory experiments and learning-rate tuning.

Overfitting and training error curves: 
The training error of \sgd+\probLS~often plateaus earlier than the one of vanilla \sgd, especially for smaller mini-batch sizes. This does not seem to impair the performance of the optimizer on the test set. 
We did not investigate this further, since it seemed like a nice natural annealing effect; the exact causes are unclear for now.
One explanation might be that the line search does indeed improve overfitting, 
since it tries to measure descent (by Wolfe conditions which rely on the \emph{noise-informed} \gp). 
This means, that if---close to a minimum---successive acceptance decisions can not identify a descent direction anymore,
diffusion might set in.

%\FloatBarrier

\subsection{Sensitivity to Design Parameters}
\label{sec:sens-design-param}

% ------------------
\begin{figure}%[ht]
  \centering
  \setlength{\figwidth}{.9\textwidth}
  \setlength{\figheight}{.07\textheight}
  %\tikzset{external/remake next}
%  {\scriptsize \input{fig/MNIST_ARCH_1000_500_250_10_m00200_paraSens_theta_reset.tikz}}  
    \includegraphics[scale=1.0]{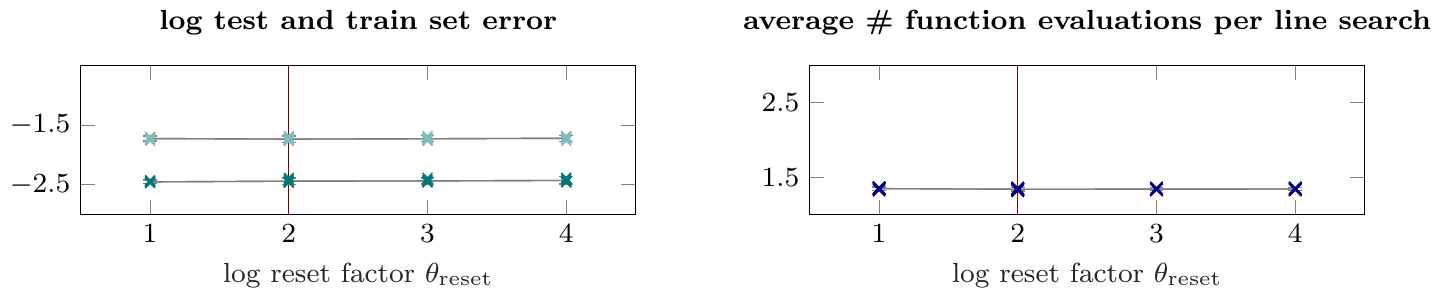}
  \caption{\emph{Sensitivity to varying hyper-parameters} $\theta_{\text{reset}}$. Plot and color coding as in Figure \ref{fig:c2extspace_cwall}. Adopted parameter in dark red at $\theta_{\text{reset}} = 100$. Resetting the \gp~scale occurs very rarely. For example for $\theta_{\text{reset}} = 100$ the reset occurred in $0.02\%$ of all line searches. }
\label{fig:theta-reset}
\end{figure}
% --------------------------------------------
\begin{figure}%[ht]
  \centering
  \setlength{\figwidth}{.9\textwidth}
  \setlength{\figheight}{.4\textheight}
  %\tikzset{external/remake next}
%  {\scriptsize \input{fig/MNIST_ARCH_1000_500_250_10_m00200_paraSens_c2_extall_cWFIX03.tikz}}  
    \includegraphics[scale=1.0]{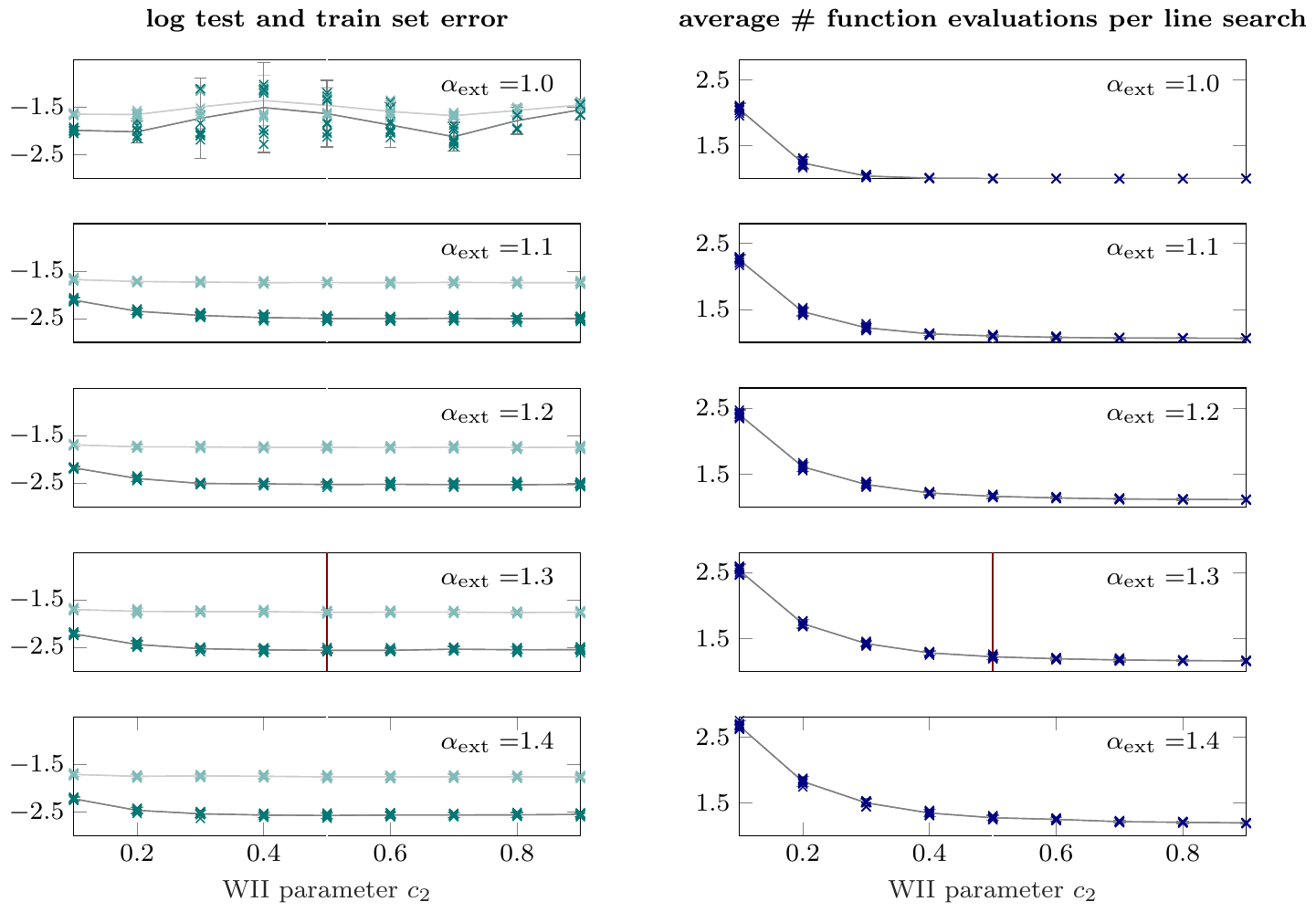}
  \caption{\emph{Sensitivity to varying hyper-parameters} $c_2$, and $\alpha_{\text{ext}}$ (\textsection \ref{sec:hyperp-estim}, \textsection \ref{sec:sens-design-param}). Runs were performed training N-II on MNIST with mini-batch size $m=200$.
For each parameter setting 10 runs with different initializations were performed.
\emph{Left column:} logarithmic test set error (light green) and train set error (dark green) after $40~000$ function evaluations; mean and $\pm $ two standard deviations of the 10 runs in gray. 
\emph{Right Column:} average number of function evaluations per line search. A low number indicates an efficient line search procedure (perfect efficiency at $1$). For most parameter combinations this lies around $\approx 1.3-1.5$. 
Only at extreme parameter values, for example $\alpha_{\text{ext}} = 1.0$, which amounts to no extrapolation at all in between successive line searches, the line search becomes unstable. The hyper-parameters adopted in the line search implementation are indicated as vertical dark red line at $\alpha_{\text{ext}} = 1.3$ and $c_2 = 0.5$.
}
\label{fig:c2extspace_cwall}
\end{figure}
% ------------------
\begin{figure}%[ht]
  \centering
  \setlength{\figwidth}{.9\textwidth}
  \setlength{\figheight}{.75\textheight}
  %\tikzset{external/remake next}
%  {\scriptsize \input{fig/MNIST_ARCH_1000_500_250_10_m00200_paraSens_c2_cwall_extFIX13.tikz}}  
    \includegraphics[scale=1.0]{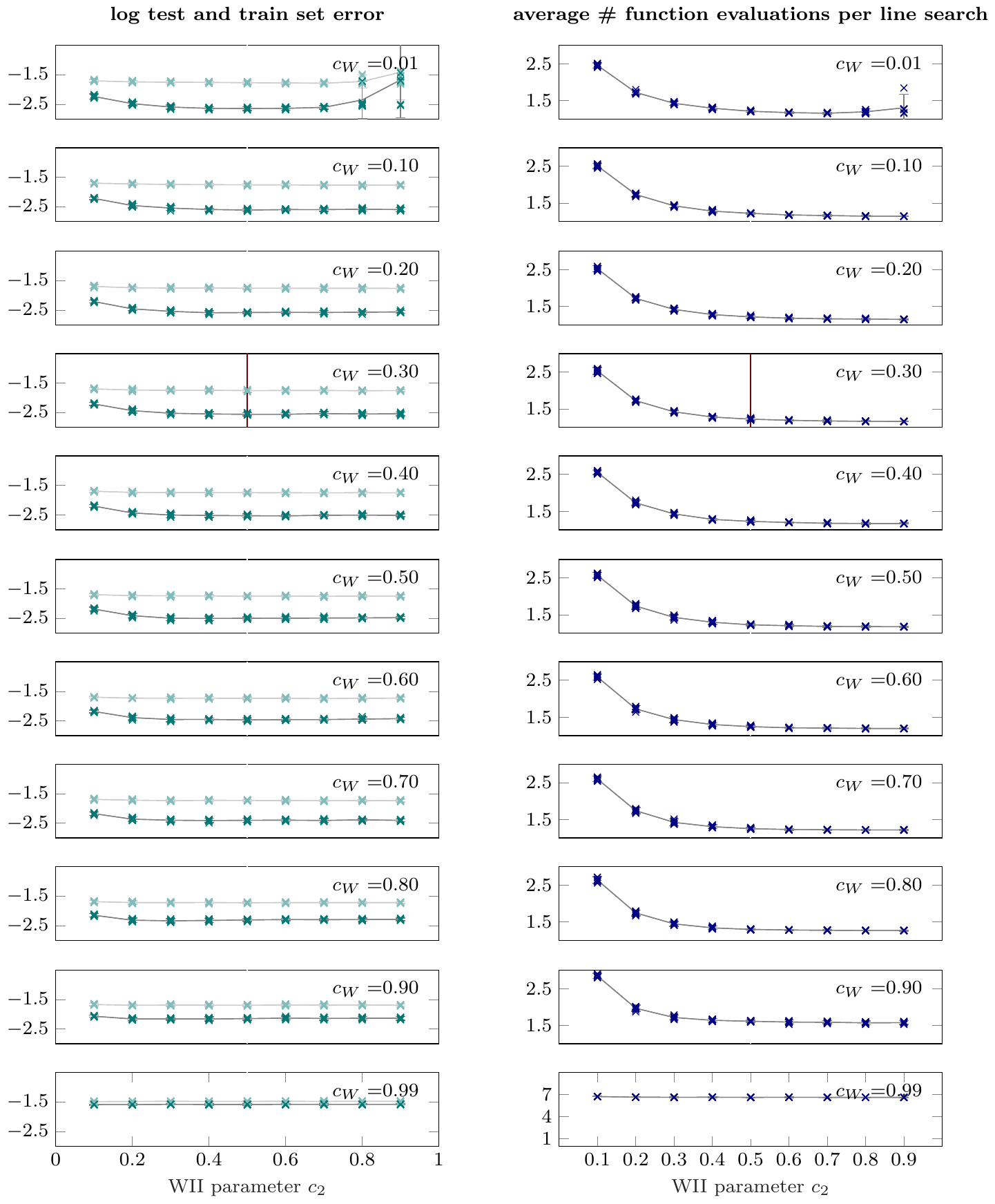}
  \caption{\emph{Sensitivity to varying hyper-parameters} $c_2$, and $c_W$ (\textsection \ref{sec:hyperp-estim}). Plot and color coding as in Figure \ref{fig:c2extspace_cwall} but this time for varying $c_W$ instead of $\alpha_{\text{ext}}$.  
\emph{Right Column:} Again a low number indicates an efficient line search procedure (perfect efficiency at $1$). For most parameter combinations this lies around $\approx 1.3-1.5$. Only at extreme parameter values for example $c_W = 0.99$, which amounts to imposing nearly absolute certainty about the Wolfe conditions, the line search becomes less efficient, though still does not break. Adopted parameters again in dark red at $c_W = 0.3$ and $c_2 = 0.5$}
\label{fig:cwc2space_extall}
\end{figure}
Most, if not all, numerical methods make implicit or explicit choices about their hyper-parameters. Most of these are never seen by the user since they are either estimated at run time, or set by design to a fixed, approximately insensitive value. Well known examples are the discount factor in ordinary differential equation solvers \citep[\textsection 2.4]{hairer87:_solvin_ordin_differ_equat_i}, or the Wolfe parameters $c_1$ and $c_2$ of classic line searches (\textsection \ref{sec:design-param-c-1}). The probabilistic line search inherits the Wolfe parameters $c_1$ and $c_2$ from its classical counterpart as well as introducing two more: The Wolfe threshold $c_W$ and the extrapolation factor $\alpha_{\text{ext}}$. $c_W$ does not appear in the classical formulation since the objective function can be evaluated exactly and the Wolfe probability is binary (either fulfilled or not). While $c_W$ is thus a natural consequence of allowing the line search to model noise explicitly, the extrapolation factor $\alpha_{\text{ext}}$ is the result of the line search favoring shorter steps, which we will discuss below in more detail, but most prominently because of bias in the line search's first gradient observation.

In the following sections we will give an intuition about the task of the most influential design parameters $c_2$, $c_W$, and $\alpha_{\text{ext}}$, discuss how they affect the probabilistic line search, and validate good design choices through exploring the parameter space and showing insensitivity to most of them.
All experiments on hyper-parameter sensitivity were performed training N-II on MNIST with mini-batch size $m=200$. 
For a full search of the parameter space $c_W$-$c_2$-$\alpha_{\text{ext}}$ we performed $4950$ runs in total with $495$ different parameter combinations. All results are reported.

\subsubsection{Wolfe II Parameter $c_2$ and Wolfe Threshold $c_W$}
\label{sec:wolfe-ii-parameter}
 As described in Section \ref{sec:hyperp-estim}, $c_2$ encodes the strictness of the curvature condition W-II. Pictorially speaking, a larger $c_2$ extends the range of acceptable gradients (green shaded are in the lower part of Figure \ref{fig:prob_ls_accept}) and leads to a lenient line search while a smaller value of $c_2$ shrinks this area, leading to a stricter line search. $c_W$ controls how certain we want to be, that the Wolfe conditions are actually fulfilled. In the extreme case of complete uncertainty about the collected gradients and function values ($\sigma_f, \sigma_{f'}\rightarrow \infty$) $\pW $ will always be $<0.25$, if the strong Wolfe conditions are imposed. In the limit of certain observations ($\sigma_f, \sigma_{f'}\rightarrow 0$) $\pW$ is binary and reverts to the classic Wolfe criteria.
An overly strict line search, therefore (e.g. $c_W = 0.99$ and/ or $c_2 = 0.1$), will still be able to optimize the objective function well, but will waste evaluations at the expense of efficiency. 
Figure \ref{fig:cwc2space_extall} explores the $c_2$-$c_W$ parameter space (while keeping $\alpha_{\text{ext}}$ fixed at 1.3). The left column shows final test and train set error, the right column the average number of function evaluations per line search, both versus different choices of Wolfe parameter $c_2$. The left column thus shows the overall performance of the optimizer, while the right column is representative for the computational efficiency of the line search. Intuitively, a line search which is minimally invasive (only corrects the learning rate, when it is really necessary) is preferred. Rows in Figure  \ref{fig:cwc2space_extall} show the same plot for different choices of the Wolfe threshold $c_W$.

The effect of strict $c_2$ can be observed clearly in Figure \ref{fig:cwc2space_extall} where for smaller values of $c_2<\approx 0.2$ the average number of function evaluations spend in one line search goes up slightly in comparison to looser restrictions on $c_2$, while still a very good perfomace is reached in terms of train and test set error. Likewise, the last row of Figure \ref{fig:cwc2space_extall} for the extreme value of $c_W = 0.99$ (demanding $99\%$ certainty about the validity if the Wolfe conditions), shows significant loss in computational efficiency having an average number of $7$ function evaluations per line search, but still does not break. Lowering this threshold a bit to $90\%$ increases the computational efficiency of the line search to be nearly optimal again.

Ideally, we want to trade off the desiderata of being strict enough to reject too small and too large steps that prevent the optimizer to converge, but being lenient enough to allow all other reasonable steps, thus increasing computational efficiency. The values $c_W = 0.3$ and $c_2 = 0.5$, which are adopted in our current implementation are marked as dark red vertical lines in Figure \ref{fig:cwc2space_extall}. 

\subsubsection{Extrapolation Factor $\alpha_{\text{ext}}$}
\label{sec:extr-fact-alph}
The extrapolation parameter $\alpha_{\text{ext}}$, introduced in Section \ref{sec:step-size-alpha}, pushes the line search to try a larger learning rate first, than the one which was accepted in the previous step.
Figure \ref{fig:c2extspace_cwall} is structured like Figure \ref{fig:cwc2space_extall}, but this time explores the line search sensitivity in the $c_2$-$\alpha_{\text{ext}}$ parameter space (abscissa and rows respectively) while keeping $c_W$ fixed at $0.3$. Unless we choose $\alpha_{\text{ext}}= 1.0$ (no step size increase between steps) in combination with a lenient choice of $c_2$ the line search performs well. For now we adopt $\alpha_{\text{ext}}= 1.3$ as default value which again is shown as dark red vertical line in Figure \ref{fig:c2extspace_cwall}.

The introduction of $\alpha_{\text{ext}}$ is a necessity and well-working fix because of a few shortcomings of the current design. First, the curvature condition W-II is the single condition that prevents too small steps and pushes optimization progress. On the other hand both W-I and W-II simultaneously penalize too large steps (see Figure \ref{fig:minimize-sketch} for a sketch).
This is not a problem in case of deterministic observation ($\sigma_f, \sigma_{f'}\rightarrow 0$), where W-II undoubtedly decides if a gradient is still too negative.
Unless W-II is chosen very tightly (small $c_2$) or $c_W$ unnecessarily large (both choices, as discussed above, are undesirable), in the presence of noise, $\pW$ will thus be more reliable in preventing overshooting than pushing progress. 
The first row of Figure \ref{fig:c2extspace_cwall} illustrates this behavior, where the performance drops somewhat if no extrapolation is done ($\alpha_{\text{ext}}=1.0$) in combination with a looser version of W-II (larger $c_2$). 

Another factor that contributes towards accepting small rather than larger learning rates is a bias introduced in the first observation of the line search at $t=0$. Observations $y'(t)$ that the \gp~gets to see are projections of the gradient sample $\nabla \hat{\mathcal{L}}(t)$ onto the search direction $s = -\nabla \hat{\mathcal{L}}(0)$. Since the first observations $y'(0)$ is computed from the same mini-batch as the search direction (not doing this would double the optimizer's computational cost) an inevitable bias is introduced of approximate size of $\cos^{-1}(\gamma)$ (where $\gamma$ is the expected angle between gradient evaluations from two independent mini-batches at $t=0$). 
Since the scale parameter $\theta$ of the Wiener process is implicitly set by $y'(0)$ (\textsection \ref{sec:scale-theta}), the \gp~becomes more uncertain at unobserved points than it needs to be; or alternatively expects the 1D-gradient to cross zero at smaller steps, and thus underestimates a potential learning rate. 
The posterior \emph{at} observed positions is little affected. The over-estimation of $\theta$ rather pushes the posterior towards the likelihood (since there is less model to trust) and thus still gives a reliable measure for $f(t)$ and $f'(t)$.
The effect on the Wolfe conditions is similar. With $y'(0)$ biased towards larger values, the Wolfe conditions, which measure the drop in projected gradient norm, are thus prone to accept larger gradients combined with smaller function values, which again is met by making small steps. Ultimately though, since candidate points at $t^{\text{cand}}>0$ that are currently queried for acceptance, are always observed and unbiased, this can be controlled by an appropriate design of the Wolfe factor $c_2$ (\textsection \ref{sec:design-param-c-1} and \textsection \ref{sec:wolfe-ii-parameter}) and of course $\alpha_{\text{ext}}$.

\subsubsection{Full Hyper-Parameter Search: $c_W$-$c_2$-$\alpha_{\text{ext}}$}
\label{sec:full-hyper-parameter}
An exhaustive performance evaluation on the whole $c_W$-$c_2$-$\alpha_{\text{ext}}$-grid is shown in Appendix \ref{app:para-sens} in Figures \ref{fig:c2cwspace10}-\ref{fig:c2cwspace14} and Figures \ref{fig:c2extspace01}-\ref{fig:c2extspace99}. As discussed above, it shows the necessity of introducing the extrapolation parameter $\alpha_{\text{ext}}$ and shows slightly less efficient performance for obviously undesirable parameter combinations. In a large volume of the parameter space, and most importantly in the vicinity of the chosen design parameters, the line search performance is stable and comparable to carefully hand tuned learning rates.

\subsubsection{Safeguarding Mis-scaled \gp s: $\theta_{\text{reset}}$}
\label{sec:safeg-mis-scal}
For completeness, an additional experiment was performed on the threshold parameter which is denoted by $\theta_{\text{reset}}$ in the pseudo-code (Appendix \ref{app:pseudocode}) and safeguards against \gp~mis-scaling. 
The introduction of noisy observations necessitates to model the variability of the 1D-function, which is described by the kernel scale parameter $\theta$ (\textsection\ref{sec:scale-theta}). Setting this hyper-parameter is implicitly done by scaling the observation input, assuming a similar scale than in the previous line search (\textsection \ref{sec:scale-theta}) . If, for some reason, the previous line search accepted an unexpectedly large or small step (what this means is encoded in $\theta_{\text{reset}}$) the \gp~scale $\theta$ for the next line search is reset to an exponential running average of previous scales ($\alpha_{\text{stats}}$ in the pseudo-code). This occurs very rarely (for the default value $\theta_{\text{reset}} = 100$ the reset occurred in $0.02\%$ of all line searches), but is necessary to safeguard against extremely mis-scaled \gp's. $\theta_{\text{reset}}$ therefore is not part of the probabilistic line search model as such, but prevents mis-scaled \gp s due to some unlucky observation or sudden extreme change in the learning rate. Figure \ref{fig:theta-reset} shows performance of the line search for $\theta_{\text{reset}} = 10, 100, 1000$ and $10~000$ showing no significant performance change. We adopted $\theta_{\text{reset}} = 100$ in our implementation since this is the expected and desired multiplicative (inverse) factor to maximally vary the learning rate in one single step. 

\FloatBarrier
% =========================================================================
\subsection{Candidate Selection and Learning Rate Traces}
\label{sec:learning-rate-traces}
\begin{figure}%[ht]
  \centering
  \setlength{\figwidth}{.9\textwidth}
  \setlength{\figheight}{.4\textheight}
  %\tikzset{external/remake next}
%  {\scriptsize \input{fig/MNIST_ARCH_1000_500_250_10_NEWPARAS_alphaTraces_ext1.30.tikz}}  
    \includegraphics[scale=1.0]{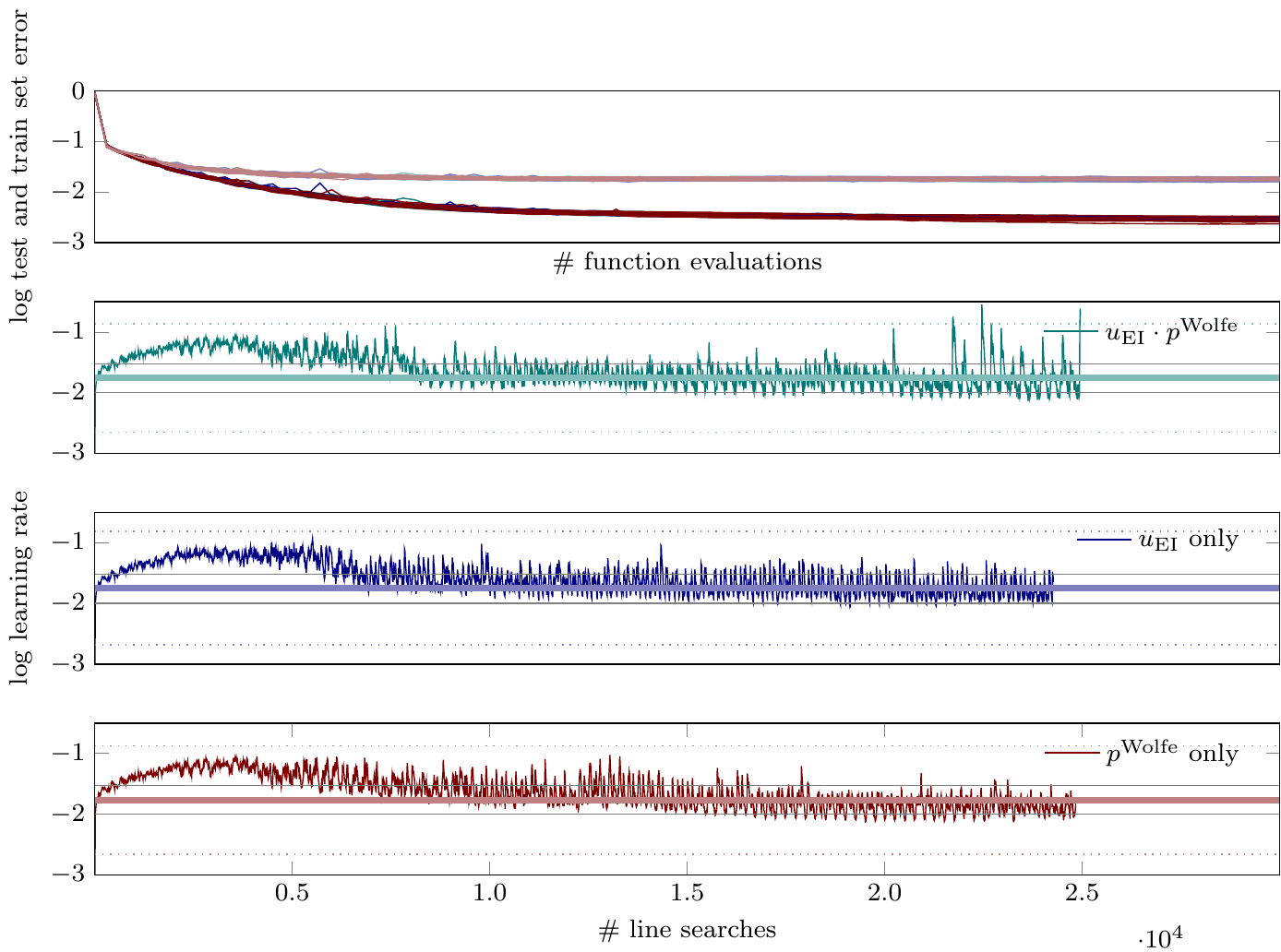}
  \caption{\emph{Different choices of acquisition function} (\textsection \ref{sec:select-eval-points}). We compare between using expected improvement $u_{\text{EI}}$ (blue), the Wolfe probability $\pW$ (red) and their product  $u_{\text{EI}}\cdot \pW$ (green) which is the default in our code. 
\emph{Top:} evolution of the logarithmic test and train set error. 
Different lines of the same color correspond to different seeds.
\emph{Rows 2-4:} show learning rate traces of a single seed (colors same as in top plot). For plotting purposes the curves were smoothed and thinned out. The thick light green, light red and light blue horizontal lines show the mean of the raw (non-smoothed) values of accepted learning rates across the whole optimization process, the dotted lines show $\pm$ two standard deviations and the gray solid lines mark a range of well performing constant learning rates. 
}
\label{fig:cand-selection-exp}
\end{figure}
\begin{figure}%[ht]
  \centering
  \setlength{\figwidth}{.9\textwidth}
  \setlength{\figheight}{.3\textheight}
  %\tikzset{external/remake next}
%  {\scriptsize \input{fig/MNIST_ARCH_1000_500_250_10_NEWPARAS_alphaTraces_noise.tikz}}  
    \includegraphics[scale=1.0]{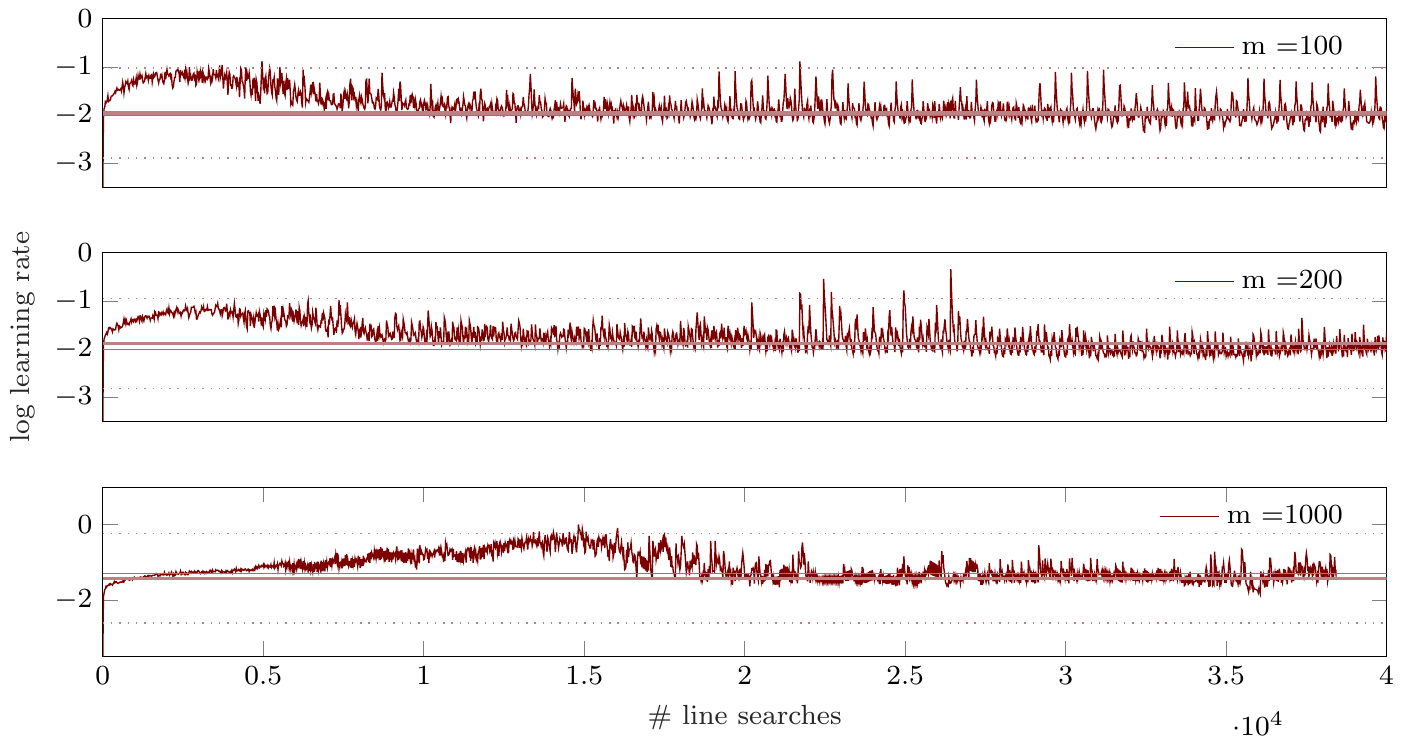}
  \caption{Traces of \emph{accepted logarithmic learning rates}. 
All runs are performed with default design parameters. 
Different rows show the same plot for different mini-batch sizes of $m = 100, 200$ and $1000$; plots and smoothing as in rows~2-4 of Figure~\ref{fig:cand-selection-exp} (details in text).
}
\label{fig:alphaTraces}
\end{figure}

\begin{figure}%[ht]
  \centering
  \setlength{\figwidth}{.9\textwidth}
  \setlength{\figheight}{.3\textheight}
  %\tikzset{external/remake next}
  %{\scriptsize \input{fig/MNIST_ARCH_1000_500_250_10_efficiency_noiseLevel.tikz}}  
   \includegraphics[scale=1.0]{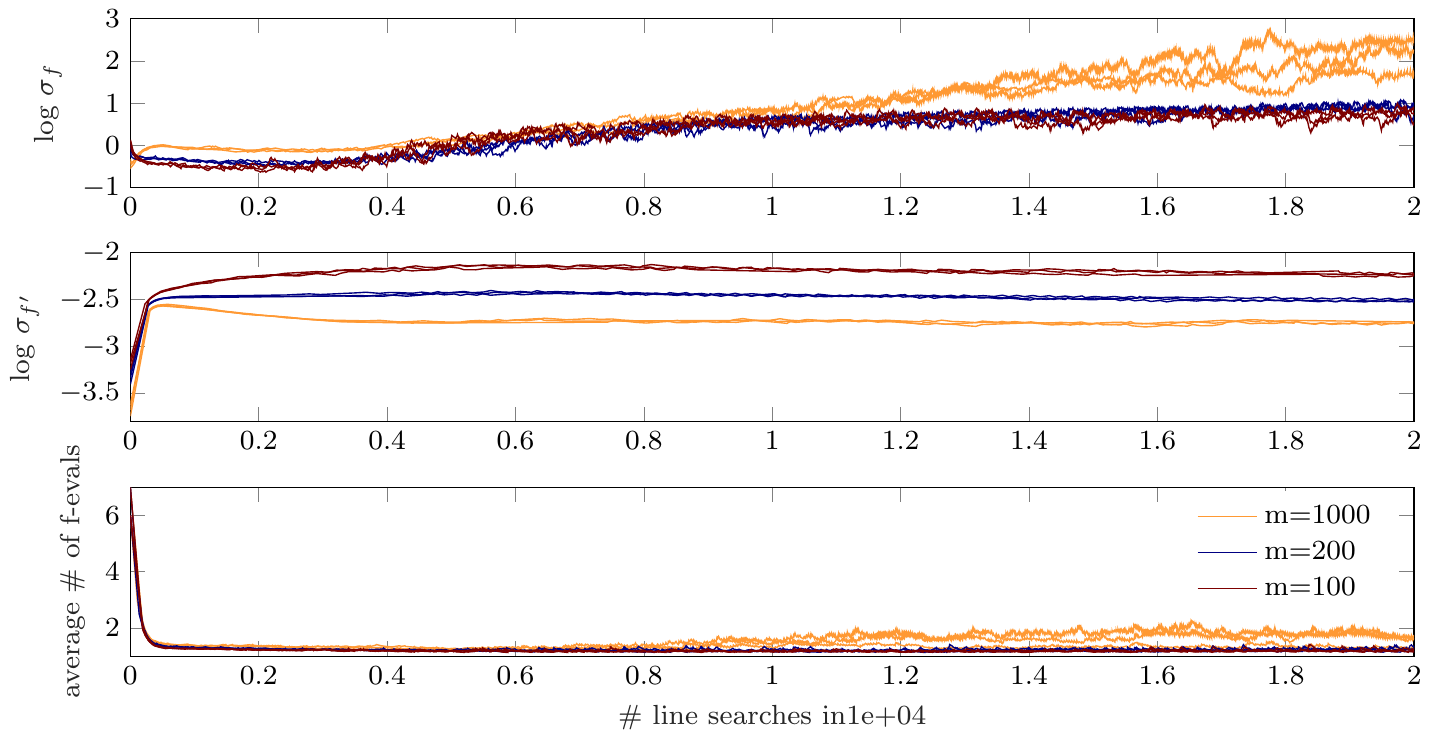}
  \caption{Traces of \emph{logarithmic noise levels $\sigma_f$} (top), $\sigma_{f'}$ (middle) and average number of function evaluations per line search (bottom). Setup and smoothing as in Figure~\ref{fig:alphaTraces}. Different colors correspond to different minibatch sizes (see legend). Curves of the same color correspond to different seeds (3 shown).
}
\label{fig:sigTraces}
\end{figure}

In the current implementation of the probabilistic line search, the choice among candidates for evaluation is done by evaluating an acquisition function $u_{\text{EI}}(t^{\text{cand}}_i)\cdot\pW(t^{\text{cand}}_i)$ at every candidate point $t^{\text{cand}}_i$; then choosing the one with the highest value for evaluation of the objective (\textsection \ref{sec:select-eval-points}). The Wolfe probability $\pW$ actually encodes precisely what kind of point we want to find and incorporates both (W-I and W-II) conditions about the function value \emph{and} to the gradient (\textsection \ref{sec:determ-conv}). However $\pW$ does not have very desirable exploration properties. Since the uncertainty of the \gp~grows to `the right' of the last observation, the Wolfe probability quickly drops to a low, approximately constant value there (Figure \ref{fig:EI-sketch}). Also $\pW$ is partially allowing for undesirably short steps (\textsection \ref{sec:extr-fact-alph}). 
The expected improvement $u_{\text{EI}}$, on the other hand, is a well studied acquisition function of Bayesian optimization trading off exploration and exploitation. It aims to globally find a point with a function value lower than a current best guess. Though this is a desirable property also for the probabilistic line search, it is lacking the information that we are seeking a point that also fulfills the W-II curvature condition. This is evident in Figure \ref{fig:EI-sketch} where $\pW$ significantly drops at points where the objective function is already evaluated but $u_{\text{EI}}$ does not.
In addition, we do not need to explore the positive $t$ space to an extend, the expected improvement suggests, since the aim of a line search is just to find \emph{a} good, acceptable point at positive $t$ and not the globally best one. The product of both acquisition function $u_{\text{EI}}\cdot\pW$ is thus a trade-off between exploring enough, but still preventing too much exploitation in obviously undesirable regions. In practice, though, we found that all three choices ((i) $u_{\text{EI}}\cdot\pW$, (ii) $u_{\text{EI}}$ only, (iii) $\pW$ only) perform comparable. The following experiments were all performed training N-II on MNIST; only the minibatch size might vary as indicated.

Figure \ref{fig:cand-selection-exp} compares all three choices for mini-batch size $m=200$ and default design parameters.
The top plot shows the evolution of the logarithmic test and train set error (for plot and color description see Figure caption). All test and train set error curves respectively bundle up (only lastly plotted clearly visible). The choice of acquisition function thus does not change the performance here. 
Rows 2-4 of Figure~\ref{fig:cand-selection-exp} show learning rate traces of a single seed. All three curves show very similar global behavior. First the learning rate grows, then drops again, and finally settles around the best found constant learning rate. This is intriguing since \emph{on average} a larger learning rate seems to be better at the beginning of the optimization process, then later dropping again to a smaller one. 
This might also explain why \sgd+\probLS~in the first part of the optimization progress outperforms vanilla \sgd~(Figure \ref{fig:noiseLevel}). Runs, that use just slightly larger constant learning rates than the best performing constant one (above the gray horizontal lines in Figure \ref{fig:cand-selection-exp}) were failing after a few steps. This shows that there is some non-trivial adaptation going on, not just globally, but locally at every step.

Figure~\ref{fig:alphaTraces} shows traces of accepted learning rates for different mini-batch sizes $m=100, 200, 1000$. Again the global behavior is qualitatively similar for all three mini-batch sizes on the given architecture. For the largest mini-batch size $m=1000$ (last row of Figure \ref{fig:alphaTraces}) the probabilistic line search accepts a larger learning rate (on average and in absolute value) than for the smaller mini-batch sizes $m=100$ and $200$, which is in agreement with practical experience and theoretical findings 
(\citet[][\textsection 4 and 7]{Hinton2012},
\citet[][\textsection 9.1.3]{Goodfellow-et-al-2016},
\citet[][]{2016arXiv161205086B}). 

Figure~\ref{fig:sigTraces} shows traces of the (scaled) noise levels $\sigma_f$ and $\sigma_{f'}$ and the average number of function evaluations per line search for different noise levels ($m=100, 200, 1000$); same colors show the same setup but different seeds. 
The average number of function evaluations rises very slightly to $\approx 1.5-2$ for minibatch size $m = 1000$ towards the end of the optimization process, in comparison to $\approx 1.5$ for $m = 100, 200$. This seems counter intuitive in a way, but since larger minibatch sizes also observe smaller value and gradients (especially towards the end of the optimization process), the relative noise levels might actually be larger.
(Although the curves for varying $m$ are shown versus the same abscissa, the corresponding optimizers might be in different regions of the loss surface, especially $m=1000$ probably reaches regions of smaller absolute gradients).
At the start of the optimization the average number of function evaluations is high, because the initial default learning rate is small ($10^{-4}$) and the line search extends each step multiple times.

% ===============================================================================
\section{Conclusion}
\label{sec:conclusion}

The line search paradigm widely accepted in deterministic optimization can be extended to noisy settings. Our design combines existing principles from the noise-free case with ideas from Bayesian optimization, adapted for efficiency. We arrived at a lightweight ``black-box'' algorithm that exposes no parameters to the user. 
Empirical evaluations so far show compatibility with the \sgd~search direction and viability for logistic regression and multi-layer perceptrons. 
The line search effectively frees users from worries about the choice of a learning rate: Any reasonable initial choice will be quickly adapted and lead to close to optimal performance.
Our matlab implementation can be found at {\url{http://tinyurl.com/probLineSearch}}.

% ===============================================================================
% Acknowledgements should go at the end, before appendices and references

%\acks{Acknowledgments will be added later.}
\acks{Thanks to Jonas Jaszkowic who prepared the base of the pseudo-code.}


\newpage
\appendix
% =================================================================
%\FloatBarrier

% =======================================================================
%\clearpage

% Note: in this sample, the section number is hard-coded in. Following
% proper LaTeX conventions, it should properly be coded as a reference:
%In this appendix we prove the following theorem from
%Section~\ref{sec:textree-generalization}:

% =========================================================================
\section*{Appendix A. -- Noise Estimation}
\manuallabel{app:noise-estimation}{A}
Section \ref{sec:noise-variances} introduced the statistical variance estimators 
\begin{equation}
  \label{eq:8}
  \begin{split}
    \Sigma'(x) &= (1-m)^{-1}(\hat{\nabla S}(x)-\nabla\hat{\mathcal{L}}(x)^{\Hm 2})\\
    \Sigma(x) &= (1-m)^{-1}(\hat{S}(x)-\hat{\mathcal{L}}(x)^{2})
  \end{split}
\end{equation}
of the function and gradient estimate $\hat{\mathcal{L}}(x)$ and $\nabla \hat{\mathcal{L}}(x)$ at position $x$. The underlying assumption is that $\hat{\mathcal{L}}(x)$ and $\nabla \hat{\mathcal{L}}(x)$ are distributed according to
\begin{equation}
  \label{eq:5}
  \begin{split}    
  \begin{bmatrix}
    \hat{\mathcal{L}}(x) \\ \nabla \hat{\mathcal{L}}(x)
  \end{bmatrix}
  &\sim \N\left(\
  \begin{bmatrix}
    \hat{\mathcal{L}}(x) \\ \nabla \hat{\mathcal{L}}(x)
  \end{bmatrix};
  \begin{bmatrix}
    \mathcal{L}(x) \\ \nabla \mathcal{L}(x)
  \end{bmatrix},
  \begin{bmatrix}
    \Sigma(x) & 0_{D\times 1}\\ 
    0_{1\times D}& \diag\Sigma'(x)
  \end{bmatrix}
\right)
  \end{split}
\end{equation}
which implies Eq~\ref{eq:3}
\begin{equation}
  \label{eq:27}
  \begin{split}    
  \begin{bmatrix}
    \hat{\mathcal{L}}(x) \\ s(x)'\cdot\nabla \hat{\mathcal{L}}(x)
  \end{bmatrix}=
  \begin{bmatrix}
    y(x) \\ y'(x)
  \end{bmatrix}
  &\sim \N\left(\
  \begin{bmatrix}
    y(x) \\ y'(x)
  \end{bmatrix};
  \begin{bmatrix}
    f(x) \\ f'(x)
  \end{bmatrix},
  \begin{bmatrix}
    \sigma_{f}(x) & 0\\ 
    0& \sigma_{f'}(x)
  \end{bmatrix}
\right).
  \end{split}
\end{equation}
where $s(x)$ is the possibly new search direction at $x$. This is an approximation since the true covariance matrix is in general not diagonal. A better estimator for the projected gradient noise would be (dropping $x$ from the notation) 
\begin{equation}
  \label{eq:20}
  \begin{split}
    \eta_{f'} 
    & = s\Trans \left[\frac{1}{m-1}\frac{1}{m}\sum_{k=1}^m(\nabla l^k - \nabla\hat{\mathcal{L}})(\nabla l^k - \nabla\hat{\mathcal{L}})\Trans\right] s\\
   & = \sum_{i, j = 1}^Ds_is_j \frac{1}{m-1}\frac{1}{m}\sum_{k = 1}^m\left(\nabla l^k_i - \nabla \hat{\mathcal{L}}_i\right)\left(\nabla l^k_j - \nabla \hat{\mathcal{L}}_j\right)\\
& = \frac{1}{m-1}\sum_{i, j = 1}^Ds_is_j \left(
  \frac{1}{m}\sum_{k = 1}^m \nabla l^k_i\nabla l^k_j 
-  \nabla \hat{\mathcal{L}}_i\nabla \hat{\mathcal{L}}_j  
-  \nabla \hat{\mathcal{L}}_j\nabla \hat{\mathcal{L}}_i  
+  \nabla \hat{\mathcal{L}}_i\nabla \hat{\mathcal{L}}_j \right)\\
& = \frac{1}{m-1} \left(
  \frac{1}{m}\sum_{k = 1}^m \sum_{i, j = 1}^Ds_i\nabla l^k_i s_j\nabla l^k_j 
- \sum_{i, j = 1}^Ds_j\nabla \hat{\mathcal{L}}_j s_i\nabla\hat{\mathcal{L}}_i \right)\\
& = \frac{1}{m-1} \left(
  \frac{1}{m}\sum_{k = 1}^m (s'\cdot \nabla l^k)^2% (s'\cdot\nabla l^k) 
-  (s'\cdot \nabla \hat{\mathcal{L}})^2\right).
%& = \frac{1}{m-1} \left(
%  \frac{1}{m}\sum_{k = 1}^m (s'\cdot \nabla l^k)^2
%- \left(\frac{1}{m}\sum_{k=1}^m (s'\cdot \nabla l^k)\right)^2 \right).
  \end{split}
\end{equation}
Comparing to $\sigma_{f'}$ yields
\begin{equation}
\label{eq:24}
  \begin{split}
    \eta_{f'} 
& = \frac{1}{m-1}\sum_{i, j = 1}^Ds_is_j \left(
  \frac{1}{m}\sum_{k = 1}^m \nabla l^k_i\nabla l^k_j 
-  \nabla \hat{\mathcal{L}}_j\nabla \hat{\mathcal{L}}_i \right)\\
& = \frac{1}{m-1}\sum_{i = 1}^Ds_i^2 \left(
  \frac{1}{m}\sum_{k = 1}^m (\nabla l^k_i)^2 
-  \nabla \hat{\mathcal{L}}_i^2 \right)\\
&\q+ \frac{1}{m-1}\sum_{i\neq j = 1}^Ds_is_j \left(
  \frac{1}{m}\sum_{k = 1}^m \nabla l^k_i\nabla l^k_j 
-  \nabla \hat{\mathcal{L}}_j\nabla \hat{\mathcal{L}}_i \right)\\
\eta_{f'} & = \sigma_{f'}+ \frac{1}{m-1}\sum_{i\neq j = 1}^Ds_is_j \left(
  \frac{1}{m}\sum_{k = 1}^m \nabla l^k_i\nabla l^k_j 
-  \nabla \hat{\mathcal{L}}_j\nabla \hat{\mathcal{L}}_i \right).
  \end{split}
\end{equation}
From Eq~\ref{eq:20} we see that, in order to effectively compute $\eta_{f'}$, we need an efficient way of computing the inner product $(s'\cdot \nabla l^k)$ for all $k$. In addition, we need to know the search direction $s(x)$ of the potential next step (if $x$ was accepted) at the time of computing $\eta_{f'}$. This is possible e.g. for the \sgd~search direction where $s(x) = -\frac{1}{m}\sum_{k=1}^m{\nabla l^k(x)}$ but potentially not possible or practical for arbitrary search directions. For all experiments in this paper we used the approximate variance estimator $\sigma_{f'}$. 

The above paragraph analyzed the independence assumption among gradient elements; the following paragraph is concerned with the independence assumption of gradient and function value $y$ and $y'$: In general $y$ and $y'$ are not independent since the algorithm draws them from the same minibatch; the likelihood including the correlation factor $\rho$ reads
\begin{equation}
  \label{eq:15}
  p(y _t,y' _t\g f) = \N\left(
    \begin{bmatrix}
      y _t\\ y' _t
    \end{bmatrix};
    \begin{bmatrix}
      f(t) \\ f'(t)
    \end{bmatrix},
    \begin{bmatrix}
      \sigma_{f}^2 & \rho \\ \rho & \sigma_{f'} ^2
    \end{bmatrix}
\right).
\end{equation}
The noise covariance matrix enters the \gp~only in the inverse of the sum with the kernel matrix of the observations. We can compute it analytically for one datapoint at position $t$, since it is only a $2\times 2$ matrix. For $\rho=0$, define:
\begin{equation}
  \label{eq:25}
  \begin{split}
      det_{\rho=0} & := [k_{tt}+\sigma_f ^2 ][\dkd_{tt} +\sigma_{f'} ^2 ] - \kd_{tt}\dk_{tt}\\
%   & = k_{tt}\dkd_{tt}+\sigma_f ^2\dkd_{tt} + k_{tt}\sigma_{f'} ^2 +\sigma_f ^2\sigma_{f'} ^2  + \kd_{tt}\dk_{tt} \\
  G^{-1}_{\rho=0} &:=  
  \begin{bmatrix}
      k_{tt}+\sigma_f ^2 & \kd_{tt} \\
      \dk_{tt} & \dkd_{tt} +\sigma_{f'} ^2 
    \end{bmatrix}^{-1}
= \frac{1}{det_{\rho=0}}
    \begin{bmatrix}
       \dkd_{tt} +\sigma_{f'} ^2 & -\kd_{tt} \\
       -\dk_{tt}  &k_{tt}+\sigma_f ^2 
    \end{bmatrix}.
  \end{split}
\end{equation}
For $\rho\neq 0$ we thus get:
\begin{equation}
  \label{eq:17}
  \begin{split}
      det_{\rho\neq 0} & := [k_{tt}+\sigma_f ^2 ][\dkd_{tt} +\sigma_{f'} ^2 ] - [\kd_{tt} + \rho][\dk_{tt} + \rho]\\
          & = det_{\rho=0} -  \rho(\dk_{tt} + \kd_{tt}) - \rho^2\\
  G^{-1}_{\rho\neq0} &:=   \begin{bmatrix}
      k_{tt}+\sigma_f ^2 & \kd_{tt} + \rho\\
      \dk_{tt} + \rho& \dkd_{tt} +\sigma_{f'} ^2 
    \end{bmatrix}^{-1}
= \frac{1}{det_{\rho\neq 0}}
    \begin{bmatrix}
       \dkd_{tt} +\sigma_{f'} ^2 & -(\kd_{tt} + \rho) \\
       -(\dk_{tt} + \rho) &k_{tt}+\sigma_f ^2 
    \end{bmatrix}\\
%&= G^{-1}_{\rho=0}\frac{det_{\rho=0} }{ det_{\rho=0} - \rho(\dk_{tt} + \kd_{tt}) - \rho^2}
%+ \frac{\begin{bmatrix}
%       0 & - \rho \\
%       -\rho &0 
%    \end{bmatrix}}{ det_{\rho=0} - \rho(\dk_{tt} + \kd_{tt}) - \rho^2}\\
&= \frac{det_{\rho=0} }{ det_{\rho\neq0}}G^{-1}_{\rho=0}
- \frac{\rho}{ det_{\rho\neq0} }
 \begin{bmatrix}
       0 & 1 \\
       1 &0 
    \end{bmatrix}\\
%
%&= G^{-1}_{\rho=0}\left[\frac{det_{\rho=0} }{ det_{\rho=0} - \rho(\dk_{tt} + \kd_{tt}) - \rho^2}
%+ \frac{ G_{\rho=0}\begin{bmatrix}
%       0 & - \rho \\
%       -\rho &0 
%    \end{bmatrix}}{ det_{\rho=0} - \rho(\dk_{tt} + \kd_{tt}) - \rho^2}\right]\\
%&= G^{-1}_{\rho=0}\left[\frac{det_{\rho=0} }{ det_{\rho=0} - \rho(\dk_{tt} + \kd_{tt}) - \rho^2}
%- \frac{ 
%  \rho\begin{bmatrix}
%      \kd_{tt}   &  k_{tt}+\sigma_f ^2 \\
%       \dkd_{tt} +\sigma_{f'} ^2  &  \dk_{tt}
%    \end{bmatrix}
%}{ det_{\rho=0} - \rho(\dk_{tt} + \kd_{tt}) - \rho^2}\right]
  \end{split}
\end{equation}
The fraction $\nicefrac{det_{\rho=0} }{ det_{\rho\neq0}}$ in the first term of the last row, is a positive scalar that scales all element of $G^{-1}_{\rho=0}$ equally (since $G_{\rho=0}$ and $ G_{\rho\neq 0}$ are positive definite matrices, we know that $det_{\rho=0}>0$, $det_{\rho\neq0}>0$). 
If $|\rho|$ is small in comparison to the determinant $det_{\rho=0}$, then $det_{\rho\neq 0}\approx det_{\rho=0}$ and the scaling factor is approximately one. 
The second term corrects off-diagonal elements in $G_{\rho\neq 0}$ and is proportional to $\rho$; if $|\rho|\ll det_{\rho=0}$ this term is small as well.

In might be possible to estimate $\rho$ as well from the minibatch in a similar style to the estimation of $\sigma_f$ and $\sigma_{f'}$; it is not clear from this analysis, if the additional computational cost would justify the improvements in the \gp-inference.

\newpage
\section*{Appendix B. -- Noise Sensitivity}
\manuallabel{app:noise-level}{B}
\begin{figure}[h]
  \centering
  \setlength{\figwidth}{.9\textwidth}
  \setlength{\figheight}{.12\textheight}
%  {\scriptsize 
%    %\tikzset{external/remake next}
%    \input{fig/MNIST_ARCH_800_10_NEWPARAS_noiseLevel_ter_vs_alpha.tikz}\\
%    %\tikzset{external/remake next}
%    \input{fig/MNIST_ARCH_800_10_NEWPARAS_noiseLevel_trainter_vs_epoch_ALL.tikz}\\
%    %\tikzset{external/remake next}
%    \input{fig/MNIST_ARCH_800_10_NEWPARAS_noiseLevel_testter_vs_epoch_ALL.tikz}}
    \includegraphics[scale=1.0]{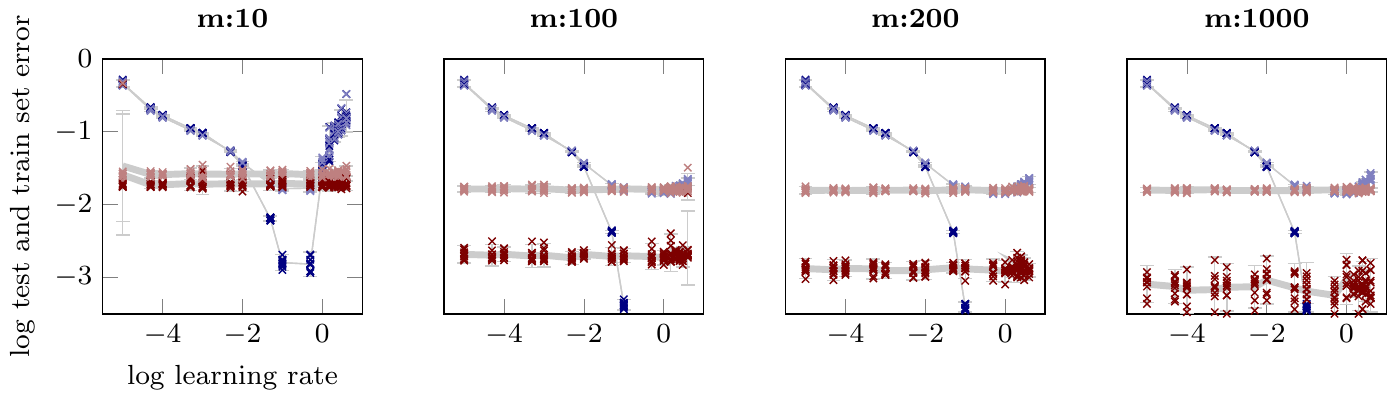}\\
    \includegraphics[scale=1.0]{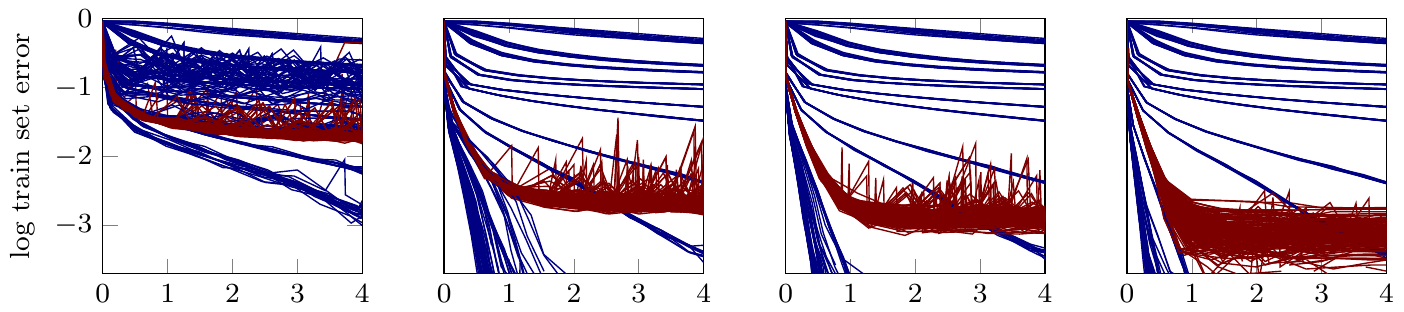}\\
    \includegraphics[scale=1.0]{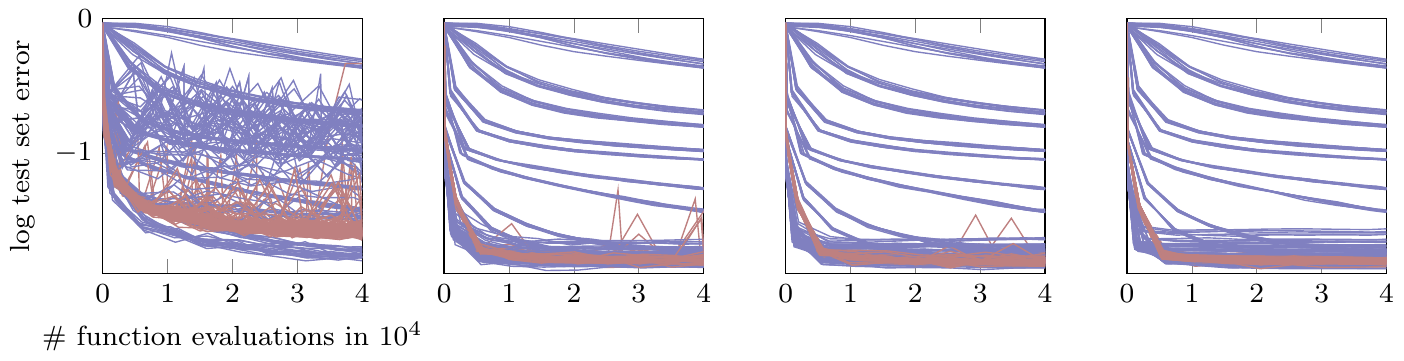}
  \caption{Performance of N-I on MNIST for \emph{varying mini-batch sizes}; plots and colors same as in Figure \ref{fig:noiseLevel} (middle plots cropped for readability). 
}
\label{fig:noiseMNIST-NI}
\end{figure}
\begin{figure}[h]
  \centering
  \setlength{\figwidth}{.9\textwidth}
  \setlength{\figheight}{.12\textheight}
%  {\scriptsize 
%    %\tikzset{external/remake next}
%    \input{fig/CIFAR10_ARCH_1000_500_250_10_NEWPARAS_noiseLevel_ter_vs_alpha.tikz}\\
%    %\tikzset{external/remake next}
%    \input{fig/CIFAR10_ARCH_1000_500_250_10_NEWPARAS_noiseLevel_trainter_vs_epoch_ALL.tikz}\\
%    %\tikzset{external/remake next}
%    \input{fig/CIFAR10_ARCH_1000_500_250_10_NEWPARAS_noiseLevel_testter_vs_epoch_ALL.tikz}}
    \includegraphics[scale=1.0]{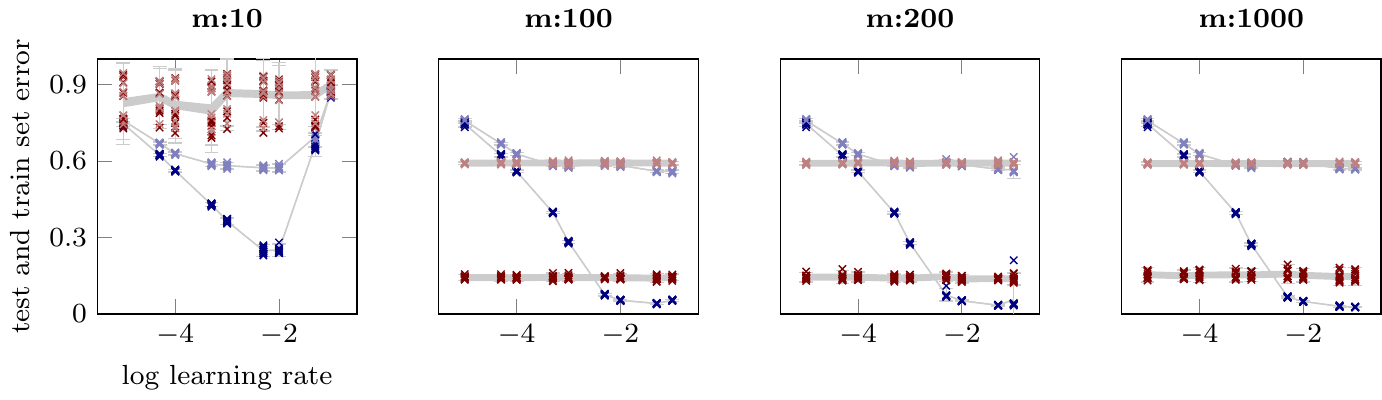}\\
    \includegraphics[scale=1.0]{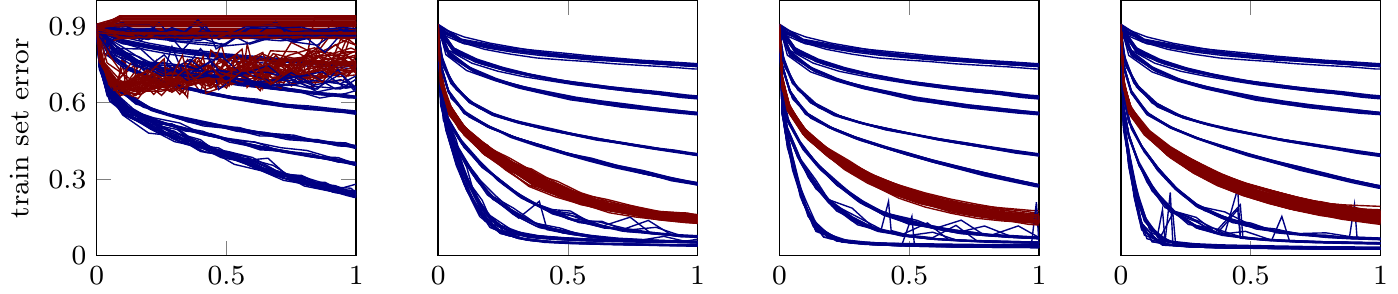}\\
    \includegraphics[scale=1.0]{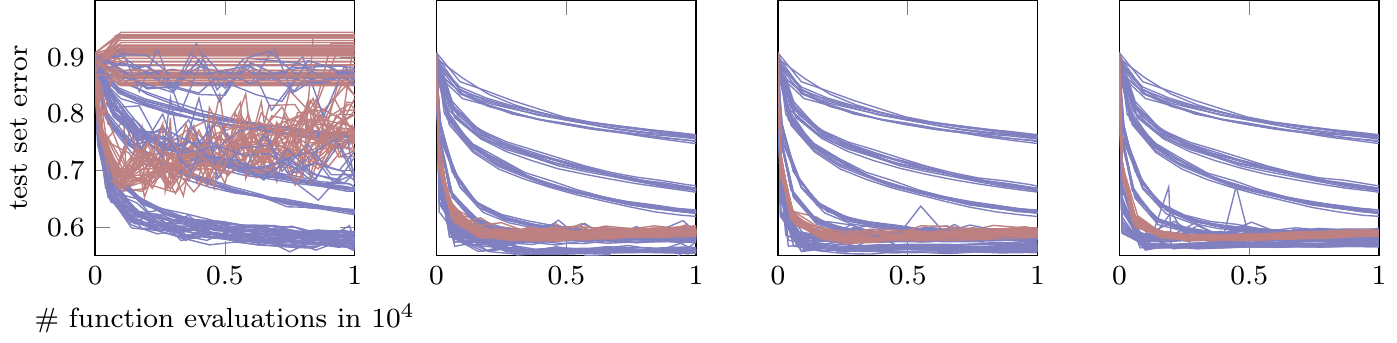}
  \caption{Performance of N-II on CIFAR-10 for \emph{varying mini-batch sizes}; plots and colors same as in Figure \ref{fig:noiseLevel}, except the scaling of the y-axis which is not logarithmic here. 
}
\label{fig:noiseCIFAR-NII}
\end{figure}
\begin{figure}[h]
  \centering
  \setlength{\figwidth}{.9\textwidth}
  \setlength{\figheight}{.12\textheight}
%  {\scriptsize 
%    %\tikzset{external/remake next}
%    \input{fig/CIFAR10_ARCH_800_10_NEWPARAS_noiseLevel_ter_vs_alpha.tikz}\\
%    %\tikzset{external/remake next}
%    \input{fig/CIFAR10_ARCH_800_10_NEWPARAS_noiseLevel_trainter_vs_epoch_ALL.tikz}\\
%    %\tikzset{external/remake next}
%    \input{fig/CIFAR10_ARCH_800_10_NEWPARAS_noiseLevel_testter_vs_epoch_ALL.tikz}}
    \includegraphics[scale=1.0]{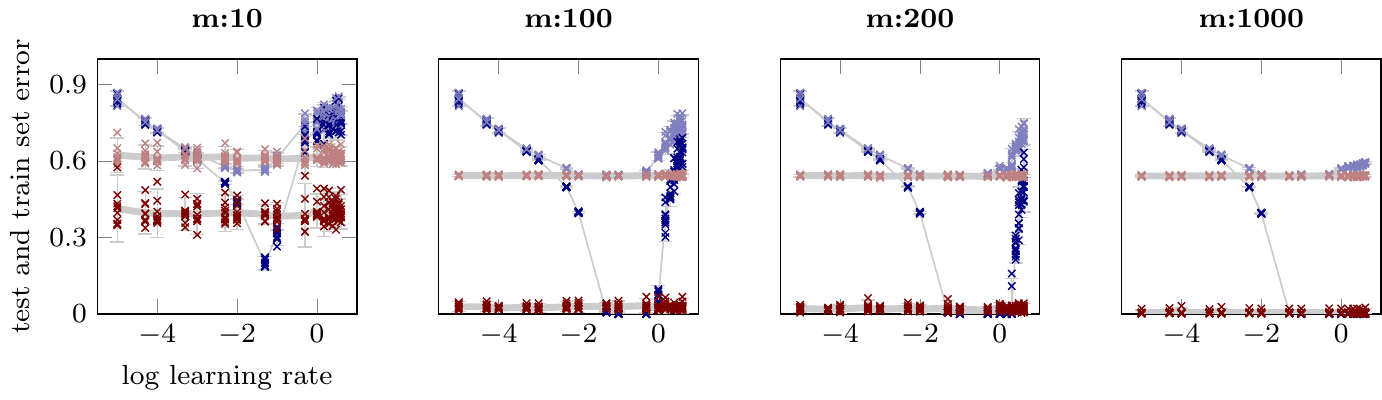}\\
    \includegraphics[scale=1.0]{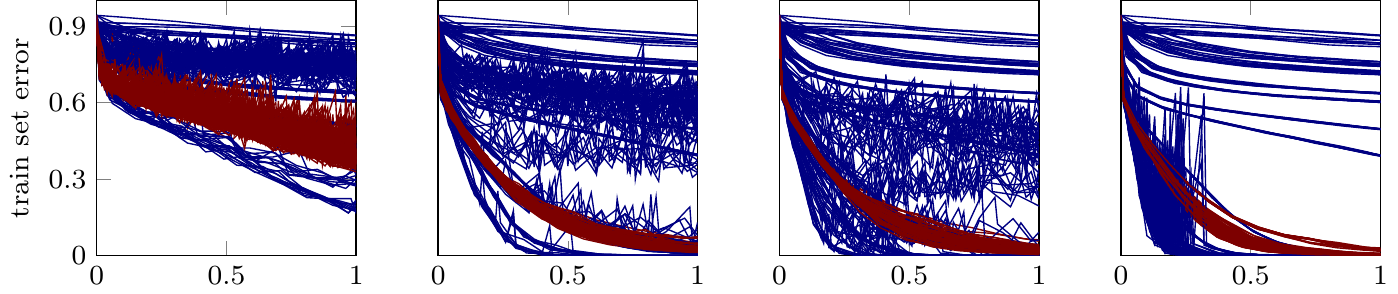}\\
    \includegraphics[scale=1.0]{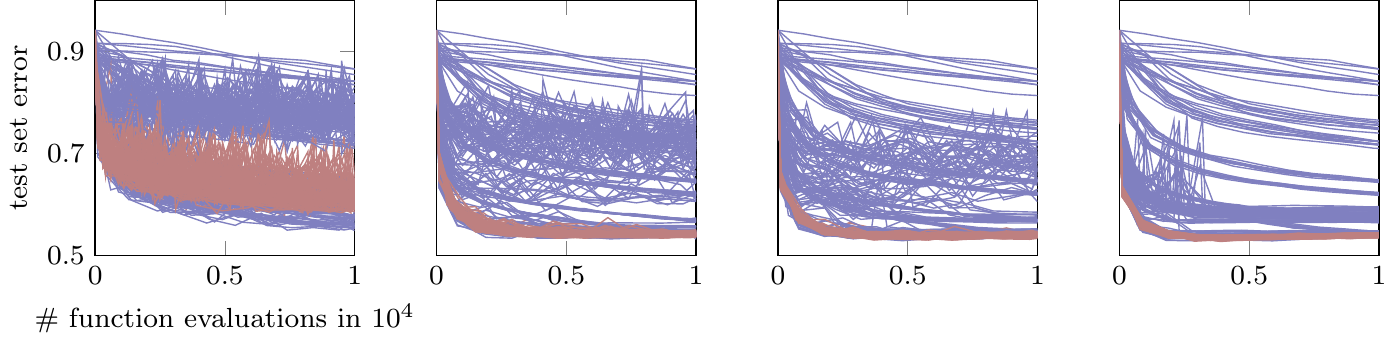}
  \caption{Performance of N-I on CIFAR-10 for \emph{varying mini-batch sizes}; plots and colors same as in Figure \ref{fig:noiseCIFAR-NII}. 
}
\label{fig:noiseCIFAR-NI}
\end{figure}
\begin{figure}[h]
  \centering
  \setlength{\figwidth}{.9\textwidth}
  \setlength{\figheight}{.12\textheight}
%  {\scriptsize 
%    %\tikzset{external/remake next}
%    \input{fig/GISETTE_ARCH_1_NEWPARAS_noiseLevel_ter_vs_alpha.tikz}\\
%    %\tikzset{external/remake next}
%    \input{fig/GISETTE_ARCH_1_NEWPARAS_noiseLevel_trainter_vs_epoch_ALL.tikz}\\
%    %\tikzset{external/remake next}
%    \input{fig/GISETTE_ARCH_1_NEWPARAS_noiseLevel_testter_vs_epoch_ALL.tikz}}
    \includegraphics[scale=1.0]{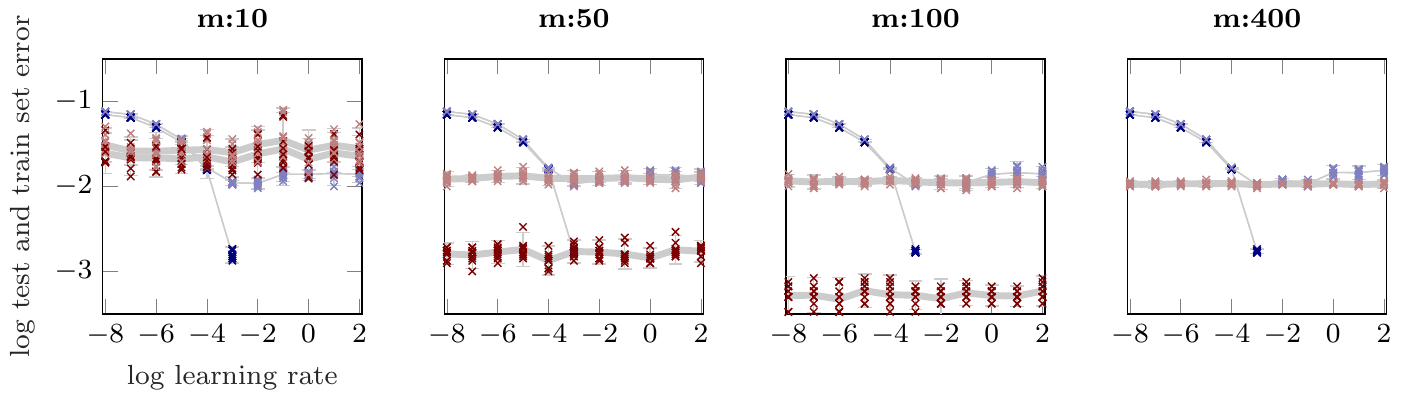}\\
    \includegraphics[scale=1.0]{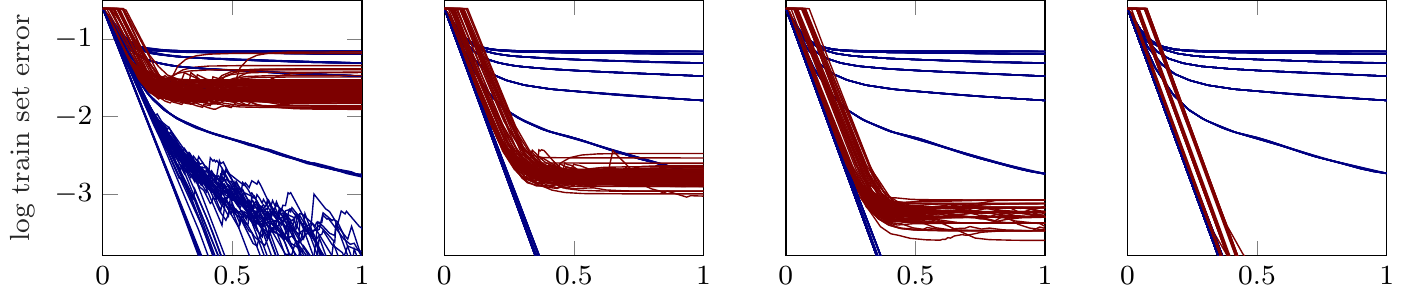}\\
    \includegraphics[scale=1.0]{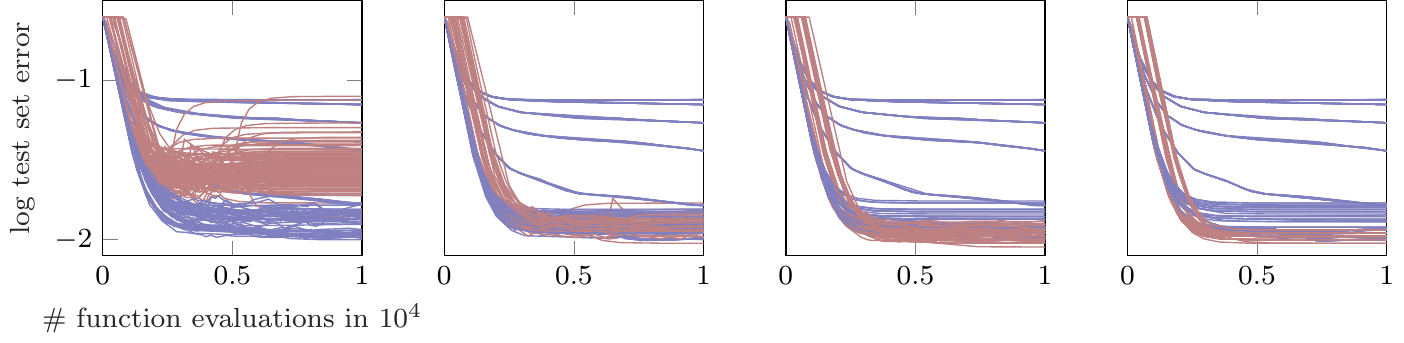}
  \caption{Performance of N-III on GISETTE for \emph{varying mini-batch sizes}; plots and colors same as in Figure \ref{fig:noiseCIFAR-NII}. 
}
\label{fig:noiseGISETTE-NIII}
\end{figure}\
\begin{figure}[h]
  \centering
  \setlength{\figwidth}{.9\textwidth}
  \setlength{\figheight}{.12\textheight}
%  {\scriptsize 
%    %\tikzset{external/remake next}
%    \input{fig/WBCD_ARCH_1_NEWPARAS_noiseLevel_ter_vs_alpha.tikz}\\
%    %\tikzset{external/remake next}
%    \input{fig/WBCD_ARCH_1_NEWPARAS_noiseLevel_trainter_vs_epoch_ALL.tikz}\\
%    %\tikzset{external/remake next}
%    \input{fig/WBCD_ARCH_1_NEWPARAS_noiseLevel_testter_vs_epoch_ALL.tikz}}
    \includegraphics[scale=1.0]{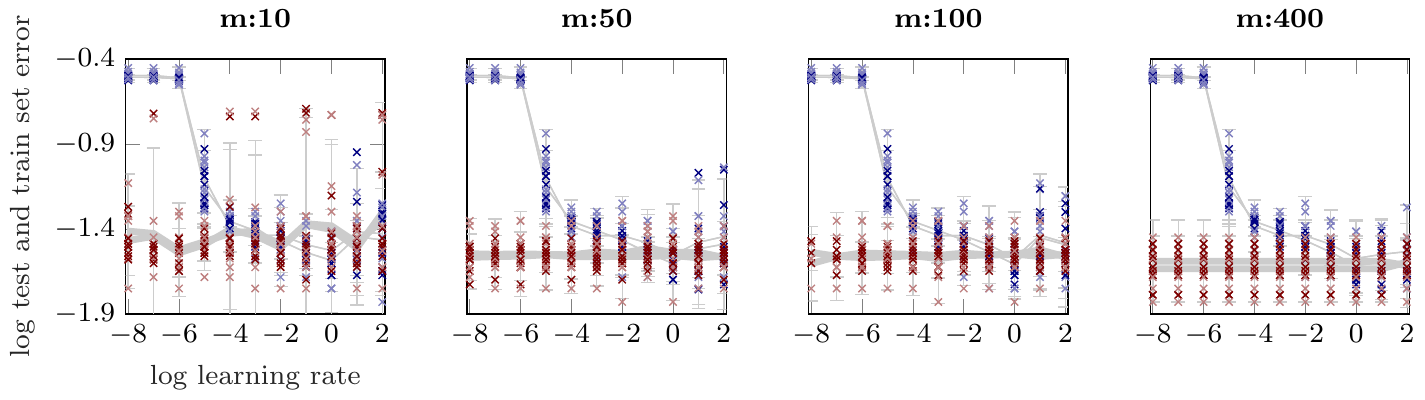}\\
    \includegraphics[scale=1.0]{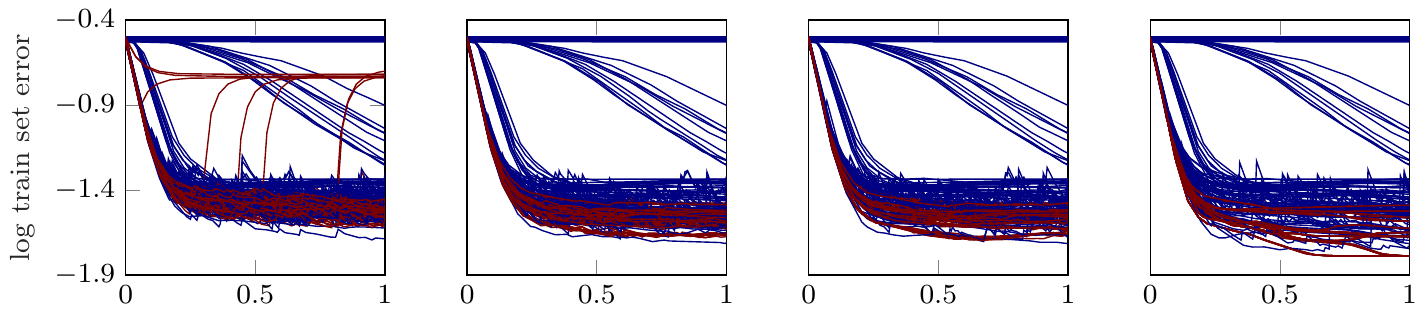}\\
    \includegraphics[scale=1.0]{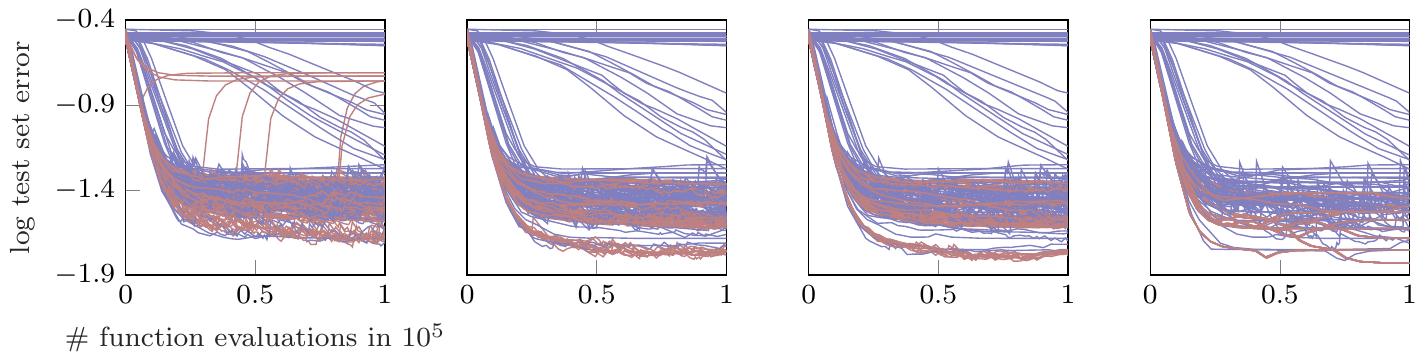}
  \caption{Performance of N-III on WDBC for \emph{varying mini-batch sizes}; plots, colors, and description same as in Figure \ref{fig:noiseGISETTE-NIII}. \emph{Remark:} since the training set is of size 400, the most right column ($m=400$) in fact runs full-batch gradient descent; this is not a problem, since the probabilistic line search can handle noise free observations as well.}
\label{fig:noiseWBCD-NIII}
\end{figure}
\begin{figure}[h]
  \centering
  \setlength{\figwidth}{.9\textwidth}
  \setlength{\figheight}{.12\textheight}
%  {\scriptsize 
%    \tikzset{external/remake next}
%    \input{fig/EPSILON_ARCH_1_NEWPARAS_noiseLevel_ter_vs_alpha.tikz}\\
%    \tikzset{external/remake next}
%    \input{fig/EPSILON_ARCH_1_NEWPARAS_noiseLevel_trainter_vs_epoch_ALL.tikz}\\
%    \tikzset{external/remake next}
%    \input{fig/EPSILON_ARCH_1_NEWPARAS_noiseLevel_testter_vs_epoch_ALL.tikz}}
    \includegraphics[scale=1.0]{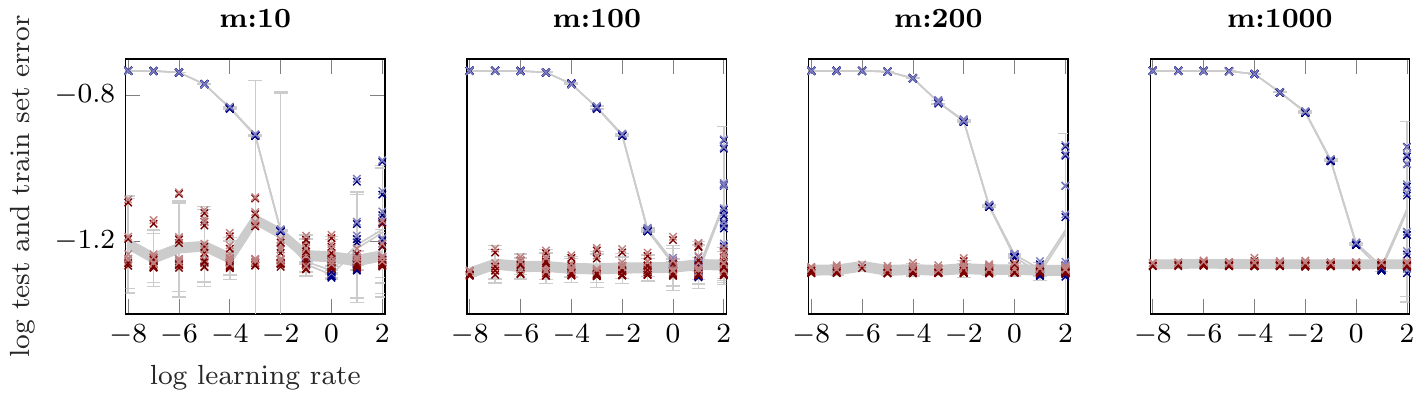}\\
    \includegraphics[scale=1.0]{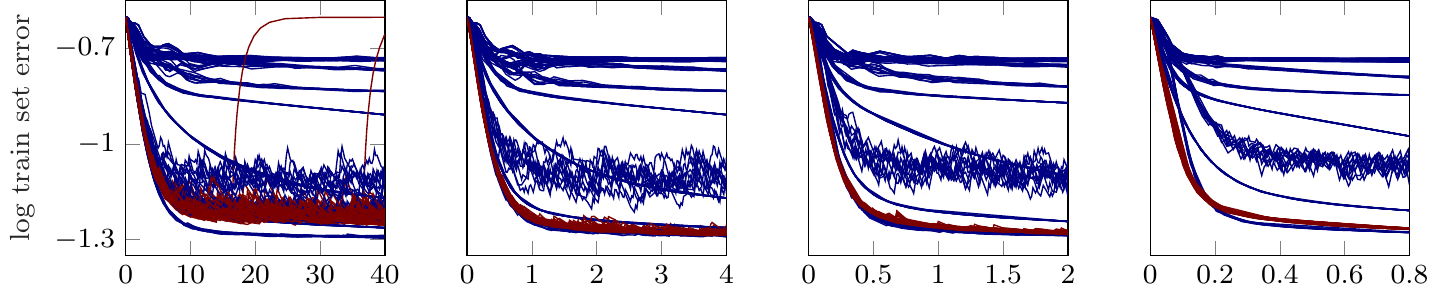}\\
    \includegraphics[scale=1.0]{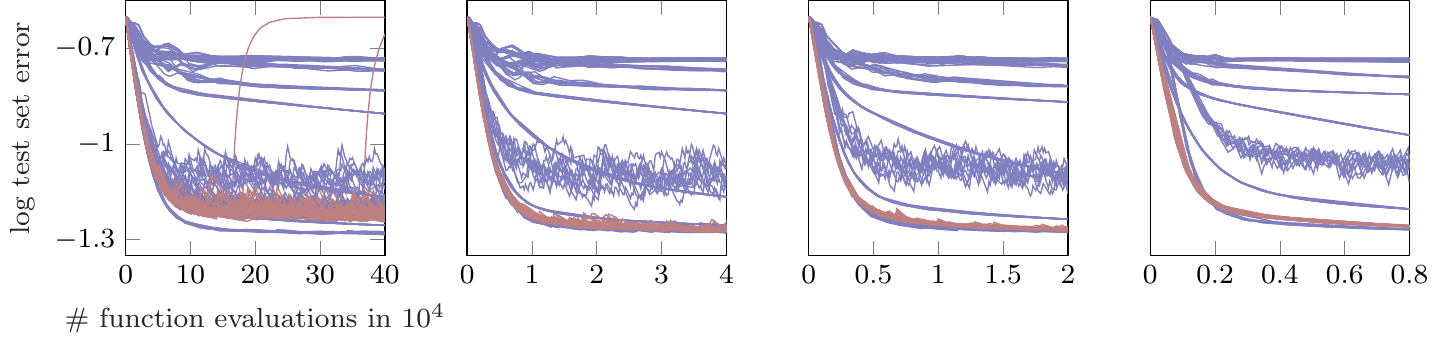}
  \caption{Performance of N-III on EPSILON for \emph{varying mini-batch sizes}; plots, colors, and description same as in Figure \ref{fig:noiseGISETTE-NIII}. EPSILON is the largest dataset, that was used in the experiments (400k samples); this did not seem to impair the performance of the line search or variance estimator.}
\label{fig:noiseEPSILON-NIII}
\end{figure}

\clearpage
\section*{Appendix C. -- Parameter Sensitivity}
\manuallabel{app:para-sens}{C}
% ------------------
\begin{figure}[h]
  \centering
  \setlength{\figwidth}{.9\textwidth}
  \setlength{\figheight}{.3\textheight}
  %\tikzset{external/remake next}
%  {\scriptsize \input{fig/MNIST_ARCH_1000_500_250_10_m00200_paraSens_cw_c2_ext10.tikz}}  
  \includegraphics[scale=1.0]{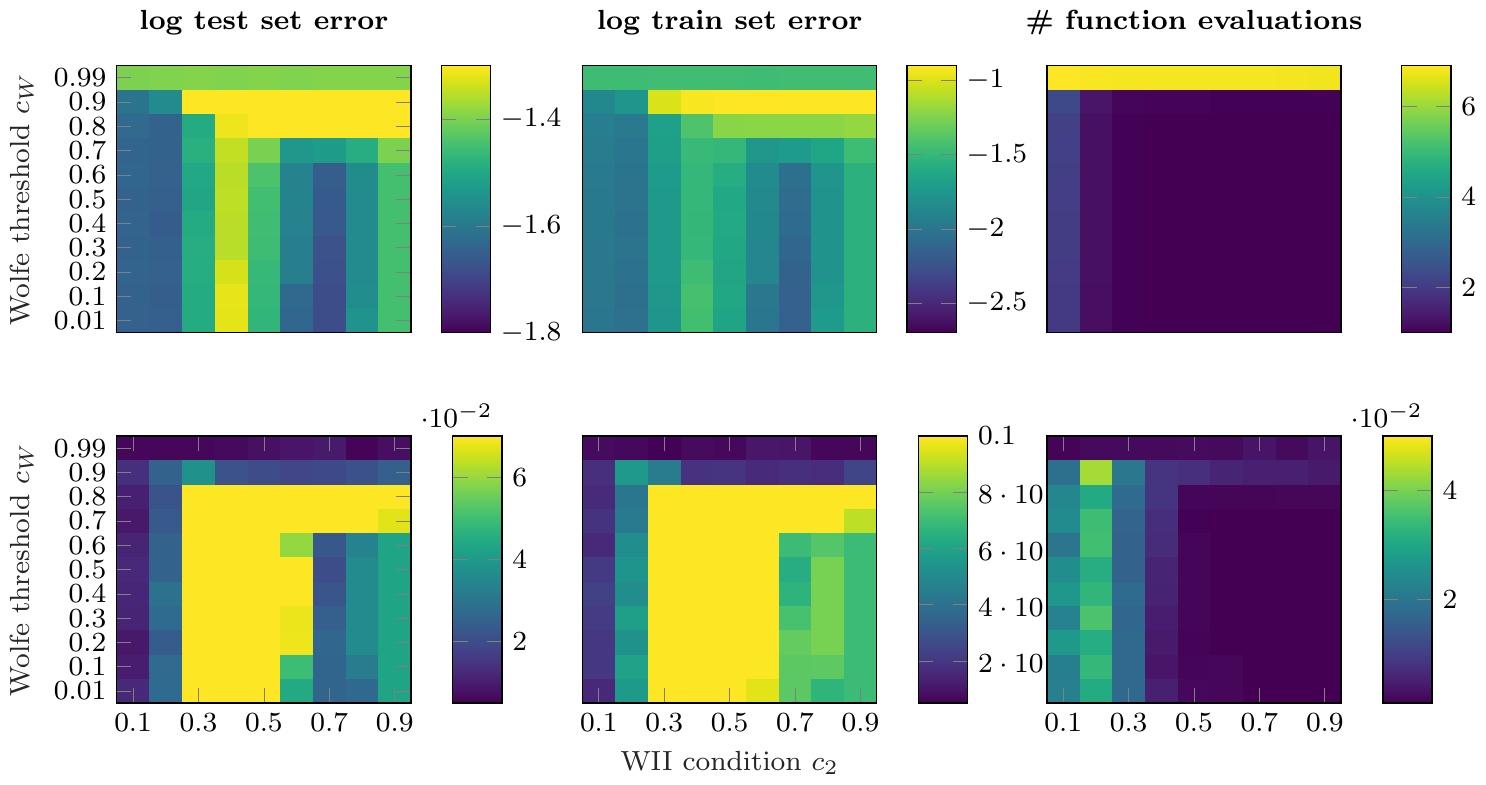}
  \caption{\emph{Sensitivity to varying hyper-parameters} $c_2$, and $c_W$ and \emph{fixed} $\alpha_{\text{ext}} = 1.0$ (\textsection \ref{sec:hyperp-estim}). Experimental setup as in Figures \ref{fig:c2extspace_cwall} and \ref{fig:cwc2space_extall}.
\emph{Top row} from left to right: logarithmic test set error, train set error, and average number of function evaluations per line search averaged over $10$ different initializations.
\emph{Bottom row:} corresponding relative standard deviations. In all plots darker colors are better. For extrapolation parameters $\alpha_{\text{ext}}>1$ (see Figures \ref{fig:c2cwspace11}, \ref{fig:c2cwspace12}, \ref{fig:c2cwspace13}, and \ref{fig:c2cwspace14}) the different parameter combinations all result in similar good performance. Only at extreme choices, for example $\alpha_{\text{ext}} = 1.0$ (this figure), which amounts to no extrapolation at all in between successive line searches, the line search becomes unstable. At the extreme value of $c_W = 0.99$, which amounts to imposing nearly absolute certainty about the Wolfe conditions, the line search becomes less efficient, though still does not break. In Figure \ref{fig:c2cwspace13} the default values adopted in the line search implementation  ($c_W = 0.3$, $c_2 = 0.5$, and $\alpha_{\text{ext}}=1.3$) are indicated as red dots.}
\label{fig:c2cwspace10}
\end{figure}
\begin{figure}[h]
  \centering
  \setlength{\figwidth}{.9\textwidth}
  \setlength{\figheight}{.3\textheight}
  %\tikzset{external/remake next}
  %{\scriptsize \input{fig/MNIST_ARCH_1000_500_250_10_m00200_paraSens_cw_c2_ext11.tikz}}   
  \includegraphics[scale=1.0]{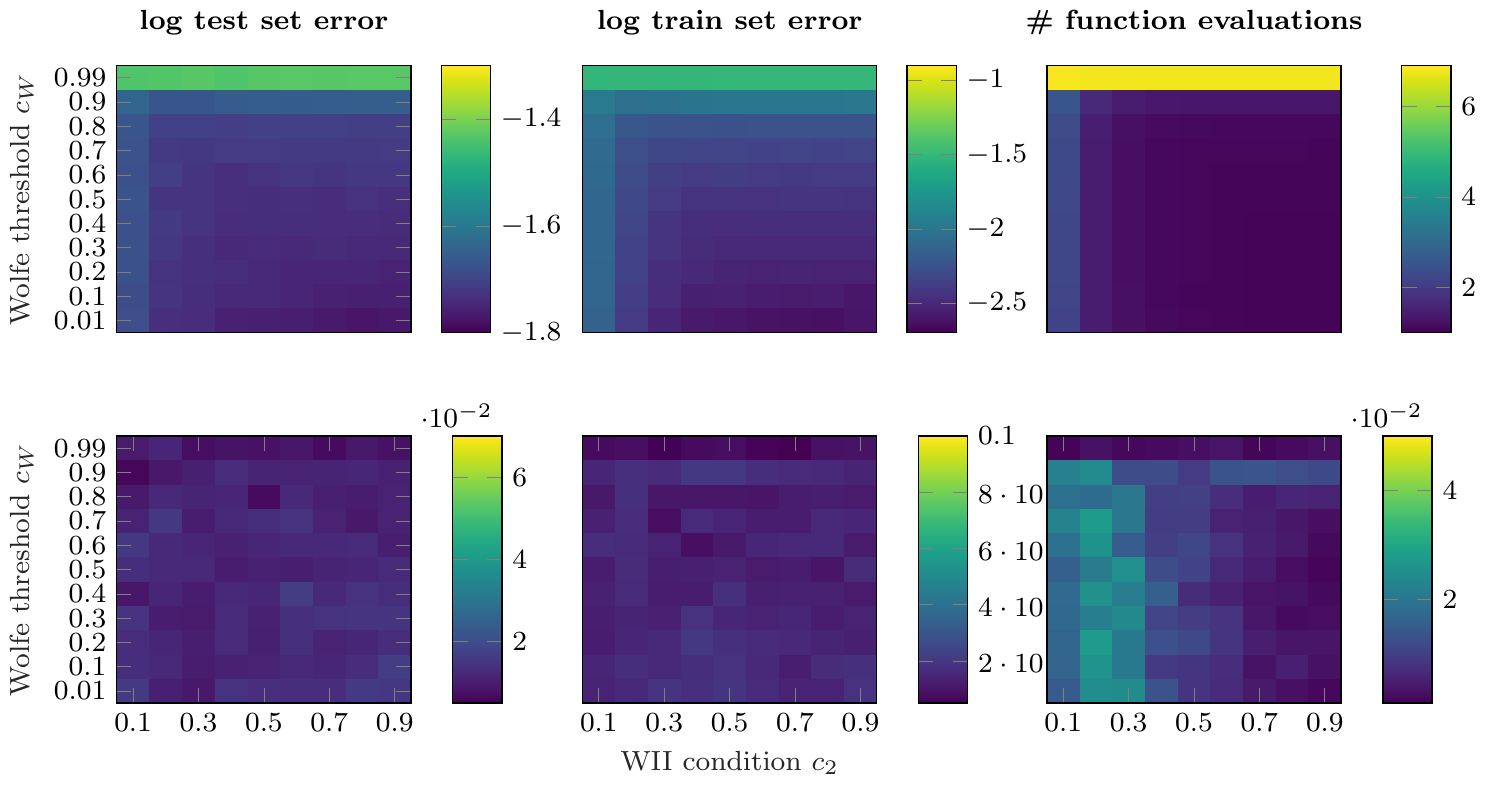}
  \caption{Same as Figure \ref{fig:c2cwspace10} but for \emph{fixed} $\alpha_{\text{ext}} = 1.1$}
\label{fig:c2cwspace11}
\end{figure}
\begin{figure}[h]
  \centering
  \setlength{\figwidth}{.9\textwidth}
  \setlength{\figheight}{.3\textheight}
  %\tikzset{external/remake next}
  %{\scriptsize \input{fig/MNIST_ARCH_1000_500_250_10_m00200_paraSens_cw_c2_ext12.tikz}}  
  \includegraphics[scale=1.0]{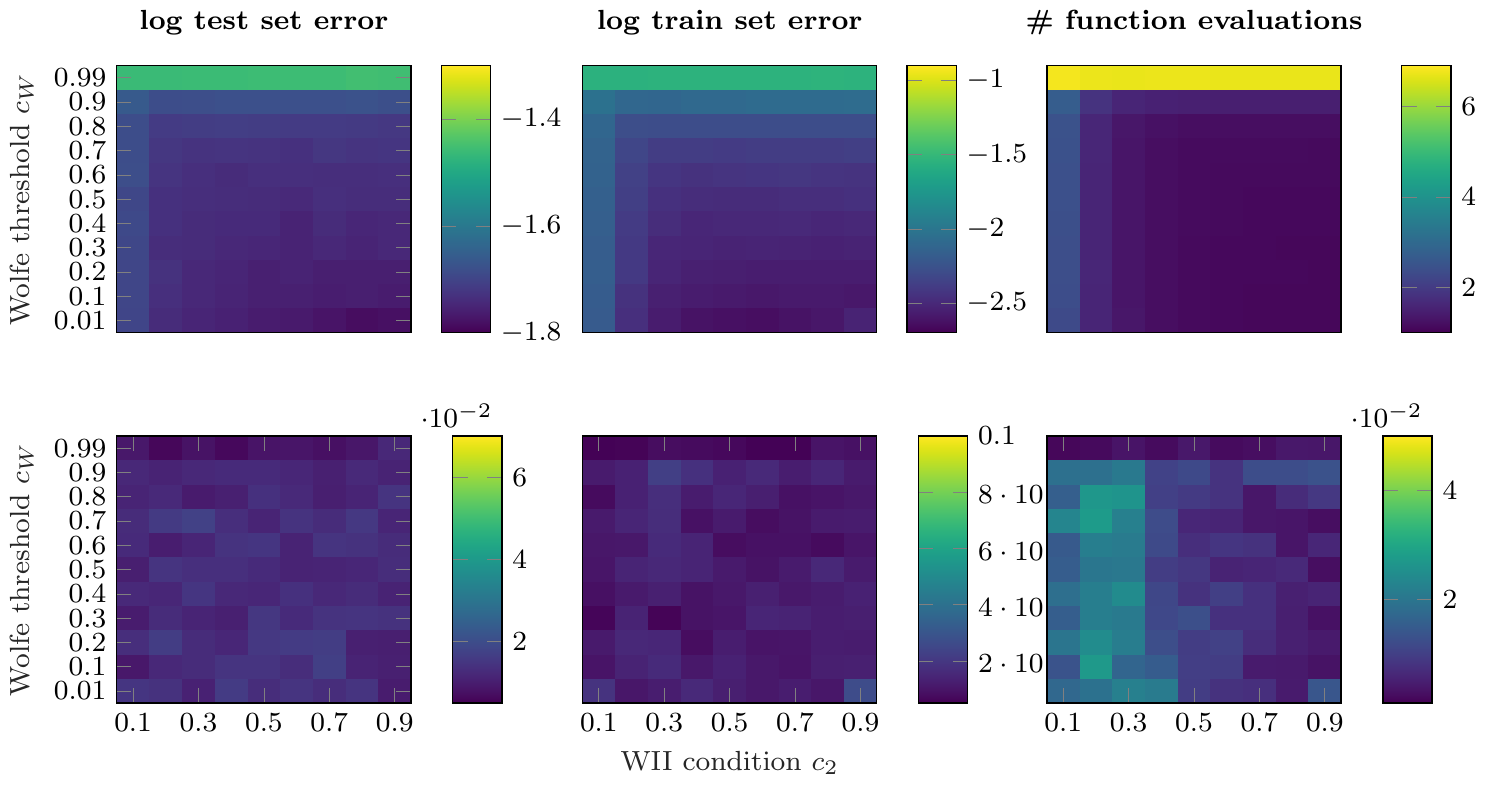}
  \caption{Same as Figure \ref{fig:c2cwspace10} but for \emph{fixed} $\alpha_{\text{ext}} = 1.2$}
\label{fig:c2cwspace12}
\end{figure}
\begin{figure}[h]
  \centering
  \setlength{\figwidth}{.9\textwidth}
  \setlength{\figheight}{.3\textheight}
  %\tikzset{external/remake next}
  %{\scriptsize \input{fig/MNIST_ARCH_1000_500_250_10_m00200_paraSens_cw_c2_ext13.tikz}}  
  \includegraphics[scale=1.0]{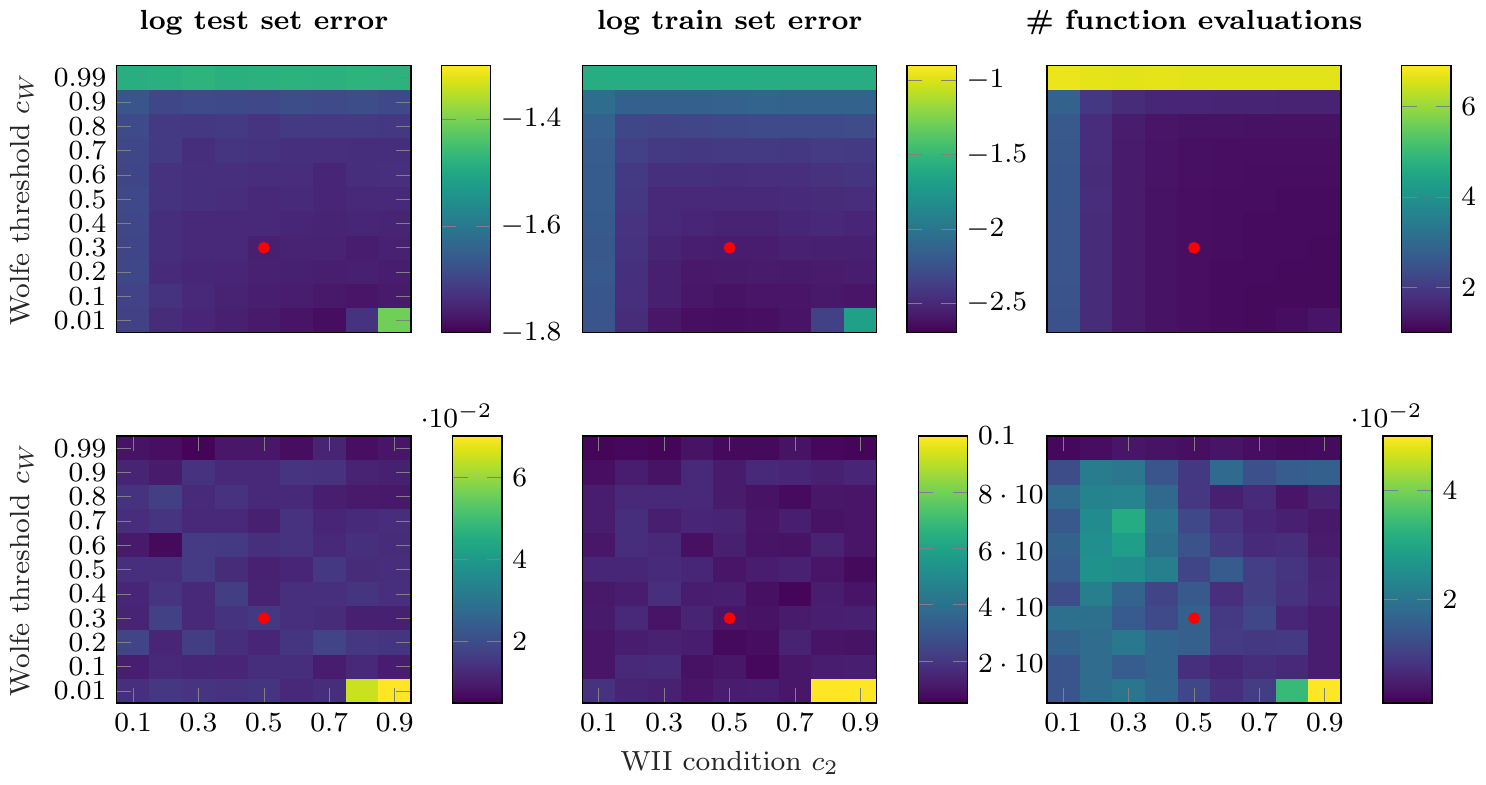}
  \caption{Same as Figure \ref{fig:c2cwspace10} but for \emph{fixed} $\alpha_{\text{ext}} = 1.3$. The default values adopted in the line search implementation ($c_W = 0.3$, $c_2 = 0.5$, and $\alpha_{\text{ext}}=1.3$) are indicated as red dots.}
\label{fig:c2cwspace13}
\end{figure}
\begin{figure}[h]
  \centering
  \setlength{\figwidth}{.9\textwidth}
  \setlength{\figheight}{.3\textheight}
  %\tikzset{external/remake next}
  %{\scriptsize \input{fig/MNIST_ARCH_1000_500_250_10_m00200_paraSens_cw_c2_ext14.tikz}}  
  \includegraphics[scale=1.0]{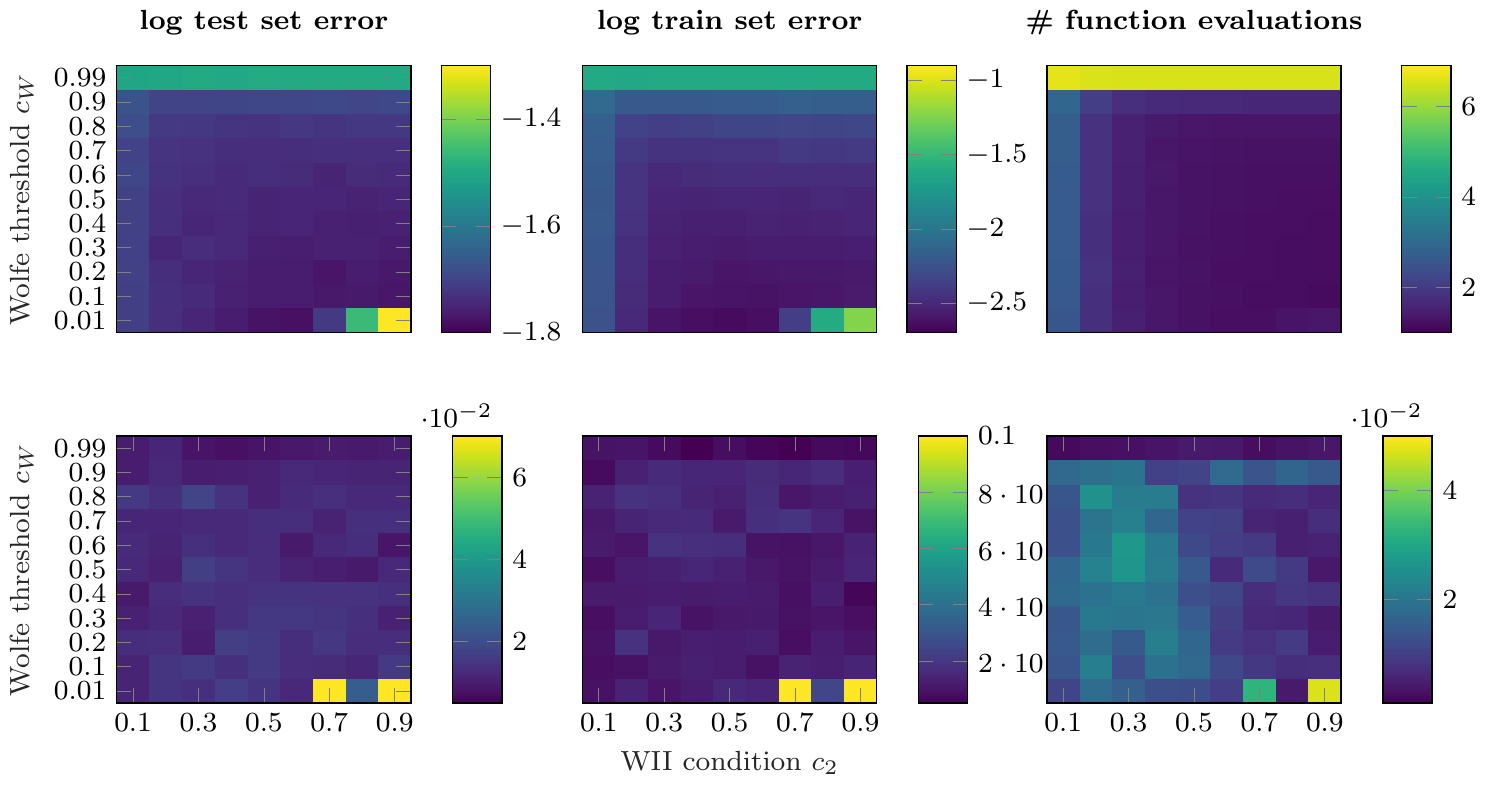}
  \caption{Same as Figure \ref{fig:c2cwspace10} but for \emph{fixed} $\alpha_{\text{ext}} = 1.4$}
\label{fig:c2cwspace14}
\end{figure}
%\clearpage
% ----------------------------------------------
% ------------------
\begin{figure}[h]
  \centering
  \setlength{\figwidth}{.9\textwidth}
  \setlength{\figheight}{.2\textheight}
  %\tikzset{external/remake next}
  %{\scriptsize \input{fig/MNIST_ARCH_1000_500_250_10_m00200_paraSens_ext_c2_cw01_imagesc.tikz}}  
  \includegraphics[scale=1.0]{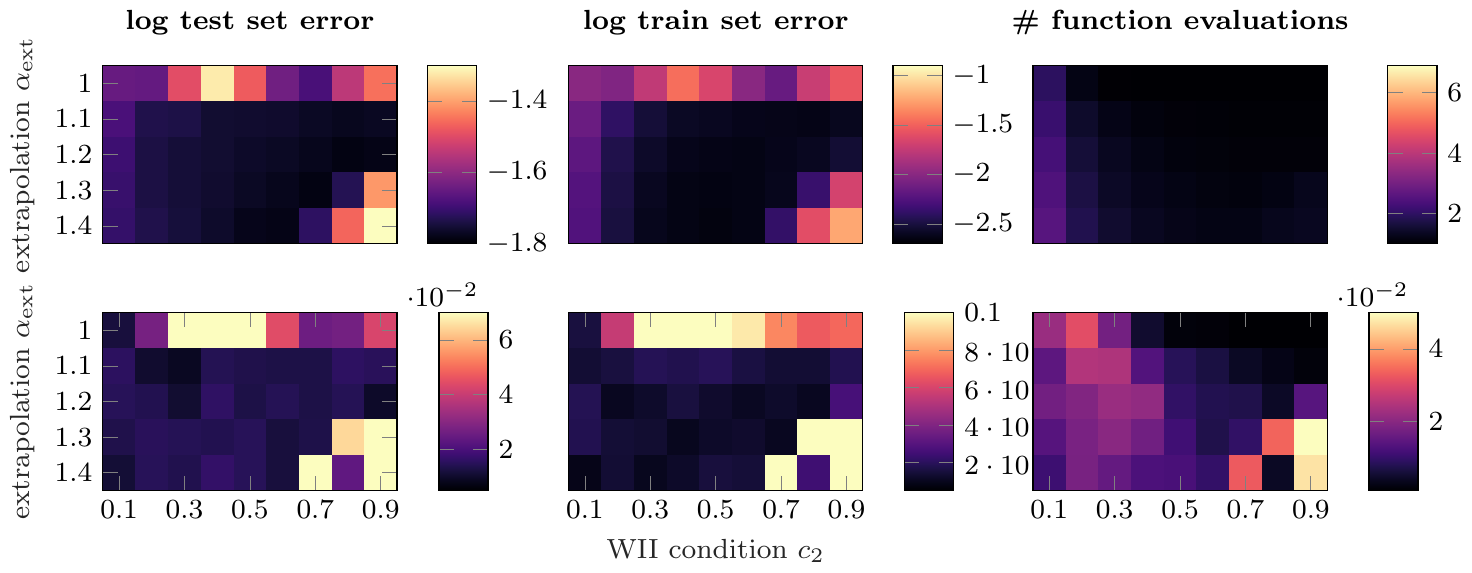}
  \caption{\emph{Sensitivity to varying hyper-parameters} $c_2$, and $\alpha_{\text{ext}}$ and \emph{fixed} $c_W = 0.01$ (\textsection \ref{sec:hyperp-estim}). Experimental setup as in Figures \ref{fig:c2extspace_cwall} and \ref{fig:cwc2space_extall} and plots like in Figure \ref{fig:c2cwspace10}.
In all plots darker colors are better. All choices of $c_W$ result in good performance though very tight choices of $c_W = 0.99$ (Figure \ref{fig:c2extspace99}), which amounts to imposing nearly absolute certainty about the Wolfe conditions, are less efficient. As described by Figure \ref{fig:c2cwspace10}, dropping the extrapolation factor $\alpha_{\text{ext}} \rightarrow 1$ in combination with a loose curvature condition (large $c_2$) renders the line search to break (top row, right half of columns in Figures \ref{fig:c2extspace01}--\ref{fig:c2extspace90}).
In Figure \ref{fig:c2extspace50} the default values adopted in the line search implementation  ($c_W = 0.3$, $c_2 = 0.5$, and $\alpha_{\text{ext}}=1.3$) are indicated as red dots.}
\label{fig:c2extspace01}
\end{figure}
\begin{figure}[h]
  \centering
  \setlength{\figwidth}{.9\textwidth}
  \setlength{\figheight}{.2\textheight}
  %\tikzset{external/remake next}
  %{\scriptsize \input{fig/MNIST_ARCH_1000_500_250_10_m00200_paraSens_ext_c2_cw10_imagesc.tikz}}    
 \includegraphics[scale=1.0]{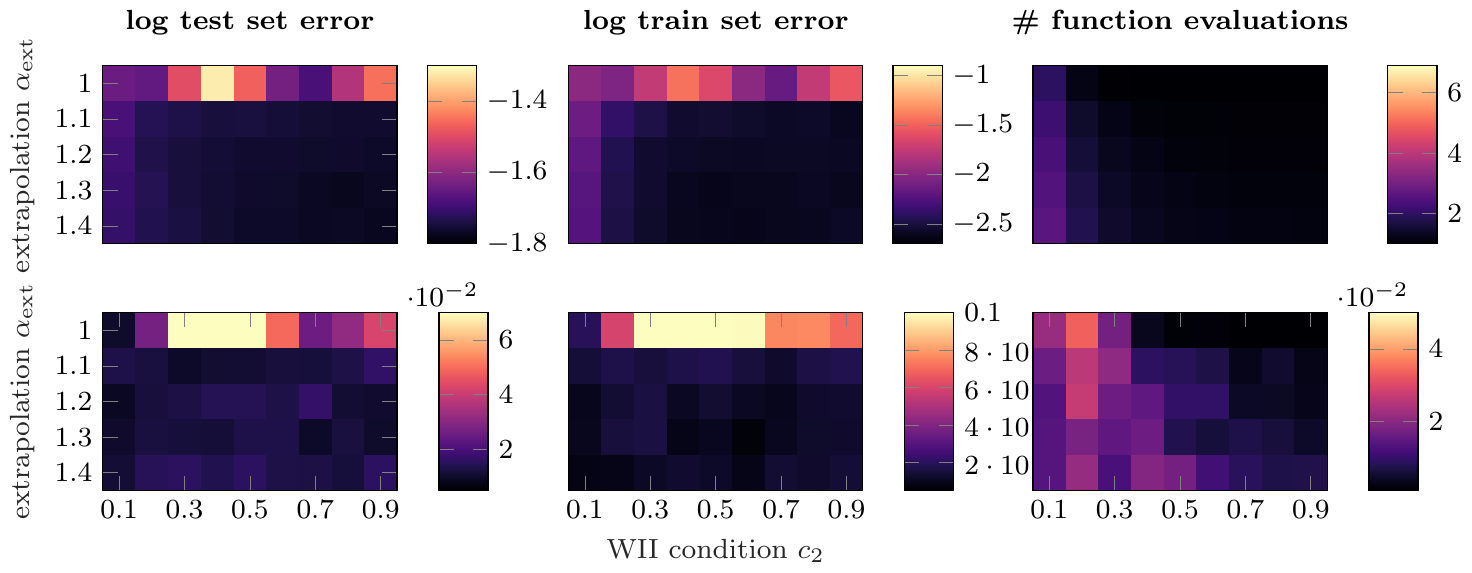}
  \caption{Same as Figure \ref{fig:c2extspace01} but for \emph{fixed} $c_W = 0.10$.}
\label{fig:c2extspace10}
\end{figure}
\begin{figure}[h]
  \centering
  \setlength{\figwidth}{.9\textwidth}
  \setlength{\figheight}{.2\textheight}
  %\tikzset{external/remake next}
  %{\scriptsize \input{fig/MNIST_ARCH_1000_500_250_10_m00200_paraSens_ext_c2_cw20_imagesc.tikz}}  
  \includegraphics[scale=1.0]{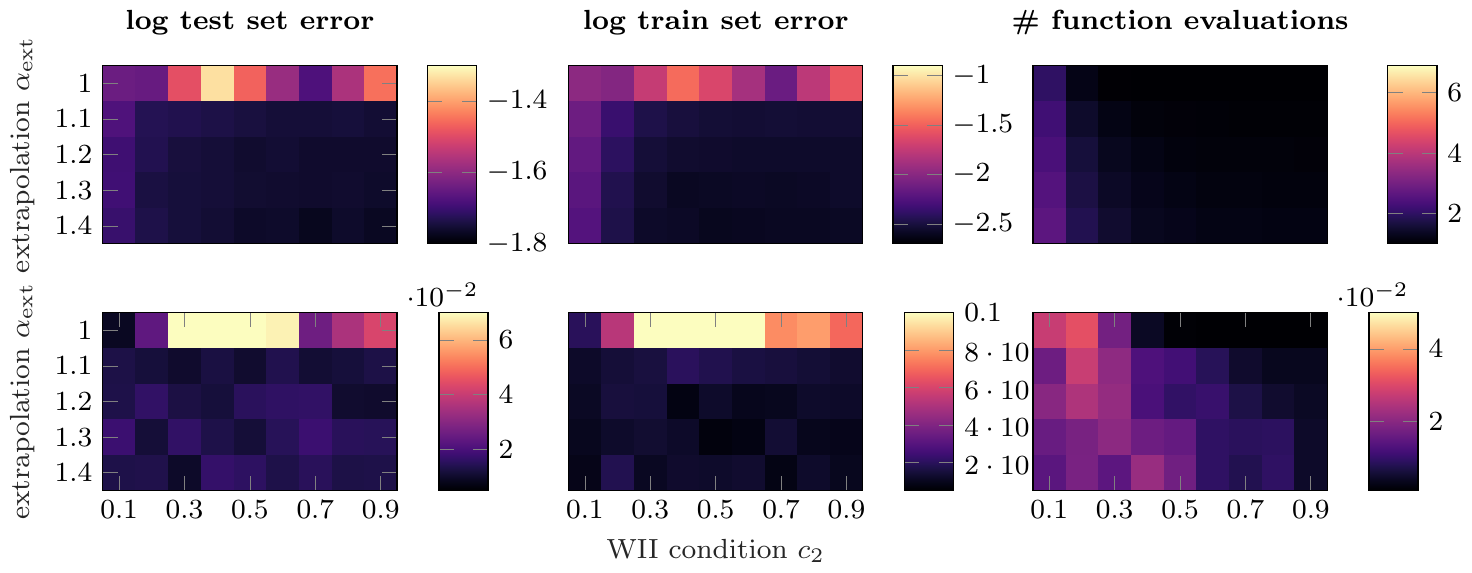}
  \caption{Same as Figure \ref{fig:c2extspace01} but for \emph{fixed} $c_W = 0.20$.}
\label{fig:c2extspace20}
\end{figure}
\begin{figure}[h]
  \centering
  \setlength{\figwidth}{.9\textwidth}
  \setlength{\figheight}{.2\textheight}
  %\tikzset{external/remake next}
  %{\scriptsize \input{fig/MNIST_ARCH_1000_500_250_10_m00200_paraSens_ext_c2_cw30_imagesc.tikz}}  
  \includegraphics[scale=1.0]{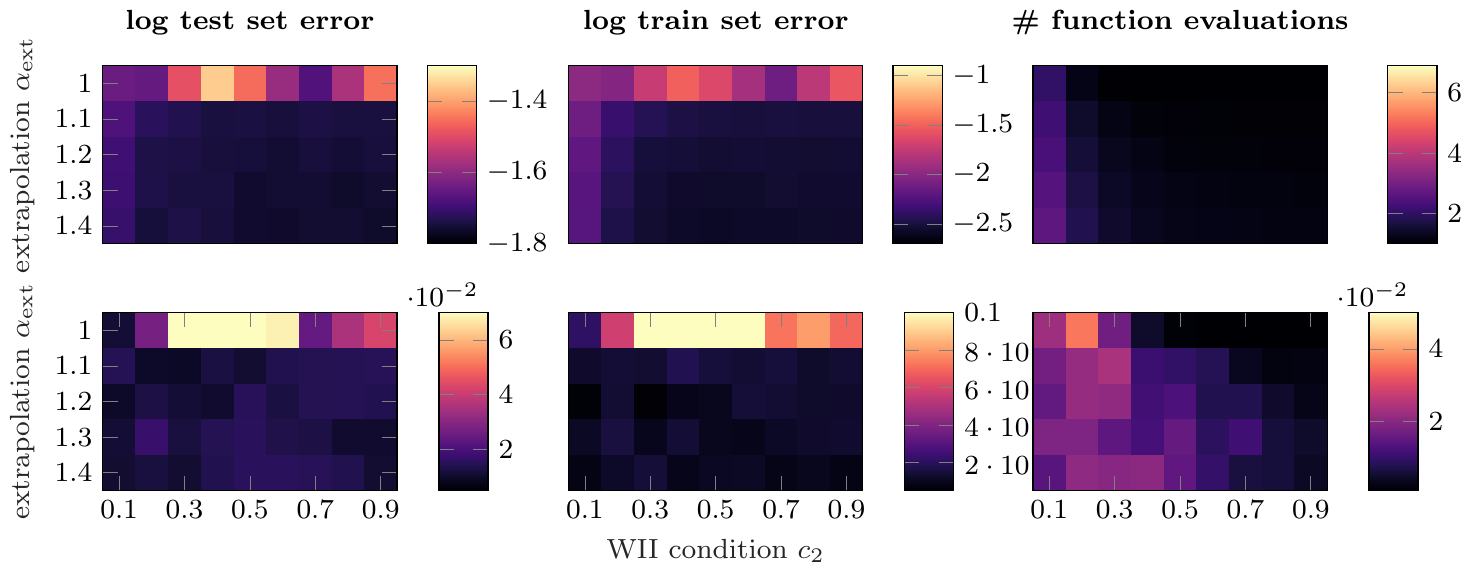}
  \caption{Same as Figure \ref{fig:c2extspace01} but for \emph{fixed} $c_W = 0.30$.}
\label{fig:c2extspace30}
\end{figure}
\begin{figure}[h]
  \centering
  \setlength{\figwidth}{.9\textwidth}
  \setlength{\figheight}{.2\textheight}
  %\tikzset{external/remake next}
  %{\scriptsize \input{fig/MNIST_ARCH_1000_500_250_10_m00200_paraSens_ext_c2_cw40_imagesc.tikz}}  
  \includegraphics[scale=1.0]{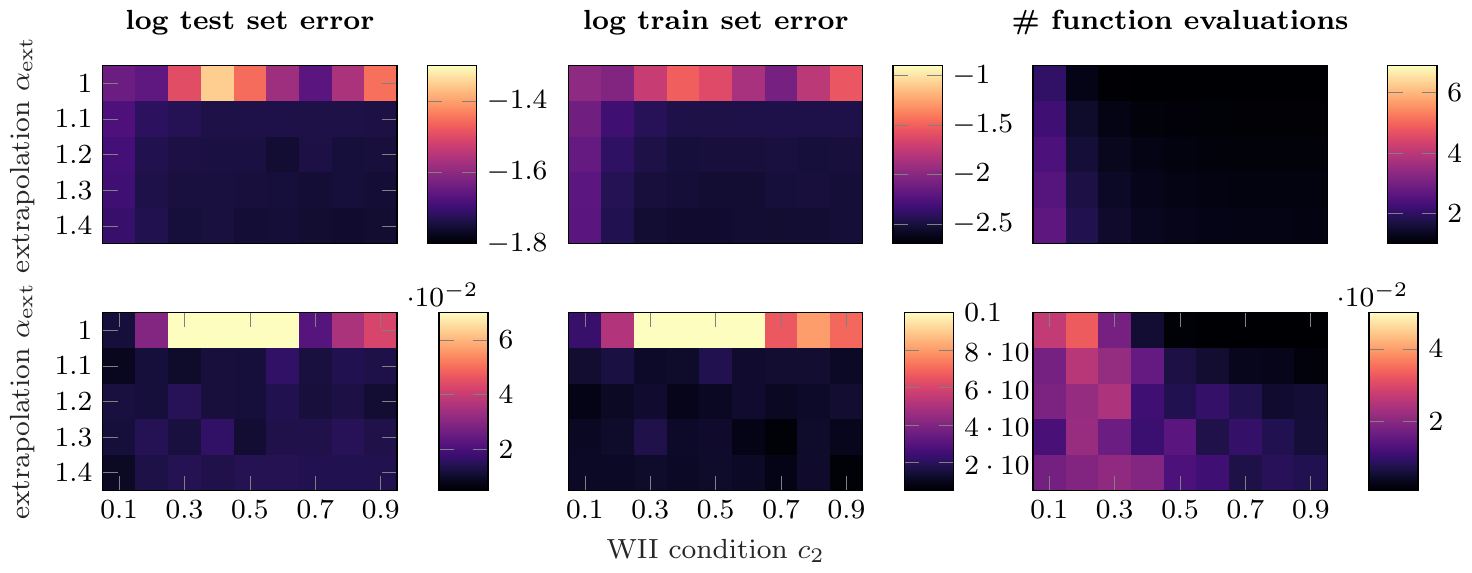}
  \caption{Same as Figure \ref{fig:c2extspace01} but for \emph{fixed} $c_W = 0.40$.}
\label{fig:c2extspace40}
\end{figure}
\begin{figure}[h]
  \centering
  \setlength{\figwidth}{.9\textwidth}
  \setlength{\figheight}{.2\textheight}
  %\tikzset{external/remake next}
  %{\scriptsize \input{fig/MNIST_ARCH_1000_500_250_10_m00200_paraSens_ext_c2_cw50_imagesc.tikz}}  
  \includegraphics[scale=1.0]{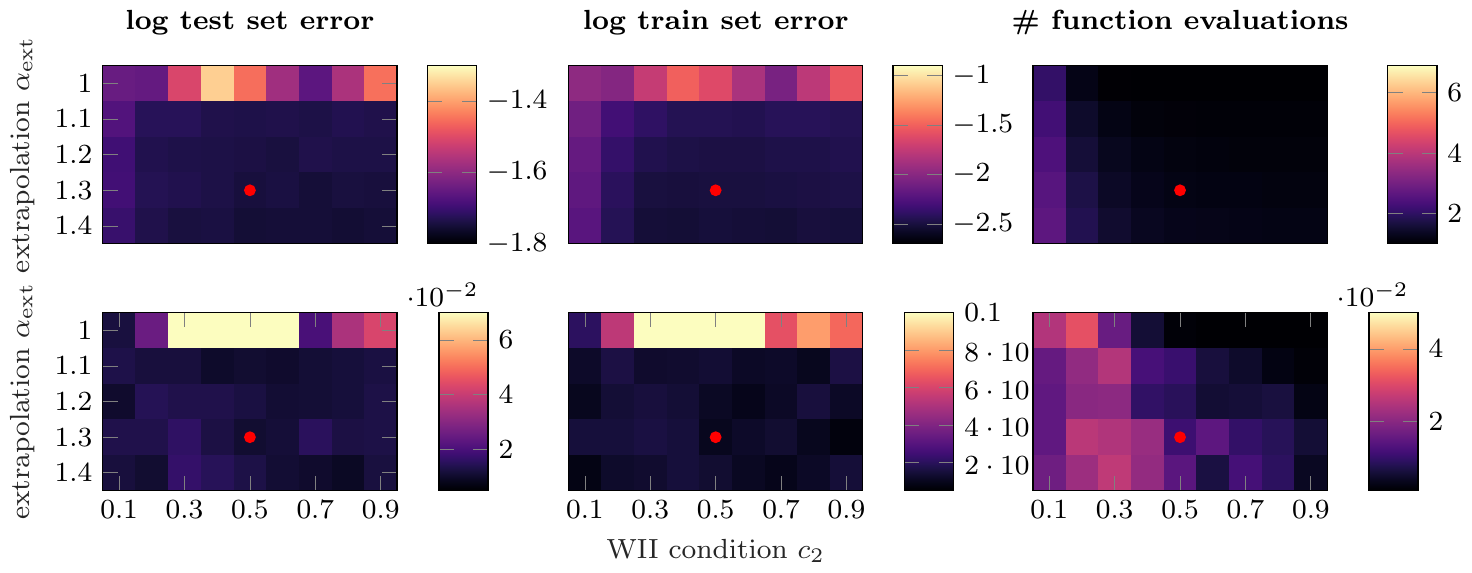}
  \caption{Same as Figure \ref{fig:c2extspace01} but for \emph{fixed} $c_W = 0.50$. The default values as red dots.}
\label{fig:c2extspace50}
\end{figure}
\begin{figure}[h]
  \centering
  \setlength{\figwidth}{.9\textwidth}
  \setlength{\figheight}{.2\textheight}
  %\tikzset{external/remake next}
  %{\scriptsize \input{fig/MNIST_ARCH_1000_500_250_10_m00200_paraSens_ext_c2_cw60_imagesc.tikz}}  
  \includegraphics[scale=1.0]{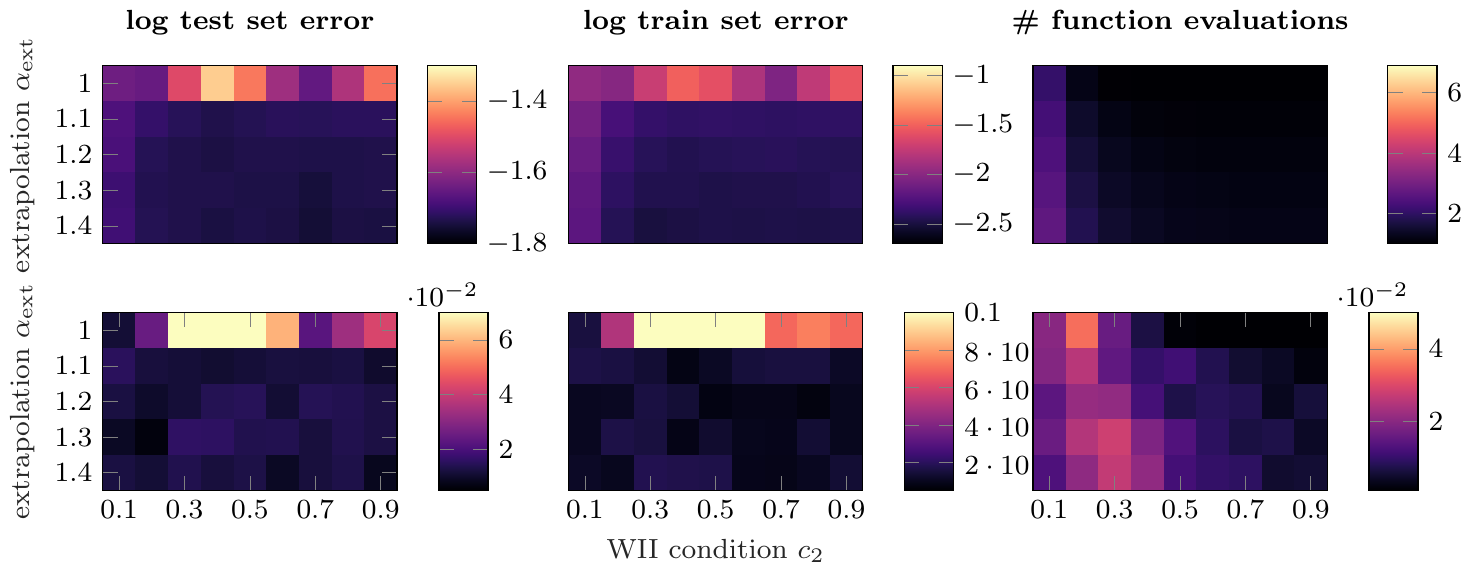}
  \caption{Same as Figure \ref{fig:c2extspace01} but for \emph{fixed} $c_W = 0.60$.}
\label{fig:c2extspace60}
\end{figure}
\begin{figure}[h]
  \centering
  \setlength{\figwidth}{.9\textwidth}
  \setlength{\figheight}{.2\textheight}
  %\tikzset{external/remake next}
  %{\scriptsize \input{fig/MNIST_ARCH_1000_500_250_10_m00200_paraSens_ext_c2_cw70_imagesc.tikz}}    
  \includegraphics[scale=1.0]{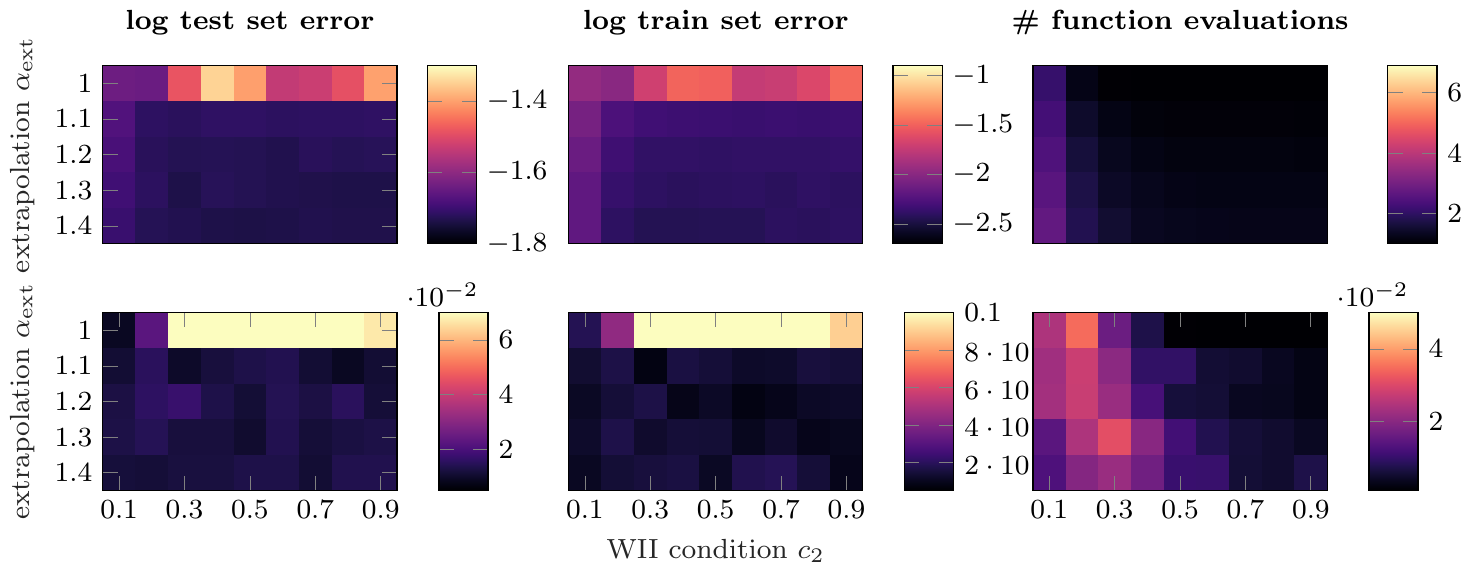}
  \caption{Same as Figure \ref{fig:c2extspace01} but for \emph{fixed} $c_W = 0.70$.}
\label{fig:c2extspace70}
\end{figure}
\begin{figure}[h]
  \centering
  \setlength{\figwidth}{.9\textwidth}
  \setlength{\figheight}{.2\textheight}
  %\tikzset{external/remake next}
  %{\scriptsize \input{fig/MNIST_ARCH_1000_500_250_10_m00200_paraSens_ext_c2_cw80_imagesc.tikz}}  
  \includegraphics[scale=1.0]{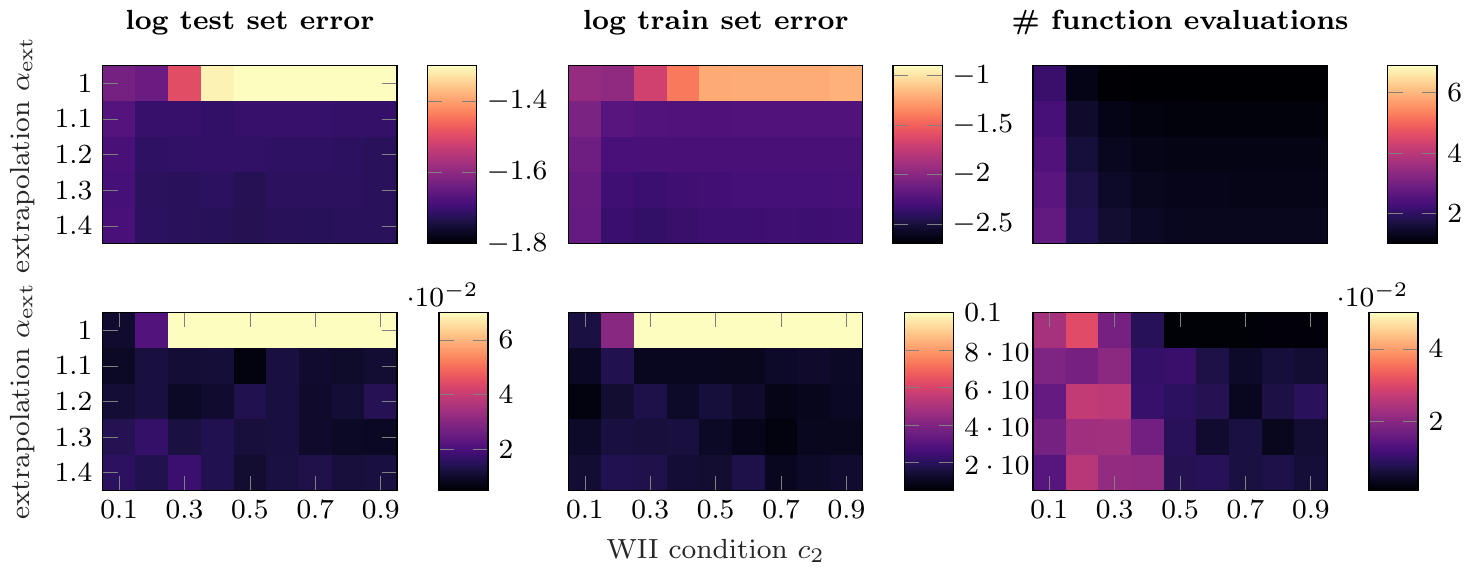}
  \caption{Same as Figure \ref{fig:c2extspace01} but for \emph{fixed} $c_W = 0.80$.}
\label{fig:c2extspace80}
\end{figure}
\begin{figure}[h]
  \centering
  \setlength{\figwidth}{.9\textwidth}
  \setlength{\figheight}{.2\textheight}
  %\tikzset{external/remake next}
  %{\scriptsize \input{fig/MNIST_ARCH_1000_500_250_10_m00200_paraSens_ext_c2_cw90_imagesc.tikz}}  
  \includegraphics[scale=1.0]{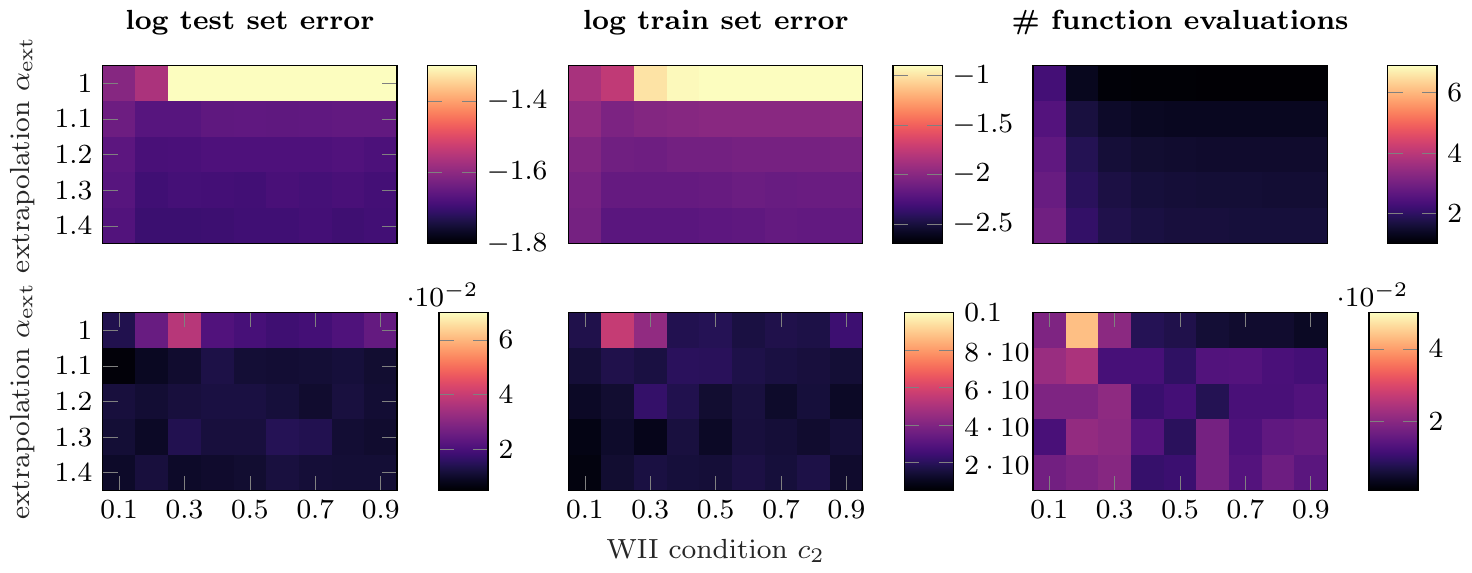}
  \caption{Same as Figure \ref{fig:c2extspace01} but for \emph{fixed} $c_W = 0.90$.}
\label{fig:c2extspace90}
\end{figure}
\begin{figure}[h]
  \centering
  \setlength{\figwidth}{.9\textwidth}
  \setlength{\figheight}{.2\textheight}
  %\tikzset{external/remake next}
  %{\scriptsize \input{fig/MNIST_ARCH_1000_500_250_10_m00200_paraSens_ext_c2_cw99_imagesc.tikz}}  
  \includegraphics[scale=1.0]{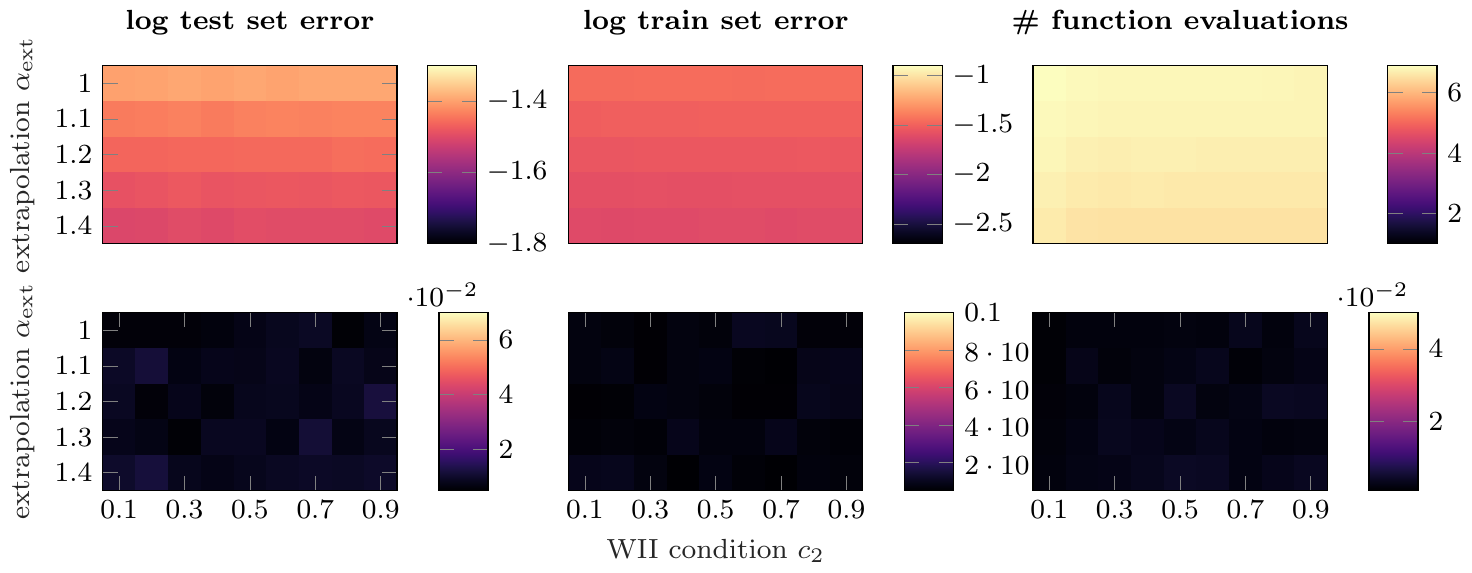}
  \caption{Same as Figure \ref{fig:c2extspace01} but for \emph{fixed} $c_W = 0.99$.}
\label{fig:c2extspace99}
\end{figure}

\clearpage
\section*{Appendix D. -- Pseudocode}
\manuallabel{app:pseudocode}{D}
Algorithm \ref{alg:sketch} of Section \ref{sec:method} roughly sketches the structure of the probabilistic line search and its main ingredients. 
This section provides a detailed pseudocode which can be used for re-implementation. 
It is based on the code which was used for the experiments in this paper. 
A matlab implementation including a minimal example can be found at {\url{http://tinyurl.com/probLineSearch}}. 
The actual line search routine is called \textsc{probLineSearch} below and is quite short. Most of the pseudocode is occupied with comments, helper function that define the kernel of the \gp, the \gp-update or Gauss cdf and pdf which we printed here for completeness such that a detailed re-implementation is possible.
%A TensorFlow \citep{tensorflow2015-whitepaper} implementation under continuous development (and therefore with eventually slightly different design than presented in this work) can be found at \url{https://github.com/ProbabilisticNumerics/probabilistic_line_search}. 
For better readability of the pseudocode we use the following color coding:
\begin{itemize}
\item \blu{blue}: comments
\item \mpg{green}: variables of the integrated Wiener process.
\item \dre{red}: most recently evaluated observation (noisy loss and gradient). If the line search terminates, these will be returned as `accepted'.
\item \ora{orange}: inputs from the main solver procedure and unchanged during each line search.
\end{itemize}
%Variables in \mpg{green} belong to the two-dimensional \gp. 
%Variables in \dre{red} belong to the lastly evaluated observation (noisy loss and gradient). 
%If the line search terminated, these are the data-points that are returned as accepted. 
%Variables in \ora{orange} are input from the solver procedure and unchanged during the line search. %An object oriented implementation could for example have \gp-class with all the \mpg{green} variables and the procedures {\sc updateGP}, {\sc cubicMinumum}, all kernel functions ({\sc m}, {\sc V},\dots)
Notation and operators:
\begin{center}
%\begin{table}
\renewcommand{\arraystretch}{1.1}
\begin{tabular}{l|l}
    operator or function & definition\\ 
\hline
$A\Hm B$& elementwise multiplication \\
$A\Hd B$& elementwise division\\
$ A^{\Hm b}$& elementwise power of $b$\\
$A'$& transpose of $A$\\
$A \cdot B$&  scalar-scalar, scalar-matrix or matrix-matrix multiplication\\
$A\slash B$& right matrix division, the same as $A \cdot B^{-1}$\\
$A\backslash B$& left matrix division, the same as $A^{-1} \cdot B$\\
$\sign(a)$& sign of scalar $a$\\
$\erf(x)$& error function $\erf(x) = \frac{2}{\sqrt{\pi}} \int_{0}^{x} e^{-t^2}dt$\\
$\max(A)$& maximum element in A\\
$\min(A)$&  minimum element in A\\
$|a|$&     absolute value of scalar $a$\\
$A < B$& elementwise `less' comparison\\
$A \leq B$& elementwise `less-or-equal' comparison\\
$A > B$& elementwise `greater' comparison\\
$A \geq B$& elementwise `greater-or-equal' comparison\\
$[a,b,c]\leftarrow f(x)$& function $f$ called at $x$ returns the values $a,b$ and $c$\\
\hline
\end{tabular}
\renewcommand{\arraystretch}{1.0}
\end{center}
\vspace{0.1cm}
For better readability and to avoid confusion with transposes, we denote derivatives for example as $dy$ and $df$ (instead of $y'$ and $f'$ as in the main text).

%\begin{center}
\begin{algorithm}
    \begin{algorithmic}[1]
        \Function{SGDSolver}{$f$}
        \LineComment{$f$ -- function handle to objective. Usage: $[y, dy, \Sigma_{f}, \Sigma_{df}]\gets f(x)$.}
        \State
        \LineComment{initial weights}
        \State $x\gets $initial weights
        \State
        \LineComment{initial step size (rather small to avoid over-shooting in very first step)}
        \State $\alpha \gets e.g.\approx 10^{-4}$
        \State $\alpha_{\text{stats}} \gets \alpha$
        \State
        \LineComment{initial function evaluation at $x$}
        \State $[y, dy, \Sigma_{f}, \Sigma_{df}]  \gets f(x)$ 
        \State
        \LineComment{initial search direction}
        \State $d\gets -dy$ 
%\algstore{myalg}
%\end{algorithmic}
%\end{algorithm}
%
%\begin{algorithm}
%  \begin{algorithmic}
%    \algrestore{myalg}
        \State
        \LineComment{loop over line searches}
        \While{\textbf{budget not used}}
            \State
            \LineComment{line search finds step size}
            \State $[\alpha, \alpha_{\text{stats}}, x, y, dy, \Sigma_{f}, 
            \Sigma_{df}]\gets$\Call{probLineSearch}{$x,d,y,dy, \Sigma_{f},\Sigma_{df},\alpha,\alpha_{\text{stats}}, f$}
            \State
            \LineComment{set new search direction}
            \State $d\gets -dy$ 
        \EndWhile
        \State
        \State\Return \returnCol{$x$}
        \EndFunction
    \end{algorithmic}
%\end{center}
\end{algorithm}

% ===============================================================================
\begin{algorithm}
\begin{algorithmic}[1]
\Function{probLineSearch}{$\globIn{x_0},\globIn{d},\globIn{f_0},\globIn{df_0},\globIn{\Sigma_{f_0}},\globIn{\Sigma_{df_0}},\globIn{\alpha_0}, \alpha_{stats}, f$}
\LineComment{$x_0$ -- current weights $[D\times 1]$}
\LineComment{$f$ -- function handle to objective. }
\LineComment{}
\LineComment{$d$ -- search direction $[D\times 1]$ (does not need to be normalized)}
\LineComment{$f_0$ -- function value at start, $f_0 = f(x_0)$}
\LineComment{$df_0$ -- gradient at start, $df_0 = \nabla f(x_0)$ $[D\times 1]$}
\LineComment{$\Sigma_{f_0}$ -- sample variance of $f_0$}
\LineComment{$\Sigma_{df_0}$ -- sample variances of $df_0$, $[D\times 1]$}
\LineComment{$\alpha_0$ -- initial step size}
\State
\LineComment{set maximum \# of $f$ evaluations per line search}
\State $\limi \gets 6$ 
\algstore{myalg}
\end{algorithmic}
\end{algorithm}

\begin{algorithm}
  \begin{algorithmic}
    \algrestore{myalg}
%\State
\LineComment{scaling and noise level of \gp}
\State $\betaf \gets |\globIn{d}' \cdot \globIn{\Sigma_{df_0}}|$ \ColorComment{scale factor}% \complexity{D}}
\State $\GPCol{\sigma_f} \gets \sqrt{\globIn{\Sigma_{f_0}}} / (\globIn{\alpha_0} \cdot \betaf)$ \ColorComment{scaled sample variance of $f_0$}
\State $\GPCol{\sigma_{df}} \gets \sqrt{((\globIn{d}^{\Hm 2})' \cdot \globIn{\Sigma_{df_0}})} / \betaf$ \ColorComment{scaled and projected sample variances of $df_0$}
\State
\LineComment{initialize counter and non-fixed parameters}
\State $\GPCol{N} \gets 1$ \ColorComment{size of \gp $=2\cdot N$}
\State $t_{\text{ext}} \gets 1$ \ColorComment{scaled step size for extrapolation}
\State $\outsCol{tt} \gets 1$ \ColorComment{scaled position of first function evaluation}
\State
\LineComment{initialize storage for \gp. Dynamic arrays of maximum size $[L+1\times 1]$}
\State $\GPCol{T} \gets [0]$ \ColorComment{scaled positions along search direction}
\State $\GPCol{Y} \gets [0]$ \ColorComment{scaled function values at $T$}
\State $\GPCol{dY} \gets [(\globIn{df_0}' \cdot \globIn{d}) / \betaf]$ \ColorComment{scaled projected gradients at $T$}%, \complexity{D}}
\State 
\LineComment{initialize \gp~with observation at start}
\State $[\GPCol{G}, \GPCol{A}] \gets $\Call{updateGP}{$\GPCol{T}, \GPCol{Y}, \GPCol{dY}, \GPCol{N}, \GPCol{\sigma_f}, \GPCol{\sigma_{df}}$}
    \State
    \LineComment{loop until budged is used or acceptable point is found}
    \For{\textbf{$\GPCol{N}$ from $2$ to $\limi + 1$}} 
        \State
        \LineComment{evaluate objective function at $tt$.}
        \State $[\outsCol{y}, \outsCol{dy}, \outsCol{\Sigma_{f}}, \outsCol{\Sigma_{df}}, \GPCol{T}, \GPCol{Y}, \GPCol{dY}, \GPCol{N}]
        \gets$\Call{evaluateObjective}{$\outsCol{tt}, \globIn{x_0}, \globIn{\alpha_0}, \globIn{d}, \GPCol{T}, \GPCol{Y}, \GPCol{dY}, \GPCol{N}, \betaf, f$}
        \State
        \LineComment{update the \gp~which is now of size $2\cdot N$.}
        \State $[\GPCol{G}, \GPCol{A}] \gets $\Call{updateGP}{$\GPCol{T}, \GPCol{Y},  \GPCol{dY} , \GPCol{N}, \GPCol{\sigma_f}, \GPCol{\sigma_{df}}$}
        \State
        \LineComment{initialize storage for candidates. Dynamic arrays of maximum size $[N\times 1]$.}
        \State $T_{\text{cand}} \gets [~]$ \ColorComment{scaled position of candidates}
        \State $M_{\text{cand}} \gets [~]$ \ColorComment{\gp~ mean of candidates}
        \State $S_{\text{cand}} \gets [~]$ \ColorComment{\gp~standard deviation of candidates}
        \State
        \LineComment{current point is above the Wolfe threshold? If yes, accept point and return.}
        \If{\Call{probWolfe}{$\outsCol{tt}, \GPCol{T}, \GPCol{A}, \GPCol{G}$}}
            \State $output\gets$\Call{rescaleOutput}{$\globIn{x_0}, \globIn{f_0}, 
              \globIn{\alpha_0}, \globIn{d}, \outsCol{tt}, \outsCol{y}, \outsCol{dy}, \outsCol{\Sigma_{f}}, \outsCol{\Sigma_{df}}, \betaf$}
            \State\Return \returnCol{$output$}
        \EndIf
        \State
        \LineComment{Wolfe conditions not satisfied at this point.}
        \LineComment{find suitable candidates for next evaluation.}
        \State
        \LineComment{\gp~mean of function values and corresponding gradients at points in $\GPCol{T}$.}
        \State $M \gets$ map function \Call{m}{$\_,\GPCol{T},\GPCol{A}$} over $\GPCol{T}$ 
        \State $dM \gets$ map function \Call{d1m}{$\_,\GPCol{T},\GPCol{A}$} over $\GPCol{T}$ 
\algstore{myalg}
\end{algorithmic}
\end{algorithm}

\begin{algorithm}
  \begin{algorithmic}
    \algrestore{myalg}
%        \State
        \LineComment{candidates 1: local minima of \gp~mean.}
        \State $T_{\text{sorted}} \gets$ sort $\GPCol{T}$ in ascending order
        \State $T_{\text{Wolfes}} \gets [~]$  \ColorComment{prepare list of acceptable points}
        \State
        \LineComment{iterate through all $N - 1$ cells, compute locations of local minima.}
        \For{$n$ \textbf{from} $1$ \textbf{to} $\GPCol{N}-1$}
            \State $T_n \gets$ value of $T_{\text{sorted}}$ at $n$
            \State $T_{n+1} \gets$ value of $T_{\text{sorted}}$ at $n+1$
            \State
            \LineComment{add a little offset for numerical stability}
            \State $t_{\text{rep}} \gets T_{n} + 10^{-6} \cdot (T_{n+1} - T_{n})$ 
            \State
            \LineComment{compute location of cubic minimum in $n^{\text{\tiny th}}$ cell}
            \State $t_{\text{cubMin}} \gets $\Call{cubicMinimum}{$t_{\text{rep}},\GPCol{T},\GPCol{A},\GPCol{N}$}
            \State
            \LineComment{add point to candidate list if minimum lies in between $T_n$ and $T_{n+1}$}
            \If{$t_{\text{cubMin}} > T_{n}$ \textbf{and} $t_{\text{cubMin}} < T_{n+1}$}
                \If{(not isnanOrIsinf($t_{\text{cubMin}}$)) and ($t_{\text{cubMin}}>0$)}
                \State $T_{\text{cand}} \gets$ append $t_{\text{cubMin}}$ 
                \State $M_{\text{cand}} \gets$ append \Call{m}{$t_{\text{cubMin}},\GPCol{T},\GPCol{A}$} 
                \State $S_{\text{cand}} \gets$ append \Call{V}{$t_{\text{cubMin}},\GPCol{T},\GPCol{G}$} 
                \EndIf
            \Else
            \State
            %else from previous page continues here
            \LineComment{most likely uphill? If yes, break.}
            \If{$n = 1$ \textbf{and} \Call{d1m}{$0,\GPCol{T},\GPCol{A}$} $ > 0$}
                 \State $r \gets 0.01$
                \State $\outsCol{tt} \gets$  $r \cdot (T_{n} + T_{n+1})$ 
                \State
                \LineComment{evaluate objective function at $tt$ and return.}
                \State $[\outsCol{y}, \outsCol{dy}, \outsCol{\Sigma_{f}}, \outsCol{\Sigma_{df}}, \GPCol{T}, \GPCol{Y}, \GPCol{dY}, \GPCol{N}]
                \gets$\Call{evaluateObjective}{$\outsCol{tt}, \globIn{x_0}, \globIn{\alpha_0}, \globIn{d}, \GPCol{T}, \GPCol{Y}, \GPCol{dY}, \GPCol{N}, \betaf, f$}
                \State
            \State $output\gets$\Call{rescaleOutput}{$\globIn{x_0}, \globIn{f_0}, 
              \globIn{\alpha_0}, \globIn{d}, \outsCol{tt}, \outsCol{y}, \outsCol{dy}, \outsCol{\Sigma_{f}}, \outsCol{\Sigma_{df}}, \betaf$}
            \State\Return \returnCol{$output$}
            \EndIf
        \EndIf
        \State
        \LineComment{check whether there is an acceptable point among the old evaluations}
        \If{$n > 1$ \textbf{and} \Call{probWolfe}{$T_{n},\GPCol{T},\GPCol{A},\GPCol{G}$}}
            \State $T_{\text{Wolfes}} \gets$ append $T_{n}$ 
        \EndIf
    \EndFor
    \State
        \LineComment{check if acceptable points exists and return}
        \If{$T_{\text{Wolfes}}$ \textbf{is not empty}}
\algstore{myalg}
\end{algorithmic}
\end{algorithm}

\begin{algorithm}
  \begin{algorithmic}
    \algrestore{myalg}
%           \State
           \LineComment{if last evaluated point is among acceptable ones, return it.}
           \If{$tt\textbf{ in } T_{\text{Wolfes}}$}
               \State $output\gets$\Call{rescaleOutput}{$\globIn{x_0}, \globIn{f_0}, 
                 \globIn{\alpha_0}, \globIn{d}, \outsCol{tt}, \outsCol{y}, \outsCol{dy}, \outsCol{\Sigma_{f}}, \outsCol{\Sigma_{df}}, \betaf$}
               \State\Return \returnCol{$output$}
           \EndIf
            \State
           \LineComment{else, choose the one with the lowest \gp~mean and re-evaluate its gradient.}
%           \LineComment{compute means of acceptable points and find position of lowest.}
           \State $M_{\text{Wolfes}} \gets$ map \Call{m}{$\_,\GPCol{T},\GPCol{A}$} over $T_{\text{Wolfes}}$
           \State $\outsCol{tt} \gets$ value of $T_{\text{Wolfes}}$ at index of $\min(M_{\text{Wolfes}})$ 
           \State
           \LineComment{evaluate objective function at $tt$.}
           \State $[\outsCol{y}, \outsCol{dy}, \outsCol{\Sigma_{f}}, \outsCol{\Sigma_{df}}, \GPCol{T}, \GPCol{Y}, \GPCol{dY}, \GPCol{N}]
           \gets$\Call{evaluateObjective}{$\outsCol{tt}, \globIn{x_0}, \globIn{\alpha_0}, \globIn{d}, \GPCol{T}, \GPCol{Y}, \GPCol{dY}, \GPCol{N}, \betaf, f$}
           \State
           \State $output\gets$\Call{rescaleOutput}{$\globIn{x_0}, \globIn{f_0}, 
            \globIn{\alpha_0}, \globIn{d}, \outsCol{tt}, \outsCol{y}, \outsCol{dy}, \outsCol{\Sigma_{f}}, \outsCol{\Sigma_{df}}, \betaf$}
            \State\Return \returnCol{$output$}
    \EndIf
    \State
    \LineComment{candidates 2: one extrapolation step}
    \State $T_{\text{cand}} \gets$ append $\max(\GPCol{T}) + t_{\text{ext}}$
    \State $M_{\text{cand}} \gets$ append \Call{m}{$\max(\GPCol{T}) + t_{\text{ext}},\GPCol{T},\GPCol{A}$}
    \State $S_{\text{cand}} \gets$ append \Call{V}{$\max(\GPCol{T}) + t_{\text{ext}},\GPCol{T},\GPCol{G}$}$^{\frac{1}{2}}$
    \State
    \LineComment{find minimal mean among $M$.}
    %\LineComment{used for \Call{expected\_improvement}{$\cdot$}}
    \State $\mu_{\text{EI}} \gets$ minimal value of $M$ 
    \State
    \LineComment{compute expected improvement and Wolfe probabilities at $T_{\text{cand}}$}
    \State $EI_{\text{cand}} \gets $\Call{expectedImprovement}{$M_{\text{cand}},S_{\text{cand}},\mu_{\text{EI}}$}
    \State $PW_{\text{cand}} \gets$ map \Call{probWolfe}{$\_,\GPCol{T},\GPCol{A},\GPCol{G}$} over $T_{\text{cand}}$
    \State
    \LineComment{choose point among candidates that maximizes $EI_{\text{cand}}\land PW_{\text{cand}}$ }
    \State $i_{\text{bestCand}} \gets \text{index of }\max(EI_{\text{cand}} \Hm PW_{\text{cand}})$
    \State $tt_{\text{bestCand}} \gets$ value of $T_{\text{cand}}$ at $i_{\text{bestCand}}$
    \State
    \LineComment{extend extrapolation step if necessary}
    \If{$tt_{\text{bestCand}}$ \textbf{is equal to} $\outsCol{tt} + t_{\text{ext}}$}
        \State $t_{\text{ext}} \gets 2 \cdot t_{\text{ext}}$
    \EndIf
    \State
    \LineComment{set location for next evaluation}
    \State $\outsCol{tt} \gets tt_{\text{bestCand}}$
    \EndFor 
        \State
%        \LineComment{evaluation limit is reached.}
        \LineComment{limit reached: evaluate a final time and return the point with lowest \gp~mean}
        \State $[\outsCol{y}, \outsCol{dy}, \outsCol{\Sigma_{f}}, \outsCol{\Sigma_{df}}, \GPCol{T}, \GPCol{Y}, \GPCol{dY}, \GPCol{N}]
        \gets$\Call{evaluateObjective}{$\outsCol{tt}, \globIn{x_0}, \globIn{\alpha_0}, \globIn{d}, \GPCol{T}, \GPCol{Y}, \GPCol{dY}, \GPCol{N}, \betaf, f$}
\algstore{myalg}
\end{algorithmic}
\end{algorithm}

\begin{algorithm}
  \begin{algorithmic}
    \algrestore{myalg}
%        \State
        \LineComment{update the \gp~which is now of size $2\cdot N$.}
        \State $[\GPCol{G}, \GPCol{A}] \gets $\Call{updateGP}{$\GPCol{T},\GPCol{Y}, \GPCol{dY} ,\GPCol{N}, \GPCol{\sigma_f}, \GPCol{\sigma_{df}}$}
        \State
        \LineComment{check last point for acceptance}
        \If{\Call{probWolfe}{$\outsCol{tt},\GPCol{T},\GPCol{A},\GPCol{G}$}}
            \State $output\gets$\Call{rescaleOutput}{$\globIn{x_0}, \globIn{f_0}, 
              \globIn{\alpha_0}, \globIn{d}, \outsCol{tt}, \outsCol{y}, \outsCol{dy}, \outsCol{\Sigma_{f}}, \outsCol{\Sigma_{df}}, \betaf$}
            \State\Return \returnCol{$output$}
        \EndIf
        \State
        \LineComment{at the end of budget return point with the lowest \gp~mean}
        \LineComment{compute \gp~means at $T$}
        \State $M \gets$ map \Call{m}{$\_,\GPCol{T},\GPCol{A}$} over $\GPCol{T}$ 
        \State $ i_{\text{lowest}} \gets$ index of minimal value in $M$ 
        \State $t_{\text{lowest}} \gets$ value of $\GPCol{T}$ at $ i_{\text{lowest}}$ 
        \State
        \LineComment{if $t_{\text{lowest}}$ is the last evaluated point, return}
        \If {$t_{\text{lowest}} \textbf{ is equal to }tt$}
            \State $output\gets$\Call{rescaleOutput}{$\globIn{x_0}, \globIn{f_0}, 
              \globIn{\alpha_0}, \globIn{d}, \outsCol{tt}, \outsCol{y}, \outsCol{dy}, \outsCol{\Sigma_{f}}, \outsCol{\Sigma_{df}}, \betaf$}
            \State\Return \returnCol{$output$}
        \EndIf
        \State
        \LineComment{else, re-evaluate its gradient and return}
           \State $\outsCol{tt} \gets$ value of $t_{\text{lowest}}$
           \State
           \LineComment{evaluate objective function at $tt$.}
           \State $[\outsCol{y}, \outsCol{dy}, \outsCol{\Sigma_{f}}, \outsCol{\Sigma_{df}}, \GPCol{T}, \GPCol{Y}, \GPCol{dY}, \GPCol{N}]
           \gets$\Call{evaluateObjective}{$\outsCol{tt}, \globIn{x_0}, \globIn{\alpha_0}, \globIn{d}, \GPCol{T}, \GPCol{Y}, \GPCol{dY}, \GPCol{N}, \betaf, f$}
           \State
            \State $output\gets$\Call{rescaleOutput}{$\globIn{x_0}, \globIn{f_0}, 
              \globIn{\alpha_0}, \globIn{d}, \outsCol{tt}, \outsCol{y}, \outsCol{dy}, \outsCol{\Sigma_{f}}, \outsCol{\Sigma_{df}}, \betaf$}
            \State\Return \returnCol{$output$}
        \EndFunction
    \end{algorithmic}
\end{algorithm}

\begin{algorithm}
    \begin{algorithmic}[1]
        \Function{rescaleOutput}{$\globIn{x_0}, \globIn{f_0}, \globIn{\alpha_0}, 
          \globIn{d}, \outsCol{tt}, \outsCol{y}, \outsCol{dy}, \outsCol{\Sigma_{f}}, \outsCol{\Sigma_{df}}, \betaf, \alpha_{\text{stats}}$}
        \LineComment{design parameters}
        \State $\alphaextr \gets 1.3$ \ColorComment{extrapolation parameter}
        \State $\fres \gets 100$ \ColorComment{reset threshold for \gp~scale}
        \State
        \LineComment{rescale accepted step size}
        \State $\alpha_{\text{acc}} \gets \outsCol{tt} \cdot \globIn{\alpha_0}$ 
        \State
        \LineComment{update weights}
        \State $x_{\text{acc}} \gets \globIn{x_0} + \alpha_{\text{acc}} \cdot \globIn{d}$ 
        \State
        \LineComment{rescale accepted function value}
        \State $f_{\text{acc}} \gets \outsCol{y} \cdot (\globIn{\alpha_0} \cdot \betaf) + \globIn{f_0}$
\algstore{myalg}
\end{algorithmic}
\end{algorithm}

\begin{algorithm}
  \begin{algorithmic}
    \algrestore{myalg}
%        \State
        \LineComment{accepted gradient}
        \State $df_{\text{acc}} \gets \outsCol{dy}$ 
        \State
        \LineComment{sample variance of $f_{\text{acc}}$}
        \State $\Sigma_{f_{\text{acc}}} \gets \outsCol{\Sigma_{f}}$ 
        \State
        \LineComment{sample variances of $df_{\text{acc}}$}
        \State $\Sigma_{df_{\text{acc}}} \gets \outsCol{\Sigma_{df}}$ 
        \State
        \LineComment{update exponential running average of scalings}
        \State $\gamma \gets 0.95$
        \State $\alpha_{\text{stats}} \gets  \gamma\cdot\alpha_{\text{stats}} +(1-\gamma)\cdot \alpha_{\text{acc}}$ 
        \State
        \LineComment{next initial step size}
        \State $\alpha_{\text{next}} \gets \alpha_{\text{acc}} \cdot \alphaextr$ 
        \State
        \LineComment{if new \gp~scaling is drastically different than previous ones reset it.}
        \If {$(\alpha_{\text{next}} < \alpha_{\text{stats}}/\fres)$ \textbf{or} $(\alpha_{\text{next}} > \alpha_{\text{stats}}\cdot\fres)$}
           \State $\alpha_{\text{next}} \gets \alpha_{\text{stats}}$
        \EndIf
        \State
        \LineComment{compressed output for readability of pseudocode}
        \State $output\gets [\alpha_{\text{next}}, \alpha_{\text{stats}}, x_{\text{acc}}, f_{\text{acc}}, df_{\text{acc}}, \Sigma_{f_{\text{acc}}}, \Sigma_{df_{\text{acc}}}]$
        \State
        \State\Return \returnCol{$output$}
        \EndFunction
    \end{algorithmic}
\end{algorithm}

\begin{algorithm}
    \begin{algorithmic}[1]
        \Function{evaluateObjective}{$\outsCol{tt}, \globIn{x_0}, \globIn{\alpha_0}, \globIn{d}, \GPCol{T}, \GPCol{Y}, \GPCol{dY}, \GPCol{N}, \betaf, f$}
        \LineComment{evaluate objective function at $tt$}
    	\State $[\outsCol{y}, \outsCol{dy}, \outsCol{\Sigma_{f}}, \outsCol{\Sigma_{df}}] 
        \gets f(\globIn{x_0}  + \outsCol{tt} \cdot \globIn{\alpha_0} \cdot \globIn{d})$ 
        \State
        \LineComment{scale output}
    	\State $\outsCol{y} \gets (\outsCol{y} - \globIn{f_0}) / (\globIn{\alpha_0} \cdot \betaf)$  
    	\State $ dy  \gets (\outsCol{dy}'\cdot \globIn{d}) / \betaf$ %\ColorComment{\complexity{D}}
        \State
        \LineComment{storage}
    	\State $\GPCol{T} \gets$ append $\outsCol{tt}$
    	\State $\GPCol{Y} \gets$ append $\outsCol{y}$ 
        \State $\GPCol{dY}  \gets$ append $ dy $ 
        \State $\GPCol{N}\gets \GPCol{N}+1$
        \State
        \State\Return \returnCol{$[y, dy, \Sigma_{f}, \Sigma_{df}, T, Y, dY, N]$}
        \EndFunction
    \end{algorithmic}
\end{algorithm}

\begin{algorithm}
    \begin{algorithmic}[1]
        \Function{cubicMinimum}{$t,\GPCol{T},\GPCol{A},\GPCol{N}$}
            \LineComment{compute necessary derivatives of \gp~ mean at $t$}
            \State $d1m_t \gets $\Call{d1m}{$t,\GPCol{T},\GPCol{A}$}
            \State $d2m_t \gets $\Call{d2m}{$t,\GPCol{T},\GPCol{A}$}
            \State $d3m_t \gets $\Call{d3m}{$t,\GPCol{T},\GPCol{A},\GPCol{N}$}
            \State $a \gets 0.5\cdot d3m_t$
            \State $b \gets d2m_t - t~\cdot d3m_t$
            \State $c \gets d1m_t - d2m_t \cdot~t + 0.5\cdot d3m_t \cdot t^2$
            \State
            \LineComment{third derivative is almost zero $\rightarrow$ essentially a quadratic, single extremum}
            \If{$|d3m_t| < 1^{-9}$} 
                \State $t_{\text{cubMin}} \gets -(d1m_t - t \cdot d2m_t)/d2m_t$
                \State\Return \returnCol{$t_{\text{cubMin}}$}
            \EndIf
            \State
            \LineComment{roots are complex, no extremum}
            \State $\lambda \gets b^2 - 4 \cdot a \cdot c$
            \If{$\lambda < 0$}
                \State $t_{\text{cubMin}} \gets +\infty$
                \State\Return\returnCol{$t_{\text{cubMin}}$}
            \EndIf
            \State
            \LineComment{compute the two possible roots}
            \State $LR \gets (-b - \sign(a)\cdot \sqrt{\lambda})/(2 \cdot a)$ \ColorComment{left root}
            \State $RR \gets (-b + \sign(a)\cdot \sqrt{\lambda})/(2 \cdot a)$ \ColorComment{right root}
            \State
            \LineComment{calculate the two values of the cubic at those points (up to a constant)}
            \State $dt_L \gets LR - t$ \ColorComment{distance to left root}
            \State $dt_R \gets RR - t$ \ColorComment{distance to right root}
            \State $CV_L \gets d1m_t \cdot dt_L + 0.5 \cdot d2m_t \cdot dt_L^2+ (d3m_t \cdot dt_L ^3)/6$ \ColorComment{left cubic value}
            \State $CV_R \gets d1m_t \cdot dt_R + 0.5 \cdot d2m_t \cdot dt_R^2+ (d3m_t \cdot dt_R ^3)/6$ \ColorComment{right cubic value}
            \State
            \LineComment{find the minimum and return it.}
            \If{$CV_L < CV_R$}
                \State $t_{\text{cubMin}} \gets LR$
            \Else
                \State $t_{\text{cubMin}} \gets RR$
            \EndIf
            \State
            \State \Return \returnCol{$t_{\text{cubMin}}$}
            \State
        \EndFunction
    \end{algorithmic}
\end{algorithm}

\begin{algorithm}
    \begin{algorithmic}[1]
        \Function{updateGP}{$\GPCol{T},\GPCol{Y},\GPCol{dY},\GPCol{N},\GPCol{\sigma_f},\GPCol{\sigma_{df}}$}
        \LineComment{initialize kernel matrices}
        \State $k_{TT} \gets [N\times N]$ matrix with zeros \ColorComment{covariance of function values}
        \State $kd_{TT} \gets [N\times N]$ matrix with zeros \ColorComment{covariance of function values and gradients}
        \State $dkd_{TT} \gets [N\times N]$ matrix with zeros \ColorComment{covariance of gradients}
        \State
        \LineComment{fill kernel matrices}
        \For{$i = 1$ to $\GPCol{N}$}
            \For{$j = 1$ to $\GPCol{N}$}
                \State $k_{TT}(i, j) \gets $\Call{k}{$\GPCol{T}(i), \GPCol{T}(j)$}
                \State $kd_{TT}(i, j) \gets $\Call{kd}{$\GPCol{T}(i), \GPCol{T}(j)$}
                \State $dkd_{TT}(i, j) \gets $\Call{dkd}{$\GPCol{T}(i), \GPCol{T}(j)$}
            \EndFor
        \EndFor
        \State
        \LineComment{build diagonal covariance matrix of Gaussian likelihood $[2N\times 2N]$.}
        \State $\Lambda \gets 
                 \begin{bmatrix}
                  \diag (\GPCol{\sigma_f}^2)_{N\times N} & 0_{N\times N}\\
                  0_{N\times N}&\diag (\GPCol{\sigma_{df}}^2)_{N\times N}
                \end{bmatrix}$
        \State
        \State $\GPCol{G} \gets$
        $\left(\begin{array}{cc}
            k_{TT} & kd_{TT} \\
            kd_{TT}' & dkd_{TT}            
        \end{array}\right) + \Lambda$
        \ColorComment{$[2N \times 2N]$ matrix}
        \State
        \LineComment{residual between observed and predicted data}
        \State $\Delta \gets$
        $\left(\begin{array}{c}
            \GPCol{Y} \\
            \GPCol{dY}
        \end{array}\right)$\ColorComment{$[2N\times 1]$ vector}
        \State
        \LineComment{compute weighted observations $A$.}
        \State $\GPCol{A} \gets \GPCol{G} \backslash \Delta $ \ColorComment{$[2N\times 1]$ vector}
        \State
        \State \Return \returnCol{$[G, A]$}
        \State
        \EndFunction
    \end{algorithmic}
\end{algorithm}

\begin{algorithm}
    \begin{algorithmic}[1]
        \Function{m}{$t,\GPCol{T},\GPCol{A}$}
            \LineComment{posterior mean at $t$}
            \State \Return $[$\Call{k}{$t,\GPCol{T}'$}$,~$\Call{kd}{$t,\GPCol{T}'$}$] \cdot \GPCol{A}$
        \EndFunction
        \State
        \Function{d1m}{$t,\GPCol{T},\GPCol{A}$}
            \LineComment{first derivative of mean at $t$} 
            \State \Return $[$\Call{dk}{$t,\GPCol{T}'$}$,~$\Call{dkd}{$t,\GPCol{T}'$}$] \cdot \GPCol{A}$
        \EndFunction
        \State
        \Function{d2m}{$t,\GPCol{T},\GPCol{A}$}
            \LineComment{second derivative of mean at $t$}
            \State \Return $[$\Call{ddk}{$t,\GPCol{T}'$}$,~$\Call{ddkd}{$t,\GPCol{T}'$}$] \cdot \GPCol{A}$
        \EndFunction
        \State
        \Function{d3m}{$t,\GPCol{T},\GPCol{A},N$}
            \LineComment{third derivative of mean at $t$}
            \State \Return $[$\Call{dddk}{$t,\GPCol{T}'$}$,~$\Call{zeros}{1, N}$] \cdot \GPCol{A}$
        \EndFunction
        \State
        \Function{V}{$t,\GPCol{T},\GPCol{G}$}
            \LineComment{posterior variance of function values at $t$}
            \State\Return \Call{k}{$t,t$}$ - [$\Call{k}{$(t,\GPCol{T}'$}$, ~$\Call{kd}{$t,\GPCol{T}'$}$] \cdot (\GPCol{G} \backslash [$\Call{k}{$t,\GPCol{T}'$}$,~$\Call{kd}{$t, \GPCol{T}'$}$]')$
        \EndFunction
        \State
        \Function{Vd}{$t,\GPCol{T},\GPCol{G}$}
            \LineComment{posterior variance of function values and derivatives at $t$}
            \State\Return \Call{kd}{$t,t$}$- [$\Call{k}{$t,\GPCol{T}'$}$ , ~$\Call{kd}{$t,\GPCol{T}'$}$] \cdot (\GPCol{G} \backslash [$\Call{dk}{$t,\GPCol{T}'$}$,~$\Call{dkd}{$t, \GPCol{T}'$}$]')$
        \EndFunction
        \State
        \Function{dVd}{$t,\GPCol{T},\GPCol{G}$}
            \LineComment{posterior variance of derivatives at $t$}
            \State\Return \Call{dkd}{$t,t$}$ - [$\Call{dk}{$t,\GPCol{T}'$}$ , ~$\Call{dkd}{$t,\GPCol{T}'$}$] \cdot (\GPCol{G} \backslash [$\Call{dk}{$t,\GPCol{T}'$}$,~$\Call{dkd}{$t, \GPCol{T}'$}$]')$
        \EndFunction
        \State
        \Function{V0f}{$t,\GPCol{T},\GPCol{G}$}
            \LineComment{posterior covariances of function values at $t=0$ and $t$}
            \State\Return \Call{k}{$0, t$}$  - [$\Call{k}{$0, \GPCol{T}'$}$,~$\Call{kd}{$0, \GPCol{T}'$}$] \cdot (\GPCol{G}\backslash[$\Call{k}{$t, \GPCol{T}'$}$ ,~$\Call{kd}{$t, \GPCol{T}'$}$]')$
        \EndFunction
        \State
        \Function{Vd0f}{$t,\GPCol{T},\GPCol{G}$}
            \LineComment{posterior covariance of gradient and function value at $t=0$ and $t$ respectively}
            \State\Return \Call{dk}{$0, t$}$  - [$\Call{dk}{$0, \GPCol{T}'$}$,~$\Call{dkd}{$0, \GPCol{T}'$}$] \cdot (\GPCol{G}\backslash[$\Call{k}{$t, \GPCol{T}'$}$ ,~$\Call{kd}{$t, \GPCol{T}'$}$]')$
        \EndFunction
%        \State
        \algstore{gphelper}
    \end{algorithmic}
\end{algorithm}
\begin{algorithm}
    \begin{algorithmic}[1]
        \algrestore{gphelper}
        \Function{V0df}{$t,\GPCol{T},\GPCol{G}$}
            \LineComment{posterior covariance of function value and gradient at $t=0$ and $t$ respectively}
            \State\Return \Call{kd}{$0, t$} $ - [$\Call{k}{$0, \GPCol{T}'$}$,~$\Call{kd}{$0, \GPCol{T}'$}$] \cdot (\GPCol{G}\backslash[$\Call{dk}{$t, \GPCol{T}'$}$ ,~$\Call{dkd}{$t, \GPCol{T}'$}$]')$
        \EndFunction
        \State
        \Function{Vd0df}{$t,\GPCol{T},\GPCol{G}$}
            \LineComment{same as \Call{V0f}{$\_$} but for gradients}
            \State\Return \Call{dkd}{$0, t$}$  - [$\Call{dk}{$0, \GPCol{T}'$}$,~$\Call{dkd}{$0, \GPCol{T}'$}$] \cdot (\GPCol{G}\backslash[$\Call{dk}{$t, \GPCol{T}'$}$ ,~$\Call{dkd}{$t, \GPCol{T}'$}$]')$
        \EndFunction
    \end{algorithmic}
\end{algorithm}

\begin{algorithm}
\begin{algorithmic}[1]
\State \Separator
\LineComment{all following procedures use the same design parameter:}
\State $\offs \gets 10$ %\ColorComment{\gp~offset for numerical stability}
\State \Separator
\Function{k}{$a,b$} \label{proc:k}
\LineComment{Wiener kernel integrated once in each argument}
\State \Return $\nicefrac{1}{3}\Hm\min(a + \offs, b + \offs)^{\Hm 3} + 0.5\Hm|a - b| \Hm \min(a + \offs, b + \offs)^{\Hm 2}$
\EndFunction
\State
\Function{kd}{$a,b$}
\LineComment{Wiener kernel integrated in first argument}
\State \Return $0.5\Hm(a < b) \Hm (a + \offs)^{\Hm 2} + (a \geq b) \Hm\left((a + \offs) \cdot (b + \offs) - 0.5 \Hm (b + \offs)^{\Hm 2}\right)$
\EndFunction
\State
\Function{dk}{$a,b$}
\LineComment{Wiener kernel integrated in second argument}
\State \Return $0.5\Hm(a > b) \Hm (b + \offs)^{\Hm 2}  + (a \leq b) \Hm ((a + \offs) \cdot (b + \offs) - 0.5 \Hm (a + \offs)^{\Hm 2})$
\EndFunction
\State
\Function{dkd}{$a,b$}
\LineComment{Wiener kernel}
\State \Return $\min(a + \offs, b + \offs)$
\EndFunction
\State
\Function{ddk}{$a,b$}
\LineComment{Wiener kernel integrated in second argument and 1x derived in first argument}
\State \Return $(a \leq b)\Hm (b - a)$
\EndFunction
\State
\Function{ddkd}{$a,b$}
\LineComment{Wiener kernel 1x derived in first argument}
\State \Return $(a \leq b)$
\EndFunction
\algstore{myalg}
\end{algorithmic}
\end{algorithm}

\begin{algorithm}
  \begin{algorithmic}
    \algrestore{myalg}
%\State
\Function{dddk}{$a,b$}
\LineComment{Wiener kernel 2x derived in first argument and integrated in second argument}
\State \Return $-(a \leq b)$
\EndFunction

\end{algorithmic}
\end{algorithm}

\begin{algorithm}
    %\caption{Wolfe Conditions}\label{alg:wolfe}
    \begin{algorithmic}[1]
        \Function{probWolfe}{$t,\GPCol{T},\GPCol{A},\GPCol{G}$}
        \LineComment{design parameters}
        \State $\cone \gets 0.05$ \ColorComment{constant for Armijo condition}
        \State $\ctwo  \gets 0.5$ \ColorComment{constant for curvature condition}
\State $\pwolfe \gets 0.3$ \ColorComment{threshold for Wolfe probability}
        \State
        \LineComment{mean and covariance values at start position ($t=0$)}
        \State $m_0 \gets $  \Call{m}{$0,\GPCol{T},\GPCol{A}$}
        \State $dm_0 \gets $\Call{d1m}{$0,\GPCol{T},\GPCol{A}$}
        \State $V_0 \gets $ \Call{V}{$0,\GPCol{T},\GPCol{G}$}
        \State $Vd_0 \gets $ \Call{Vd}{$0,\GPCol{T},\GPCol{G}$}
        \State $dVd_0 \gets $ \Call{dVd}{$0,\GPCol{T},\GPCol{G}$}
        \State
        \LineComment{marginal mean and variance for Armijo condition}
        \State $m_a \gets m_0 - $\Call{m}{$t,\GPCol{T},\GPCol{A}$}$ + \cone \cdot t \cdot dm_0$
        \State $V_{aa} \gets V_0 + (\cone \cdot t)^{ 2} \cdot dVd_0 + $\Call{V}{$t$}$ + 2 \cdot (\cone \cdot t \cdot (Vd_0 - $\Call{Vd0f}{$t$}$)- $\Call{V0f}{$t$}$)$
        \State
        \LineComment{marginal mean and variance for curvature condition}
        \State $m_b  \gets $\Call{d1m}{$t$}$ - \ctwo \cdot dm_0$
        \State $V_{bb}  \gets \ctwo^2 \cdot dVd_0 - 2 \cdot \ctwo \cdot $\Call{Vd0df}{$t$}$ + $\Call{dVd}{$t$} 
        \State
        \LineComment{covariance between conditions}
        \State $V_{ab} \gets -\ctwo \cdot (Vd_0 + \cone \cdot t \cdot dVd_0) 
        + \ctwo \cdot $\Call{Vd0f}{$t$}$ 
        +             $\Call{V0df}{$t$}$ 
        + \cone \cdot t \cdot $\Call{Vd0df}{$t$}$- $\Call{Vd}{$t$}
        \State
        \LineComment{extremely small variances $\rightarrow$ very certain (deterministic evaluation)}
        \If{$V_{aa} \leq 10^{-9}$ \textbf{and} $V_{bb} \leq 10^{-9}$}
            \State $p_{\text{Wolfe}} \gets (m_a \geq 0)\cdot(m_b \geq 0)$
            \State
            \LineComment{accept?}
            \State $p_{\text{acc}} \gets p_{\text{Wolfe}}>\pwolfe$
            \State \Return \returnCol{$p_{\text{acc}}$}
        \EndIf
        \State
        \LineComment{zero or negative variances (maybe something went wrong?)}
        \If{$V_{aa} \leq$ 0 \textbf{or} $V_{bb} \leq$ 0}
            \State \Return \returnCol{0}
        \EndIf
\algstore{myalg}
\end{algorithmic}
\end{algorithm}

\begin{algorithm}
  \begin{algorithmic}
    \algrestore{myalg}
%        \State
        \LineComment{noisy case (everything is alright)}
    \LineComment{correlation}
        \State $\rho \gets V_{ab}/\sqrt{V_{aa} \cdot V_{bb}}$  
        \State
        \LineComment{lower and upper integral limits for Armijo condition}
        \State $low_a \gets -m_a/\sqrt{V_{aa}}$ 
        \State $up_a \gets +\infty$ 
        \State
        \LineComment{lower and upper integral limits for curvature condition}
        \State $low_b \gets -m_b/\sqrt{V_{bb}}$ 
        \State $up_b \gets \left(2 \cdot \ctwo \cdot \left(|dm_0| +2 \cdot \sqrt{dVd_0}\right)-m_b\right)/\sqrt{V_{bb}}$ 
        \State
        \LineComment{compute Wolfe probability}
        \State $p_{\text{Wolfe}} \gets $\Call{bvn}{$low_a,up_a,low_b,up_b,\rho $}
        \State
        \LineComment{accept?}
        \State $p_{\text{acc}} \gets p_{\text{Wolfe}}>\pwolfe$
        \State \Return \returnCol{$p_{\text{acc}}$}
        \State
        \BoxComment{\centering The function \Call{bvn}{$low_a, up_a, low_b, up_b, \rho$} evaluates the 2D-integral 
          \begin{equation*}
            \int_{low_a}^{up_a}\int_{low_b}^{up_b}
            \mathcal{N}\left(
                \begin{bmatrix}
                  a \\
                  b
                \end{bmatrix};
                \begin{bmatrix}
                  0 \\
                  0
                \end{bmatrix},
              \begin{bmatrix}
                  1 & \rho \\
                  \rho & 1
                \end{bmatrix}
            \right)\mathrm{d}a\mathrm{d}b.
          \end{equation*}}
        \State
        \EndFunction
    \end{algorithmic}
\end{algorithm}

\begin{algorithm}
\begin{algorithmic}[1]
\Function{gaussCDF}{$z$}
\LineComment{Gauss cumulative density function}
\State \Return $0.5 \Hm\left(1 + \erf(z / \sqrt{2})\right)$
\EndFunction
\State
\Function{gaussPDF}{$z$}
\LineComment{Gauss probability density function}
\State \Return $\exp\left({-0.5\Hm z^{\Hm 2}}\right)\Hd\sqrt{2 \pi}$
\EndFunction
\State
\Function{expectedImprovement}{$m,s,\eta$}
\LineComment{\citet{jones1998efficient}}
\State \Return $(\eta - m)\Hm$\Call{gaussCDF}{$(\eta - m)\Hd s$} $+ s \Hm$\Call{gaussPDF}{$(\eta - m)\Hd s$}
\EndFunction
\end{algorithmic}
\end{algorithm}

% ====================================================================
\FloatBarrier
%\clearpage
\vskip 0.2in
\small
\bibliography{bibfile}

\end{document}